\newcounter{myctr}

\documentclass{ws-ijhr}

\usepackage{graphicx}            %
\usepackage{amsmath}             %
\usepackage{amssymb}             %
\usepackage{amsfonts}            %
\usepackage{enumitem}            %
\usepackage{multirow}            %
\usepackage{siunitx}             %
\usepackage{url}                 %
\usepackage{xspace}              %
\usepackage[T1]{fontenc}         %
\usepackage{hyperref} %
\usepackage{balance}
\usepackage[bottom]{footmisc}
\usepackage{todonotes}

\graphicspath{{figures/}}
\DeclareGraphicsExtensions{.pdf,.png,.jpg,.jpeg}

\newcommand{\degreem}{^{\circ}} %

\newcommand{\seclabel}[1]{\label{sec:#1}}

\newcommand{\figlabel}[1]{\label{fig:#1}}
\newcommand{\tablabel}[1]{\label{tab:#1}}
\newcommand{\eqnlabel}[1]{\label{eqn:#1}} 

\newcommand{\secref}[1]{Section~\ref{sec:#1}\xspace}

\newcommand{\figref}[1]{Fig.~\ref{fig:#1}\xspace}
\newcommand{\tabref}[1]{Table~\ref{tab:#1}\xspace}

\newcommand{\nop}{NimbRo\protect\nobreakdash-OP\xspace}

\newcommand{\noptwo}{NimbRo\protect\nobreakdash-OP2\xspace}
\newcommand{\nopx}{NimbRo\protect\nobreakdash-OP2X\xspace}

\newcommand{\cmnew}{CM740\xspace}

\newcommand{\igus}{igus\textsuperscript{\tiny\circledR}\xspace}
\newcommand{\iguhop}{igus Humanoid Open Platform\xspace}

\newcommand{\degree}{$\degreem$\xspace}

\begin{document}

\markboth{G. Ficht, H. Farazi et al.}{NimbRo-OP2X: Affordable Adult-sized 3D-printed Open-Source Humanoid Robot for Research}

\catchline{}{}{}{}{}

\title{NIMBRO-OP2X: AFFORDABLE ADULT-SIZED 3D-PRINTED\\OPEN-SOURCE HUMANOID ROBOT FOR RESEARCH}

\author{GRZEGORZ FICHT\textsuperscript{1}, HAFEZ FARAZI\textsuperscript{1}, DIEGO RODRIGUEZ\textsuperscript{1}, \\
DMYTRO PAVLICHENKO\textsuperscript{1}, PHILIPP ALLGEUER\textsuperscript{1}, \\
ANDR\'E BRANDENBURGER\textsuperscript{2}, SVEN BEHNKE\textsuperscript{1}}

\address{Institute for Computer Science VI, University of Bonn\\
Endenicher Allee 19a, 53115 Bonn, Germany\\
\textsuperscript{1}\{ficht, farazi, rodriguez, pavliche, pallgeuer, behnke\}@ais.uni-bonn.de\\
\textsuperscript{2}andre.brandenburger@uni-bonn.de}

\maketitle

\begin{history}
\received{Day Month Year} %
\revised{Day Month Year}  %
\accepted{Day Month Year} %
\end{history}

\begin{abstract}

For several years, high development and production costs of humanoid robots restricted researchers interested in working in the field.  
To overcome this problem, several research groups have opted to work with simulated or smaller robots, whose acquisition costs are significantly lower. 
However, due to scale differences and imperfect simulation replicability, results may not be directly reproducible on real, adult-sized robots. 
In this paper, we present the \nopx, a capable and affordable adult-sized humanoid platform aiming to significantly lower the entry barrier for humanoid robot research.
With a height of \SI{135}{cm} and weight of only \SI{19}{kg}, the robot can interact in an unmodified, human environment without special safety equipment. 
Modularity in hardware and software allow this platform enough flexibility to operate in different scenarios and applications with minimal effort. 
The robot is equipped with an on-board computer with GPU, which enables the implementation of state-of-the-art approaches for object detection and human perception demanded by areas such as manipulation and human-robot interaction. 
Finally, the capabilities of the \nopx, especially in terms of locomotion stability and visual perception, are evaluated. 
This includes the performance at RoboCup 2018, where \nopx won all possible awards in the AdultSize class.

\end{abstract}


\section{Introduction}

The vision of humanoid robots performing work alongside humans has been the motivation for their development, by both researchers and companies.
Ever since 1973, when the Waseda Robot~\cite{wabot} was presented, numerous attempts have been made to produce humanoid robots capable
of navigating and interacting in unaltered human environments. One of these efforts originated in Japan with the development of Honda's P series robots~\cite{Hirai1998}, 
which finally led to the emergence of Asimo~\cite{asimo}. At a similar time,
the joint efforts of Kawada Industries, Kawasaki Heavy Industries and AIST, produced the HRP series of robots~\cite{yokoi2004experimental}~\cite{hrp2}~\cite{kaneko2008humanoid}~\cite{kaneko2011humanoid}~\cite{Kaneko2009}.
Only until recently, these robots (especially the HRP-2~\cite{hrp2}) were the only universal human-sized platforms that allowed for 
multi-faceted research. This has recently changed with the introduction of multiple new full-sized humanoid robots all over the world. 
Examples include the german DLR-TORO~\cite{englsberger2014overview}, american NASA Valkyrie~\cite{radford2015valkyrie} and spanish TALOS~\cite{stasse2017talos}.

In terms of availability, however, the situation has not significantly changed. 
The platforms either remain closed-source, or are sold at a premium price. 
To cope with this problem, many researchers work with smaller robots in miniaturized environments or simulators. 
Nonetheless, the gap between simulation and reality is still an unresolved problem, and thus, the results obtained may not always translate directly into real-world applications. 

Not only the acquisition but also the operation of real robot hardware can prove to be problematic.
Production and servicing time are aspects that needs to be also factored in.
A broken part often requires to be sent to the manufacturer for repairing, halting development for weeks.
Reducing the cost, complexity and minimum knowledge barrier would allow the parts to be repaired or completely rebuilt by the end-user in a matter of days.
Breaking of components critical for operation would no longer be such a limiting factor.
To further research in humanoid robotics, an inexpensive, openly available and capable platform is necessary. We have contributed to the goal of 
producing such a platform, starting with the \SI{90}{cm} tall \nop~\cite{Schwarz2012}, where both the hardware and ROS-based software were open-source. 
The manufacture and assembly was quite complex with numerous milled aluminum and carbon-composite parts. This process was streamlined and 
simplified with the \iguhop~\cite{Allgeuer2015b} by transitioning to a fully 3D printed design. Both of these robots were child-sized however, which limited
their final use cases. Expanding on the flexibility provided by 3D printing, a larger, asimo-sized \noptwo~\cite{ficht2017nop2} was developed.
By using commercially available components, we kept the overall cost down while still demonstrating impressive performance~\cite{Ficht2018Grown}. 

The experience gained with all of these platforms led to the creation of the \nopx~\cite{ficht2018nimbro}, which is the subject of this paper. Its chief characteristics include a completely 3D printed structure,
(including external gears in several joints), an on-board GPU-enabled computing unit in a standard Mini-ITX form factor, a new series of more powerful, intelligent
actuators, and an overall cost reduction. Both the hardware~\cite{NOP2Hardware} and software components~\cite{IguhopSoftware} of the robot are completely open-source. By providing
the research community with a light-weight, robust robot, equipped with a GPU for parallel computing, we hope to foster research not only limited to motion planning but deep learning as well.

The remainder of this paper is organized as follows. In Sec.~\ref{design}, the design concepts used in the \nopx are described. Sections \ref{hardware} and \ref{software}
describe the robot's open-source hardware and software accordingly. In Sec.~\ref{eval} the capabilities of \nopx are assessed both in a qualitative and quantitative way.

\section{Core Design Concept} 
\label{design}
\begin{figure}[t]
	\centering
	\includegraphics[height=0.45\linewidth]{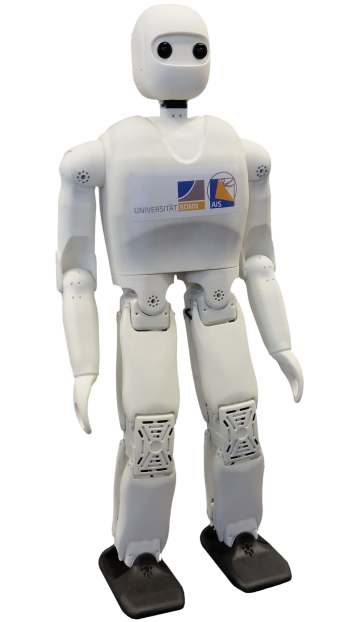}\hspace{1cm}
	\includegraphics[height=0.45\linewidth]{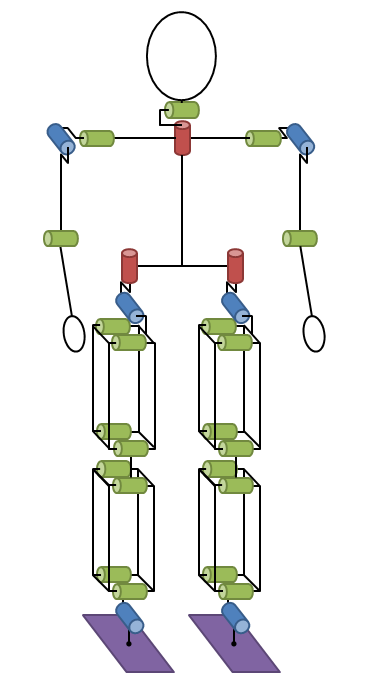}
	\caption{Complete, assembled \nopx (left) along with its kinematics (right).}
	\figlabel{op2x_robot}
\end{figure}

The main goal when designing the \nopx was to create a minimalistic but highly-capable robot, which can act as a baseline for different custom applications.
Such a platform requires a customizable baseline, with a wide range of features that works with minimum effort.
Modularity plays a key role in achieving this, in both the hardware and software of the robot. This mindset is the result of our experience
with creating capable, open-source and cost-effective humanoid robots. 

With respect to hardware, modularity can be achieved by maximizing the usage of standard and commercial off-the-shelf (COTS) components.
3D printing can complement this approach by providing high flexibility when modifying and replacing structural parts.
Such a design procedure minimises the amount of procurement, development and maintenance problems. 
Exchanging COTS components might require certain 3D printed parts to be redesigned,
this however can be done in-house and without excessive lead-time. 
Additionally, parts can be modified for specific tasks, e.g., by adding grippers or by altering the foot shape. 
By focusing on the usage of standard components instead of custom ones, 
the platform is not only modular but also cost-efficient. 
Actuators can be purchased in bulk, while a computing unit can be selected based on
user criteria. 
These could involve computing power, thermal properties, weight, price and availability. 
Although a custom-tailored solution might provide beneficial results, 
it requires time to develop into a reliable and mature product.
These developments lead to an extended project duration and shift the focus away from the main goal. 

The flexibility of the hardware requires equally adaptable software to fully utilize the spectrum of possibilities. 
A complete framework is needed which clearly separates, and implements commonly used lower-level device control and several layers of abstraction. 
The Robot Operating System (ROS) middleware serves this purpose well. 
With ROS, large control tasks can be split into smaller and separate ROS processes called \textit{nodes},
which communicate over a defined interface by publishing \textit{messages} on to \textit{topics}.
A plugin-scheme allows to further expand on the modularity, as the base functions can be reused.

The complete platform has been open-sourced in order to foster research in the area of humanoid robotics.
Providing access to an easily adaptable robot allows researchers to focus on their respective topics. 
By building a community around a common platform, efforts which focus on fundamental features required to operate the robot, are not duplicated.
The entry barrier for using real hardware is greatly lowered and allows for a wider dissemination and accelerated progress of the state of the art.

\section{Hardware Description} 
\label{hardware}

\begin{table}
\renewcommand{\arraystretch}{1.2}
\caption{\nopx specifications.}
\tablabel{OP2X_specs}
\centering
\footnotesize
\begin{tabular}{c c c}
\hline
\textbf{Type} & \textbf{Specification} & \textbf{Value}\\
\hline
\multirow{4}{*}{\textbf{General}} & Height \& Weight & \SI{135}{cm}, \SI{19}{kg}\\
& Battery & 4-cell LiPo (\SI{14.8}{V}, \SI{8.0}{Ah})\\
& Battery life & \SI{20}{}--\SI{40}{\minute}\\
& Material & Polyamide 12 (PA12)\\
\hline
\multirow{6}{*}{\textbf{PC}}
& Mainboard & Z370 Chipset, Mini-ITX\\
& CPU & Intel Core i7-8700T, \SI{2.7}{}--\SI{4.0}{GHz}\\
& GPU & GTX 1050 Ti, 768 CUDA Cores\\
& Memory & \SI{4}{GB} DDR4 RAM, \SI{120}{GB} SSD\\
& Network & Ethernet, Wi-Fi, Bluetooth\\
& Other & 8$\,\times\,$USB 3.1, 2$\,\times\,$HDMI, DisplayPort\\
\hline
\multirow{4}{*}{\textbf{Actuators}} 
& Total & 34$\,\times\,$Robotis XH540-W270-R\\
& Stall torque & \SI{12.9}{Nm} \\
& No load speed & \SI{37}{rpm} \\
& Control mode & Torque, Velocity, Position, Multi-turn\\
\hline
\multirow{6}{*}{\textbf{Sensors}} & Encoders & 12\,bit/rev\\
& Joint current (torque) & 12\,bit\\
& Gyroscope & 3-axis (L3G4200D chip)\\
& Accelerometer & 3-axis (LIS331DLH chip)\\
& Camera & Logitech C905 (720p)\\
& Camera lens & Wide-angle lens with 150\degree\!FOV\\
\hline
\end{tabular}
\end{table}

A summary of the hardware features in our configuration of the \nopx is shown in \tabref{OP2X_specs}.
With a height of \SI{135}{cm} the robot is large enough to allow meaningful interaction in an unmodified, human environment. 
The low weight of approximately \SI{19}{kg} makes operation of the robot easy and safe, as a gantry is unnecessary.
The fully 3D printed structure of the robot contributes greatly to this, as the utilized ribbing provides the needed structural rigidity with little additional weight. 
The robot is equipped with 34 actuators, organized into 18 joints through the efficient use 
of geared transmissions, and a mixture of parallel and serial kinematics.
\nopx also possesses a wide array of sensors that provide feedback to the control methods.
Due to usage of standardized and COTS components and the 3D printed nature of the robot, 
the functionality as well as the structure can be further expanded upon to meet user demands.
As the customizations with respect to user requirements are applied at a much later stage (or not at all), the overall entry level costs remain low and increase proportionally only by adding features.
A set of the \nopx components required to achieve the specifications in \tabref{OP2X_specs} is in a similar price 
range to that of the popular stationary dual-arm research platform Baxter. Compared to similarly sized humanoid robots, \nopx
is at least an order of magnitude less expensive to procure, and does not have fees associated with software licensing. 

\subsection{Exoskeleton} 

\begin{figure}[t]
\parbox{\linewidth}{\centering\includegraphics[width=0.95\linewidth]{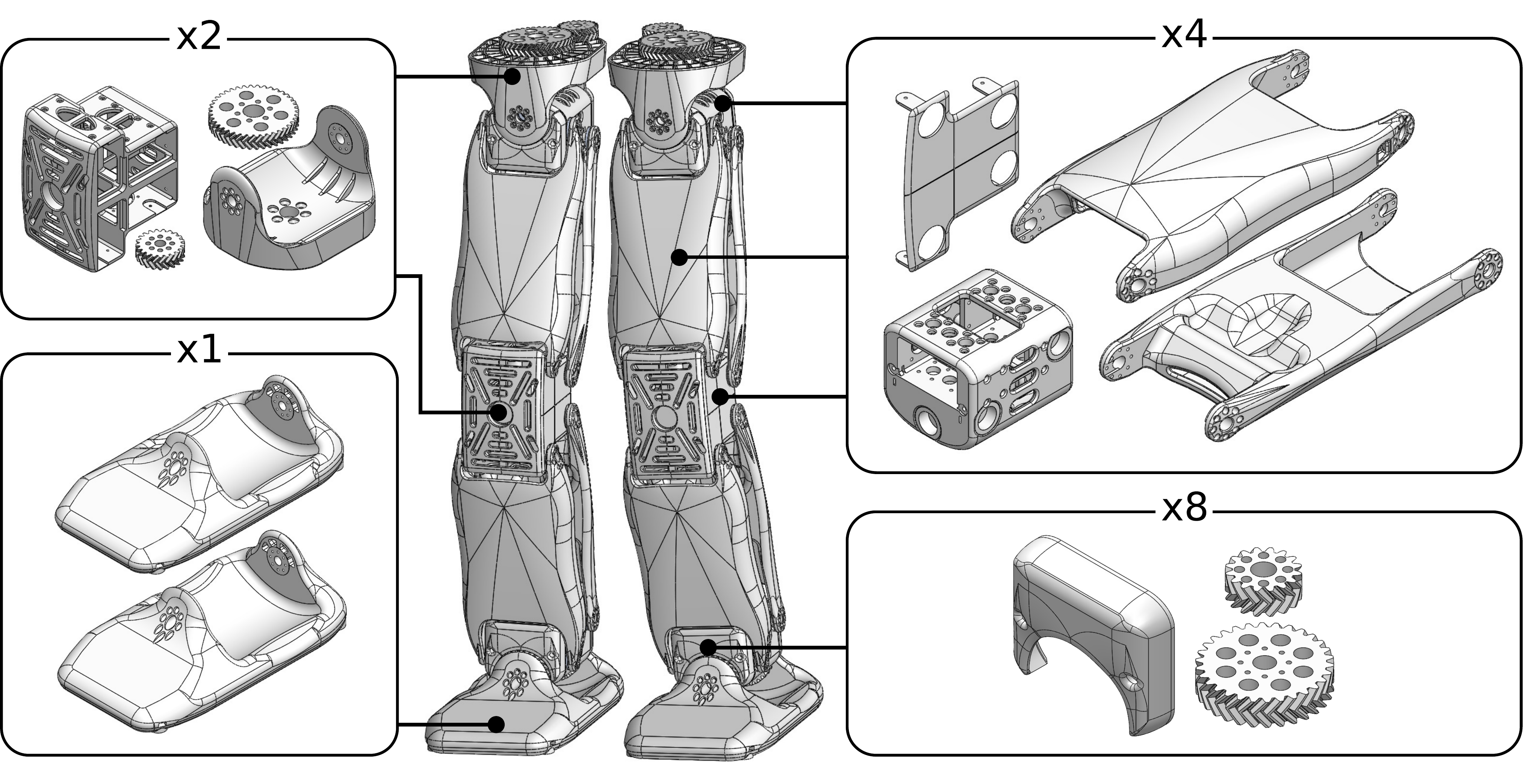}}
\caption{Reuse and symmetry of the parts in the legs. The feet are the only unique components in the complete leg kinematic chain.}
\figlabel{legsCAD}
\end{figure}
The 3D printed frame of the \nopx is one of its distinctive features. The plastic parts are made using a 
Selective Laser Sintering (SLS) process from Polyamide 12 (PA12) material in layers of \SI{0.1}{mm} increments. 
Compared to the more-common additive process of Fused Deposition Modeling (FDM), the layers are fused together 
making the connection between them much stronger. This prevents any layer delamination, making for a more uniform and rigid structure. 
Due to the increased strength and stiffness, parts made with SLS technology can be used not only for appearance, 
but to bear the whole load of the robot. Combined with the ability to produce shapes of varied complexity, 
the production process is streamlined and the total number of parts can be greatly reduced. The parts have also been
optimized to minimize weight while maintaining the necessary rigidity. This is achieved through a mixture of wall-thickness
variation and strategic usage of ribbing structures printed directly into the frame. 

At its core the \nopx is a meticulous redesign of the \noptwo~\cite{ficht2017nop2}, that introduces several upgrades over its predecessor.
Multiple portions of the design have been remodeled to be structurally stronger, while lowering the entire weight of the 3D printed shell by approximately \SI{0.5}{kg}.
The total weight was reduced in spite of increasing the size of the torso, which is the single largest and heaviest part of the robot.
The change was made in order to accommodate for a larger computing unit, based on a standard Mini-ITX motherboard.
The weight saving was mostly done by making the leg parts narrower. As the legs in the \noptwo were made out of simple shapes and flat surfaces,
they were prone to bending. The narrower leg and the overall rounder and smoother design provides a significant improvement in terms of rigidity.
In total, the plastic parts weight is \SI{10.334}{kg}, which accounts for the \SI{54}{\%} of the complete mass of the robot.
In general, the metal structure of robotic platforms are used to dissipate the heat coming off from the actuators, this is 
unfortunately not possible with our light-weight plastic parts. This issue has been resolved by incorporating venting holes into the structure
of the components in key locations. The design of multiple components also takes advantage of the natural symmetry of humanoids.
This is especially evident in the legs, as it can be seen on \figref{legsCAD}.

\subsection{Actuation} 

\begin{figure}[t]
\centering
\def\svgwidth{0.9\linewidth}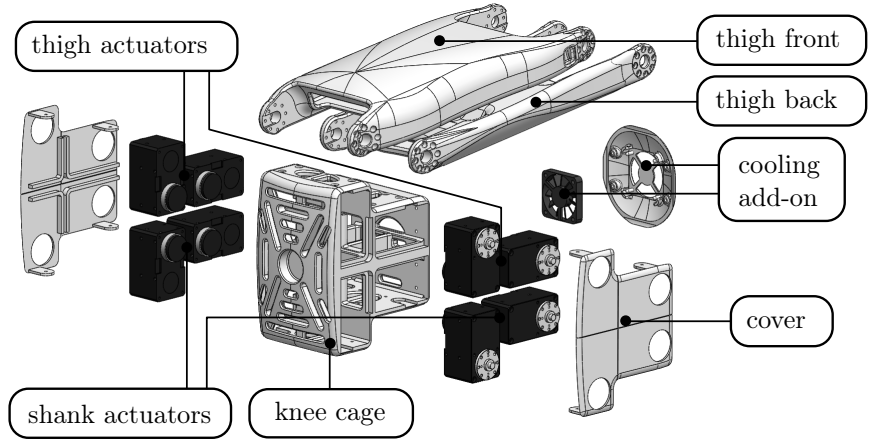
\caption{Knee design incorporating parallel kinematics. Four upper and lower actuators control the thigh and shank respectively. 
All of the servos slide into place and are fastened with screws. A fan can be mounted on the back of the knee for additional cooling. }
\figlabel{kneeCAD}
\end{figure}

One of the factors that contributed to the refined appearance of the \nopx was the decision to use a new series of intelligent actuators.
Incorporating the Robotis Dynamixel XH540 into the \nopx brought several improvements over the MX106 from the \noptwo.
The more notable physical improvements of the XH540 include a \SI{29}{\%} increase in output torque, a fully metal casing for heat dissipation and redesigned gearbox.
These not only increase the total power of the robot, but add to the overall durability and uniform operation under stress.
On the software side, the actuators can be controlled using the same protocol as the previous generation, but provide an additional subset of features.
Through the inclusion of reliable current sensing, the standard position controller is enhanced with torque control.
The feature that adds the most versatility to the design is the external port, which allows for effortless sensor integration.
Up to three analog or digital devices can be connected to a single actuator and operated using the standard protocol.

For a humanoid robot of this size, a certain amount of torque is necessary to interact with and navigate the environment in a meaningful way.
A single XH540 would not be enough to sufficiently power the significantly loaded joints, such as the ankle or knee. For this purpose, we have 
decided to incorporate a mixture of parallel kinematics and external gearing in the leg design. The parallel kinematics are powered by eight actuators,
with all of them located in the knee. The thigh and shank each utilize four servos with one acting as a master for the three slaves. This ensures synchronized movement
between the front and back parts of the legs, which is necessary to provide uniform, unopposed leg movement.
In spite of the heat dissipation capabilities, a high concentration of actuators in such a compact knee design produces a considerable amount of heat when under load.
We incorporated a fan add-on into the design, that pulls air through the knee vents to cool down the whole joint. A blow-up of the knee construction, along with the upper part
of the parallel kinematics can be seen on \figref{kneeCAD}. 

\begin{figure}[t]
\centering
\def\svgwidth{0.9\linewidth}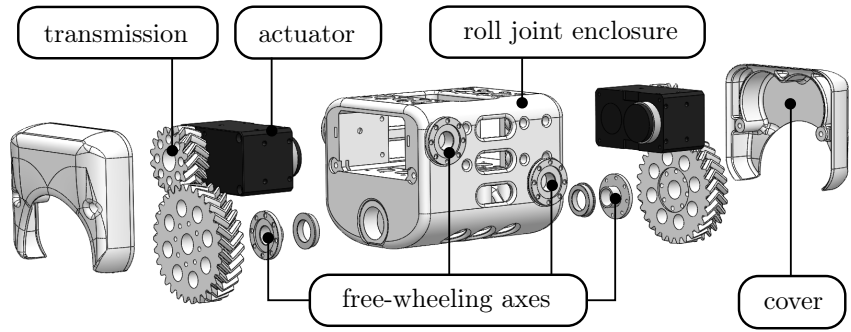
\caption{Design of the hip and ankle roll joints. Two actuators with reduction gearing provide the necessary torque to support the robot's weight. 
The free-wheeling axes on the side connect the thigh or shank part to the joint. The driven gear further attaches to the hip bracket or foot.}
\figlabel{hiproll}
\end{figure}
Placing the complete parallel kinematic mechanism inside of the knee allowed for space savings in the hip and ankle roll joints, which mirror each other by design. 
Here, two servos work together in another master-slave configuration, which drives a set of external gears to further increase the possible torque. The transmission used 
in these joints is 16:30, which provides a raw torque increase of \SI{1.875}{} at the cost of angular velocity. As the produced torque is relatively large, we have 
included light-weight gear covers in the design. These do not only prevent foreign objects from coming in contact with the gearbox and potentially jamming it, but also allow for safer handling of the robot.
What remains in the leg structure is the hip-yaw joint. It is embedded in the robot's torso and is powered by a single actuator with a 21:40 geared reduction. 
The driven gear lays in between the torso and the hip bracket. The whole assembly moves on an \igus PRT slewing ring bearing mounted in the torso, and has a common hole going through 
the rotation axis that allows for effortless cable feeding. The joints with additional transmission are depicted on \figref{hiproll} and \figref{hipyaw}.

\begin{figure}[t]
	\centering
	{
		\fontsize{7pt}{11pt}\selectfont
		\def\svgwidth{0.9\linewidth}
		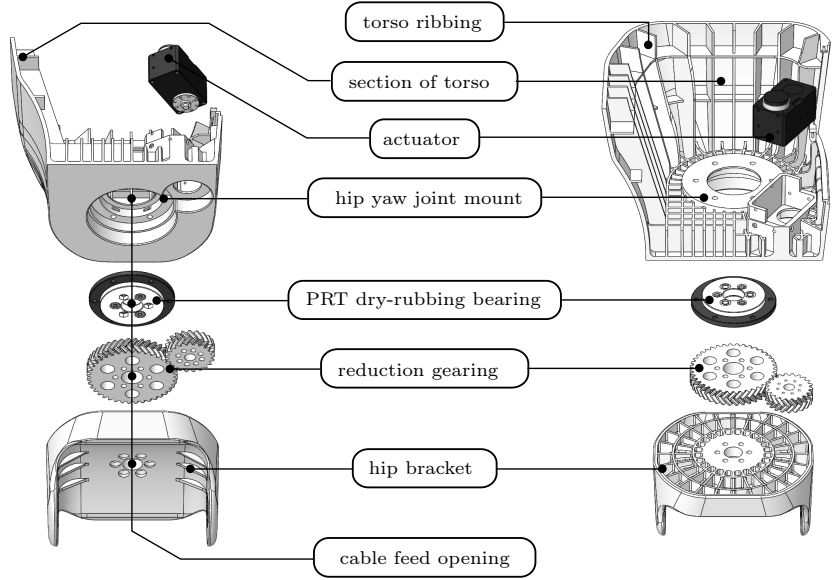
	}
	\caption{Section of torso including the hip yaw joint assembly.}
	\figlabel{hipyaw}
	\vspace{-2ex}
\end{figure}
Instead of the typically used spur gears, we utilize double helical gears which are characterized by a higher torque density and smoother operation. The complex shape of these gears make them difficult 
to produce out of metal with subtractive manufacturing. These limitations are not an issue with 3D printing, where virtually any shape can be recreated with high fidelity.
Apart from selecting a module $m$, the design of helical gears requires a  specified helix angle $\psi$. Its value is usually in the range between \SI{15}{\degree} and \SI{45}{\degree}, with some commonly 
used values being \SI{15}{}, \SI{23}{}, \SI{30}{}, and \SI{45}{\degree}~\cite{gopinath2010gears}. Higher $\psi$ values result in smoother operation, but increase the axial load.
The pitch diameter $d$, required to determine the distance between transmission axes is then:
\begin{equation}
d = \tfrac{Z m}{\cos(\psi)}, \eqnlabel{gear_diameter}
\end{equation}
We chose the helix angle as $\psi = 41.4096\degreem$, and set the module to $m=1.5$. With these values, $\cos(\psi)$ is equal to \SI{0.75}{} which makes the diameter double the tooth count and
simplifies the design process. By using double helical gears we cancel out the produced axial forces, with the two gear halves working in opposite directions.
As with the exoskeleton, gears are printed using an SLS process to increase durability. The material used is the \igus I6, dedicated for 3D printing gears. Its chief characteristis are 
a low friction coefficient and no requirement for lubrication. To increase tooth overlap, total gear thickness was set to \SI{14}{mm}~(\SI{7}{mm} for each side of the helix).
The thicker, 3D printed gears are only \SI{54}{\%} of the weight, when compared to the brass gears of the \noptwo.
Since the robot's creation in june of 2018, we have not had a single gear break or loosen, which proves the durability of the solution.

\subsection{Electronics} 

One of the more notable features of the \nopx is the completely standardized PC in a Mini-ITX form-factor. Such a hardware configuration can be 
modified at will, with respect to the user requirements. Further upgrades can be done at the component level, making for a flexible solution.
The space requirements were taken into account in the torso's design and included a GPU connected into the motherboard's PCIe slot.
This was motivated by the current trends in research, with parallel computing and machine learning techniques at the forefront.
Accordingly, we have fitted the \nopx with an Intel Core i7-8700T processor with 6 cores running 12 threads 
and an Nvidia GTX 1050Ti GPU with 768 CUDA cores. For an on-board computing system, this is more than sufficient to perform calculations
using state-of-the-art computer vision and motion control methods.

The main component of the control electronics is the \cmnew sub-controller board, produced by Robotis from South Korea.
Equipped with a single STM32F103RE microcontroller, the \cmnew handles all of the necessary interfacing 
between the computing unit and connected peripherals. These include all of the XH540 actuators connected on a RS485 bus, 
an integrated 6-axis Inertial Measurement Unit (IMU) composed of a 3-axis accelerometer and gyroscope,
as well as a basic I/O board for direct control over the robot. This interface panel is equipped with three buttons, 
which are used to fade-in and fade-out the robot, start and stop the higher layer behavior control and finally, 
to forcefully reset the power supplied to the servos in case of an emergency. Additionally, seven LEDs (including two RGB)
are used to display information about the robot's state, such as communications status and mode of operation. 
These basic functions can be extended and modified at will, as the microcontroller firmware has been fully rewritten 
and open-sourced. The improvements over the stock firmware were focused mainly on achieving faster and more reliabile 
communication, however other features have been implemented. Further interfacing to the robot can be done through the various
options provided by the chipset on the PC side e.g. USB, HDMI, Ethernet, WiFi, Bluetooth and more. In our case,
USB ports are used to control the robot through a Joystick or connect a Logitech C905 720p camera for visual feedback.
\begin{figure*}
	\centering
	\includegraphics[width=0.80\linewidth, trim={50pt 5pt 80pt 10pt}, clip]{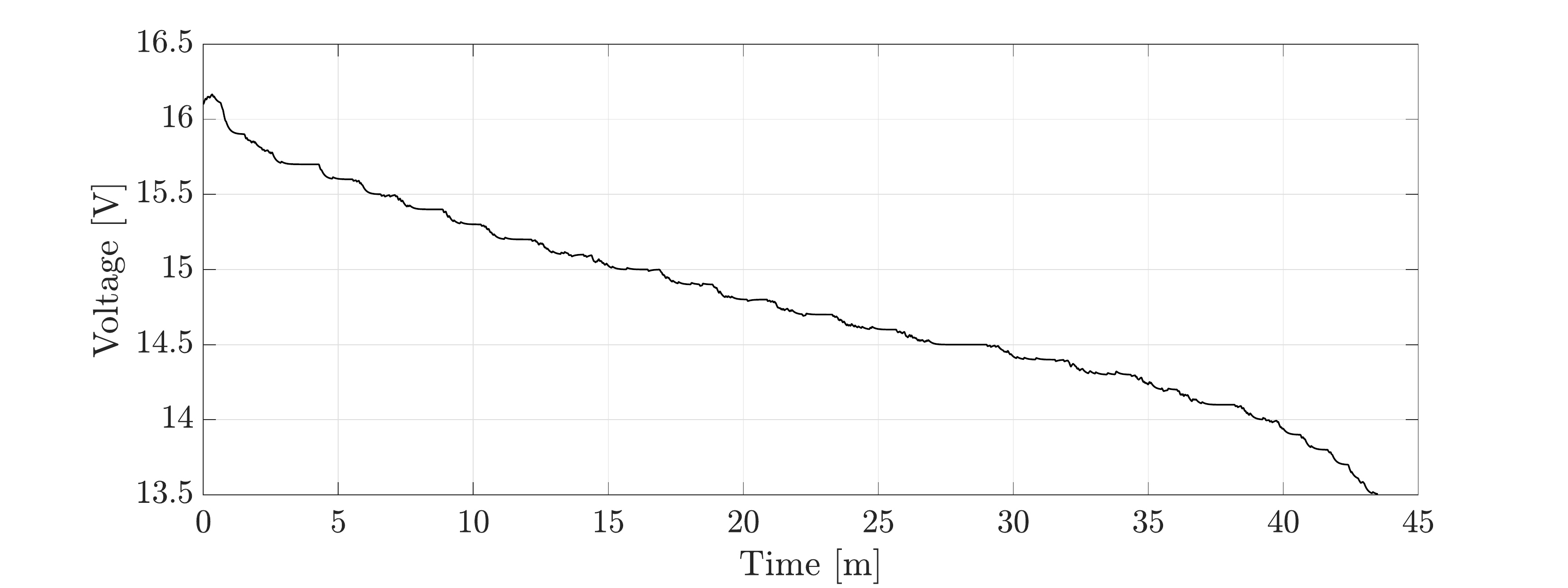}
	\caption{Discharge rate of the battery under normal operation, when the robot is standing with the CPU and GPU under full load.}
	\figlabel{battery}
\end{figure*}

The whole system is powered through a 4-cell Lithium Polymer (LiPo) battery. The input is then split to supply power to two independently regulated circuits. 
In this case, the actuators are isolated from the control electronics through a BTS555 smart high-side power switch circuit, which limits inrush currents when turning the robot on. 
Due to the high number of connected devices, this effect could potentially damage the electronics of the \nopx. 
The control electronics, along with the PC---although quite capable in terms of raw computing 
specifications---do not need to consume much current. As an example, our computing setup has a 
Thermal Design Power (TDP) of only 105W, which is powered through a single industrial wide-input 250W M4-ATX power supply.
The size and capacity of the battery has been selected to provide enough runtime, and not add unnecessary weight for the robot to carry.
We decided on an inexpensive, generic, relatively small($\SI{170}{}\times\SI{65}{}\times\SI{30}{mm}$) and light-weight(\SI{760}{g} fully charged) 8000mAh battery.
Despite the small size, it can still power the robot for a relatively long period. During a stress test (for which results are shown on \figref{battery}) when the robot
was standing with the CPU and GPU under full load, the battery lasted 43 minutes until reaching a safe minimum voltage of \SI{3.3}{V} per cell. 
\section{Software Overview} 
\label{software}

\begin{figure}[!t]
	\hspace{-2ex}
	\parbox{\linewidth}{\centering
		\includegraphics[width=0.6\linewidth]{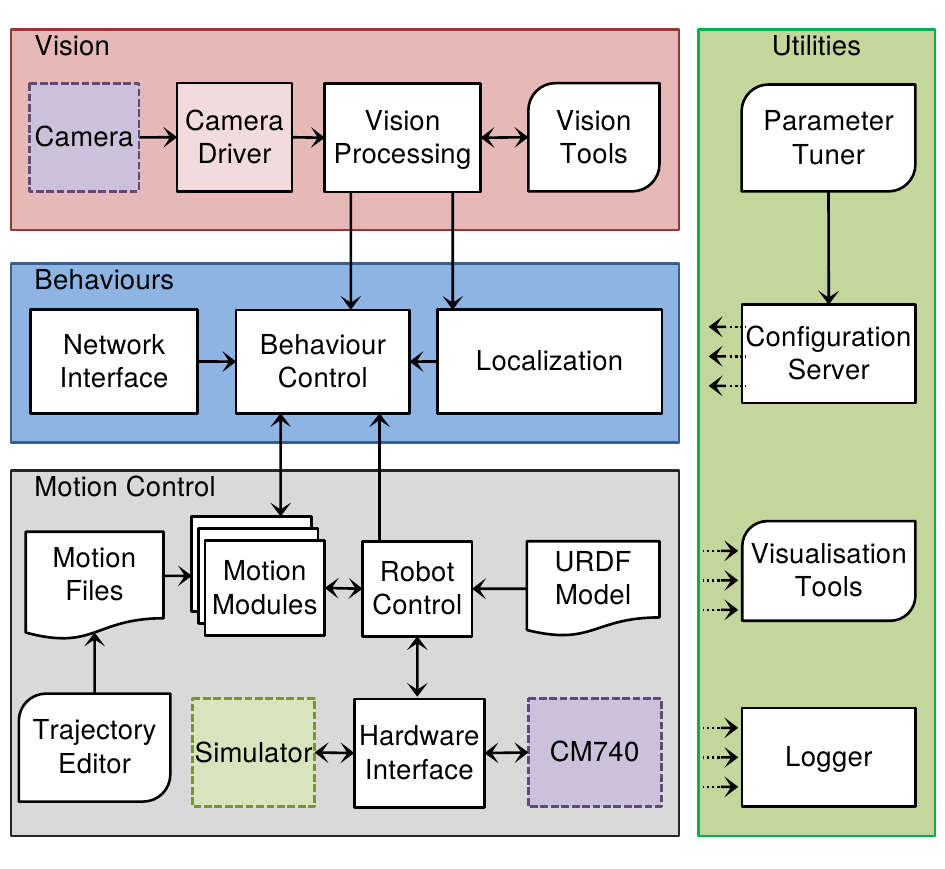}}
	\caption{Simplified architecture of our open-source ROS framework.}
	\figlabel{software_architecture}
	\vspace{-2ex}
\end{figure}

The true capability of the robot comes from our open-source ROS~\cite{quigley2009ros} based software. 
The software repository on GitHub has started in 2013~\cite{schwarz2013humanoid}~\cite{IguhopSoftware} and continues to expand each year. 
We have many unique clone and downloads each week, and many research groups have used it or were inspired by it~\cite{chen2017robocup}~\cite{biddulph2018comparing}~\cite{dehkordiunbounded}~\cite{raziichiro}.  
The software was developed with the target application of humanoid robot soccer, but as we show later on in this section, 
new functionality can be easily implemented and adapted for virtually any other application to be realised. A simplified diagram of the 
framework can be seen in \figref{software_architecture}. 

Each robot can be launched and configured directly on the command line inside SSH sessions of the robot computer. 
Although launching and operating the robot through the command line allows for a significant amount of freedom and flexibility, 
it is time-consuming and prone to errors and requires in-depth knowledge of the framework. To simplify the operation, 
a web application is hosted directly on the robot and is accessible through any standard web 
browser on any device. The web application also hosts a built-in terminal so that full control of the robot can be taken, 
regardless of the operating system.

\subsection{Sensor data processing} 

Obtaining feedback on the current robot state is essential in the operation of closed-loop motion generation algorithms. 
For compatibility reasons, all of our robots (including \nopx) have the actuators running in position control mode, however more modes
are available such as velocity, multi-turn and current-based torque mode. The position is measured through magnetic encoders with a resolution
of 4096 ticks per revolution, and read out using the dynamixel protocol. Further in the ROS software, these values are processed by the hardware 
interface and later interpolated by the tf2 library in combination with the URDF model. The hardware interface also performs
conversions that account for the parallel kinematics and external gearing, as it is not internally supported by ROS. The conversions
are set up automatically, based on dependencies from an extended URDF model with additional virtual serial kinematic leg chains for reference.
Using joint name aliases, the positions and commands are scaled, copied, negated, summed or subtracted to reflect the parallel kinematic behavior. 
One of the chief characteristics of the XH540 actuators is the inclusion of current sensing and control options. The current measurements are quite 
reliable and can be used to accurately measure the torque applied by the motor. We have identified the torque constant $K_T$ of the XH540 through 
linear regression to be \SI{3.8511}{Nm/A}, with a slight offset of \SI{-0.0821}{Nm} attributed to the idle current. The result of the linear regression
can be seen on \figref{torque}.

\begin{figure*}
	\centering
	\includegraphics[width=0.80\linewidth, trim={50pt 5pt 80pt 10pt}, clip]{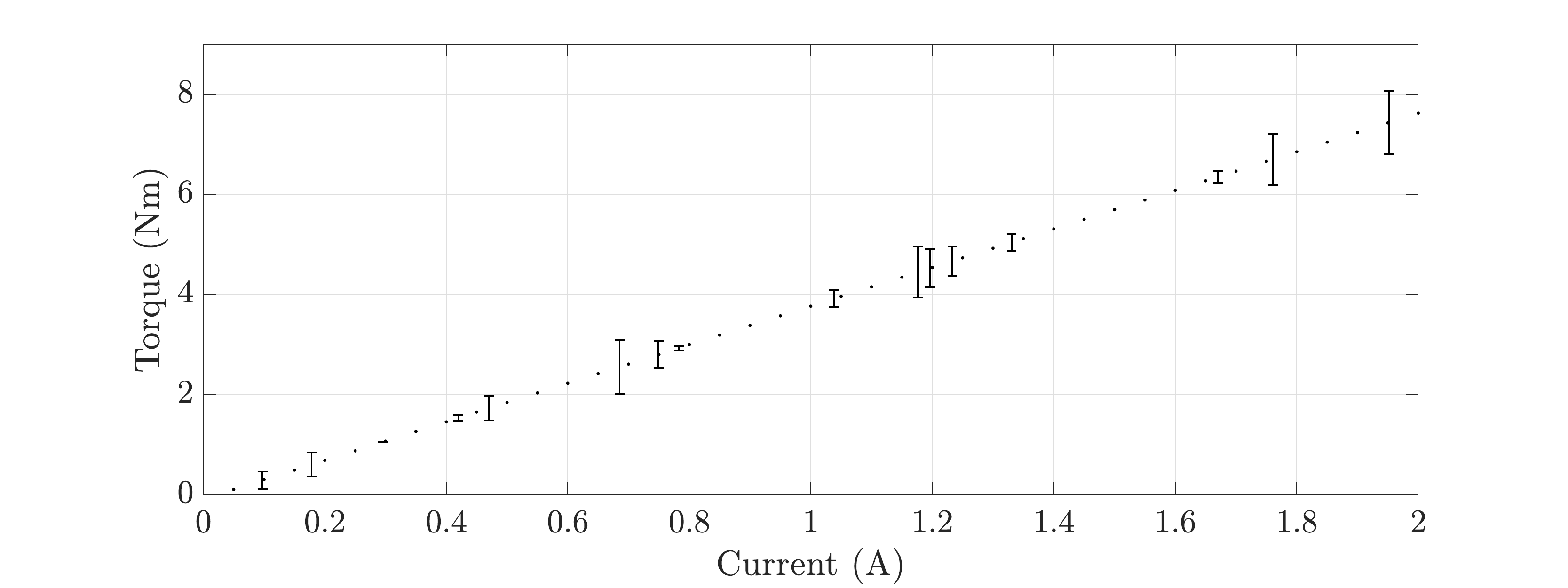}
	\caption{Linear regression performed on applied torque and measured current.}
	\figlabel{torque}
\end{figure*}

The proprioception with the joint sensors is completed with an estimation of the 3D orientation of the robot.
For that purpose, we apply a 3D nonlinear complementary filter to the readings obtained from the combined 6-axis accelerometer and gyrometer IMU.
Adding a 3-axis magnetometer is possible, and can increase the accuracy of the estimation, especially in measuring the rotation around the global z-axis.
The software on the \cmnew microcontroller side already supports this, and requires only that the sensor be connected.
The resulting orientation estimate is then stored in a quaternion format, and can be converted to any desired form of representation.
Fused angles~\cite{AllgeuerFusedAngles} is one of these representations, and is widely used in the gait of the \nopx. The deviations 
from the desired orientation in sagittal and lateral planes can be separated, which is useful for performing corrective, balancing actions.

Each of the robots, exclusively perceive the environment through a Logitech C905 webcam equipped with a wide-angle lens (150\degree) and infrared cut-off filter. The wide-angle lens is helping the robot to see more of the field within a single captured frame. The inferred cut-off filter is useful to perceive more stable and realistic colors in the presence of the white light. For projecting the identified object into an ego-centric world view, we need camera parameters. An standard chessboard calibration is used to find intrinsic camera parameters. For online estimation of the extrinsic parameters, we consider known kinematics and posture of the robot. Despite having the kinematic model of the robot, small hardware variations still result in huge projection errors, especially for distant objects. We remedy this by utilizing Nelder-Mead~\cite{nelder1965simplex} method to calibrate the position and orientation of the camera frame in the head. More details about our vision system are described in Sec.~\ref{vision_sec}.

\subsection{Gait generation}

Given that the actuator control scheme handles the low level tracking 
performance of the joints of the robot in a coordinated way, motions such as
walking gaits can be built on top. 
Bipedal walking is an important skill for humanoid robots, 
as it is the primary form of locomotion and motion, 
especially for example in the context of the RoboCup competition, where the 
\nopx has to navigate and traverse an artificial grass field. 
The variability and unpredictability of such a surface is a source 
of disturbances during walking, and hence mandates a robust gait trajectory and 
feedback loop for success. The main gait generation and stabilization algorithm 
used on the \nopx is presented as follows, followed by the description of an 
approach to optimize the sagittal arm gains using Bayesian optimization.

\subsubsection{Walking with direct fused angle feedback}

The walking algorithm of the \nopx is based on producing joint trajectories 
that result in a semi-stable open-loop walking gait, and then 
to stabilize this gait through integration of multiple heuristics, called \textit{corrective actions}. 
This results in a modification of the gait trajectories 
in order to specifically affect the balance in a targeted way. 
The open-loop portion of the gait is generated by a central pattern generator algorithm that can essentially be 
summarized as a map
\begin{equation}
f_{CPG}: (-\pi,\pi] \to \mathbb{J},\, \mu \mapsto J,
\end{equation}
where $\mu$ is the so-called gait phase, $J$ is a joint configuration of the 
robot, and $\mathbb{J}$ is the set of all possible such configurations. When the 
robot is commanded to walk, $\mu$ starts at 0, and is incremented in each time 
step by a constant amount, with values past $\pi$ wrapping around to $-\pi$. The 
open-loop walking pose that is commanded to the actuator control scheme in each 
cycle is then a simple evaluation of $f_{CPG}(\mu)$. The definition of 
$f_{CPG}$ can be manually specified and tuned for each 
robot, but the parameterized waveforms that proved to work for the \nopx have 
been used on a wide range of other platforms in the past, demonstrating its 
generality and wider applicability.

\begin{figure}
	\centering
	\includegraphics[width=0.9\linewidth]{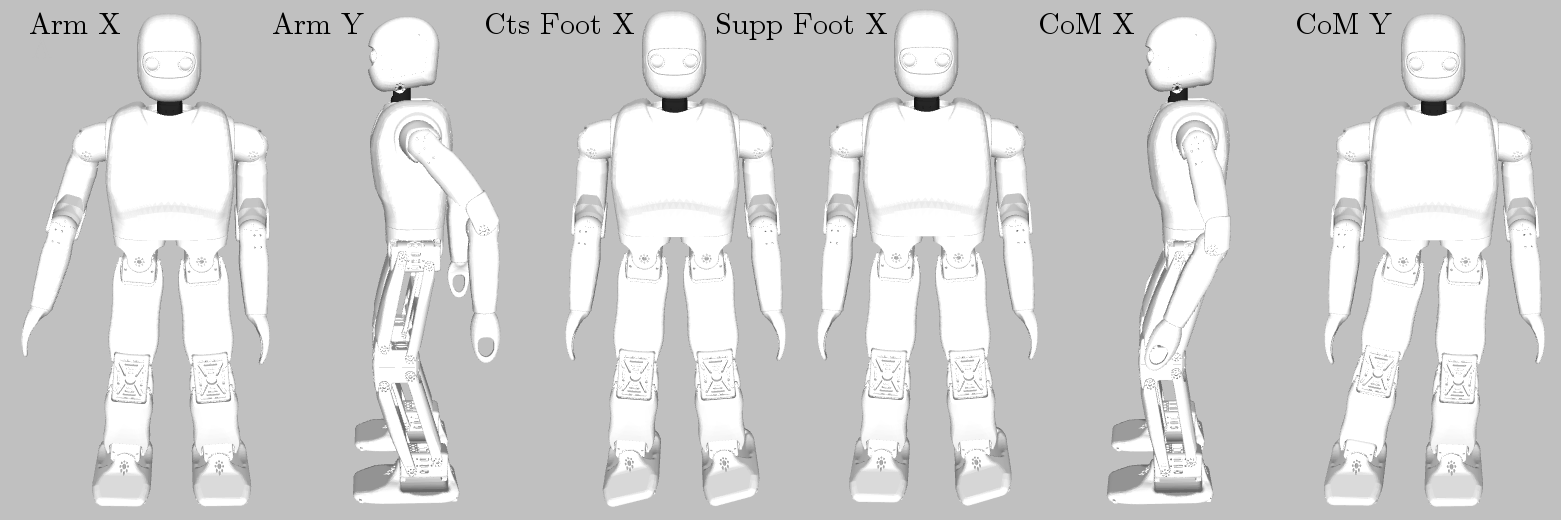}
	\caption{Example poses of the \nopx, demonstrating the various implemented 
	corrective actions aside from timing. From left to right: arm angle X, arm angle 
	Y, continuous foot angle X, support foot angle X, CoM shift X and CoM shift Y. 
	The magnitudes of the corrective actions have been exaggerated for illustrative 
	effect.}
	\figlabel{corrective_actions}
\end{figure}

In order to apply the heuristic corrective actions to the gait trajectories 
generated by $f_{CPG}$ for feedback purposes, $f_{CPG}(\mu)$ is first expressed 
in terms of the \emph{abstract pose space}, which is an alternative way of 
numerically describing a robot pose~\cite{Allgeuer2016a}. 
Essentially, the abstract space describes the cartesian pose of each leg  using three leg 
angles $\xi_{Lx}$, $\xi_{Ly}$, $\xi_{Lz}$, two foot angles relative to the torso $\xi_{Fx}$,$\xi_{Fy}$, 
and a dimensionless leg retraction parameter $\eta$ in the range $[0,1]$. Analogous parameters
exist for the arms. This representation can be ocmputed from the joint angles $q$:
\begin{equation}
\begin{aligned} 
\eta &= 1-cos(\frac{1}{2}q^{knee}_{pitch}), &\xi_{Lz} &= q^{hip}_{yaw},\\
\xi_{Ly} &= q^{hip}_{pitch}+\frac{1}{2}q^{knee}_{pitch}, &\xi_{Lx} &= q^{hip}_{roll},\\
\xi_{Fy} &= \xi_{Ly}+q^{ankle}_{pitch}+\frac{1}{2}q^{knee}_{pitch}, \qquad &\xi_{Fx} &= \xi_{Lx}+q^{ankle}_{roll},
\end{aligned}
\end{equation}
Once in the abstract space, the required corrective changes to the 
arm angles and foot angles are made, and the result is converted back to the 
joint space to give $f_{CL}(\mu)$. The corrective actions that have been 
implemented on the \nopx to augment the open-loop central pattern generated gait 
are (see \figref{corrective_actions} for example visualisations of each of these):
\begin{itemize}
\item \textbf{Arm angle:} The arm X and Y angles are adjusted to shift the 
weight of the arms, and thereby also apply reaction moments to the torso. 
\item \textbf{Continuous foot angle:} The foot angle X (i.e.\ `roll') is 
adjusted in a continuous manner for both feet to push the robot further to one 
direction laterally.
\item \textbf{Support foot angle:} The foot angle X is adjusted transiently for 
the support foot from the moment of touch down to the moment of lift-off, to 
apply an impulsive ground reaction force to the robot to bias its lateral 
balance.
\item \textbf{CoM shifting:} The inverse kinematic position of the feet relative 
to the torso are adjusted in the horizontal plane relative to the torso. As the 
CoM of the robot is assumed to be situated at a constant offset relative to the torso frame, this
results in shifting the position of the assumed CoM over the robot's support polygon. 
\item \textbf{Timing:} The rate of progression of gait phase $\mu$ is adjusted 
in order to speed up or slow down the open-loop gait, causing the next step to 
be taken earlier or later in time.
\end{itemize}
As the parallel kinematics of the \nopx restrict the ability to tilt the feet 
in the sagittal direction relative to the torso, these corrective actions
are only a subset of the corrective actions that were developed in
\cite{Allgeuer2016a}.The only corrective action that does not, like previously described, simply require 
a component to be added to $f_{CPG}(\mu)$ expressed in the abstract space is the 
timing adjustment. This corrective action works instead by modifying the 
increment that is applied to $\mu$ in each time step.

The sole source of feedback for the calculation of the activations of the 
corrective actions is the orientation of the torso of the robot. The orientation of the torso is expressed 
in terms of the fused angles rotation representation, i.e.\ the fused pitch and 
roll~\cite{AllgeuerFusedAngles}, and the deviations $d_\theta, d_\phi$ of 
these values from their nominal constant or sinusoidal waveforms. 
If for example, the robot is leaning 0.1 radians further forwards than it should 
be, then $d_\theta$, the deviation in the fused pitch, would be $+0.1$.
The corrective actions are then triggered and applied depending on their respective
gains to generate behaviour that would make the robot tilt further backwards 
again.

The most predominant corrective actions, responsible for the majority of the 
transient stability of the robot, are the arm angle and support foot angle 
actions. These are activated by constructing smoothed and deadbanded 
proportional $K_p$ and derivative $K_d$ components of $d_\theta$ and $d_\phi$, and scaling 
them by various gains on a per-component and per-action basis. While this works
well for large temporary disturbances, the consistent long-term effect of 
smaller biases cannot be effectively counteracted. To deal specifically with 
these, an integral term based on an exponentially weighted (`leaky') integrator 
(with a gain of $K_i$) is constructed as well, and used to activate the continuous 
foot angle and CoM shifting feedback mechanisms. These then make small but consistent long-term 
changes to the balance of the robot, to avoid problems like a consistent torso 
tilt in one direction, or for example also `limping' effects on one leg. 
The only remaining corrective action, the timing adjustment, is activated by 
weighting and deadbanding the smoothed fused roll deviation $d_\phi$:
and applying independent speed-up and slow-down gains to determine the 
contribution of the increment of $\mu$ that is applied in that time step. 
The slow-down gain is selected to lower the gait frequency to avoid premature stepping when the robot is tipping outwards. 
Analogically, the speed-up gain is chosen to raise the frequency and place the swing foot in time when the robot is tipping inwards.

All of the utilized heuristics are combined and applied simultaneously, 
i.e.\ superimposed on the gait trajectories in the abstract space. 
The closed-loop gait is significantly more stable than the open-loop gait, 
and allows for fast walking of the robot. The gait will work on low-cost platforms
with imprecise actuation or lacking the necessary sensors to estimate ground reaction forces, 
such as the \nopx. The limitation of the gait also stems from this, as every corrective action
requires gains tuned on a per-robot basis for them to contribute to the overall balance.
As manual tuning requires specialist knowledge on the behaviour of the corrective actions~(as per \cite{Allgeuer2016a}),
the next subsection deals with a more automated approach to tuning the gains. 
Numerical and other qualitative results of the walking can be found in \secref{walking_stability}.

\subsubsection{Bayesian gait optimization}

The application of fused angle feedback mechanisms introduces additional 
parameters, which highly influence the performance of the gait. 
The tuning process of these parameters can only be performed by experts, as it requires 
substantial knowledge over the gait and the effect of some parameters can not be 
easily observed. 
Furthermore, every individual robot can have different optimal 
parameters, not only due to variations in the robot model, but also due to 
mechanical inaccuracies.
In other words, the gait of each robot needs to be parametrized individually which creates overhead for the robot operator. 

To address these drawbacks, we introduce an approach to optimize 
the parameters of the fused angle feedback mechanisms. This method was already 
successfully applied to the \iguhop~\cite{Rodriguez2018Combining} and the \nopx~\cite{ficht2018nimbro}. 
In contrast to traditional methods, our approach not 
only utilizes information from the real system, but also incorporates knowledge 
from a simulator. In this manner, we reduce the hardware wear-off and accelerate the 
optimization process by evaluating parameters faster than real-time.

To further reduce the load on the real system, our approach uses 
sample-efficient \textit{Bayesian Optimization}, allowing efficient trade-off 
between exploration and exploitation. This trade-off can be controlled by the 
choice of an adequate \textit{Kernel function} $k$ that parametrizes a 
\textit{Gaussian Process} (GP). In 
this approach, we utilize a composite kernel, consisting of the two terms 
$k_{sim}$, resembling the performance in simulation, and $k_\epsilon$, which 
relates to the estimated error between the simulation and the real system. 
By introducing an augmented parameter vector $\mathbf{a_i}=(\mathbf{x_i},\delta_i)$, where $\delta$ symbolizes whether an experiment has been performed in simulation 
or on the real system, the kernel function can de expressed as:
\begin{equation}\label{eq:compositeKernel} 
k(\mathbf{a_i},\mathbf{a_j})=k_{sim}(\mathbf{x_i},\mathbf{x_j} 
)+k_\delta(\delta_i,\delta_j)k_{\epsilon}(\mathbf{x_i}, \mathbf{x_j})\,
\end{equation} 

where $k_{\delta}=1$ if both experiments have been performed in the real world 
and $k_{\delta}=0$ otherwise. Due to $k_\epsilon$, it is 
possible to model a highly complex, non-linear mapping between the 
simulation and the real-world. The selection of the next evaluation point is 
performed by maximizing the expected change in \textit{Entropy}, which contributes 
to increase the sample-efficiency of our approach~\cite{hennig2012}. Furthermore, to limit 
hardware use, we bias the selection of query points by a weight factor towards 
the simulation. For the two kernel functions, we choose the Rational-Quadratic 
Kernel, as it has shown to be suitable in our previous work~\cite{Rodriguez2018Combining}. 

We propose a cost function including the proportional fused feedback $e_P$, 
which is based on the fused angle deviations and thus measures the robots 
stability. Also, we account for the sagittal ($\alpha$) and lateral ($\beta$) 
plane separately to reduce the impact of noise. Additionally, we prefer 
parameters with low torques, which is formulated by penalizing high gains 
through a regularization term $\nu(\mathbf{x})$. This results in the cost 
functions:
\begin{equation}\label{eq:finala}
J_{\alpha}(\mathbf{x}) = \int_0^T{\Vert e_{P\alpha}(\mathbf{x})\Vert_1}dt + 
\nu(\mathbf{x})
\end{equation}
\begin{equation}\label{eq:finalb}
J_{\beta}(\mathbf{x}) = \int_0^T{\Vert e_{P\beta}(\mathbf{x})\Vert_1}dt + 
\nu(\mathbf{x})\,,
\end{equation}
where the proportional fused feedback $e_P$ is modeled as a function of the 
gait parametrization $\mathbf{x}$.

\begin{figure}
	\centering
	\resizebox{\linewidth}{!}{
\begingroup%
  \makeatletter%
  \providecommand\color[2][]{%
    \errmessage{(Inkscape) Color is used for the text in Inkscape, but the package 'color.sty' is not loaded}%
    \renewcommand\color[2][]{}%
  }%
  \providecommand\transparent[1]{%
    \errmessage{(Inkscape) Transparency is used (non-zero) for the text in Inkscape, but the package 'transparent.sty' is not loaded}%
    \renewcommand\transparent[1]{}%
  }%
  \providecommand\rotatebox[2]{#2}%
  \newcommand*\fsize{\dimexpr\f@size pt\relax}%
  \newcommand*\lineheight[1]{\fontsize{\fsize}{#1\fsize}\selectfont}%
  \ifx\svgwidth\undefined%
    \setlength{\unitlength}{532.06677246bp}%
    \ifx\svgscale\undefined%
      \relax%
    \else%
      \setlength{\unitlength}{\unitlength * \real{\svgscale}}%
    \fi%
  \else%
    \setlength{\unitlength}{\svgwidth}%
  \fi%
  \global\let\svgwidth\undefined%
  \global\let\svgscale\undefined%
  \makeatother%
  \begin{picture}(1,0.36082758)%
    \lineheight{1}%
    \setlength\tabcolsep{0pt}%
    \put(0,0){\includegraphics[width=\unitlength,page=1]{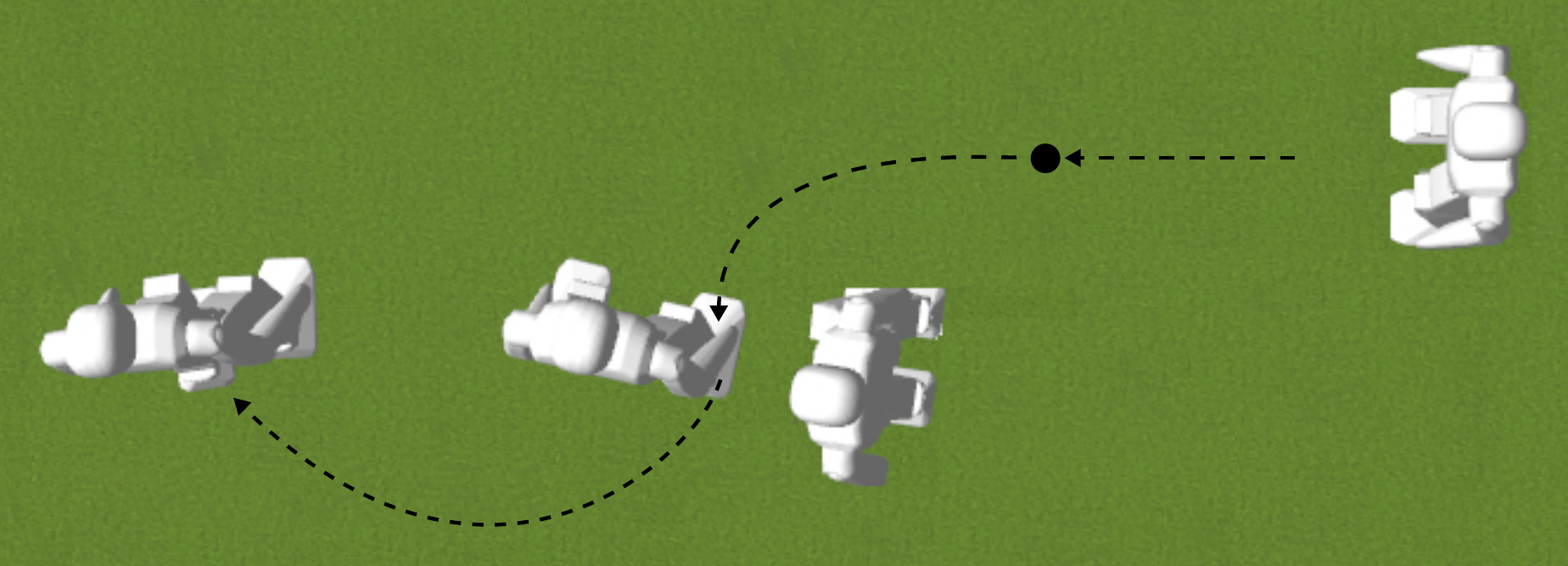}}%
    \put(0.71511949,0.2675224){\color[rgb]{0.97647059,0.97647059,0.97647059}\makebox(0,0)[lt]{\lineheight{0}\smash{\begin{tabular}[t]{l}Forward\end{tabular}}}}%
    \put(0.56725391,0.26675497){\color[rgb]{0.97647059,0.97647059,0.97647059}\makebox(0,0)[lt]{\lineheight{0}\smash{\begin{tabular}[t]{l}Left\end{tabular}}}}%
    \put(0.27913216,0.03520509){\color[rgb]{0.97647059,0.97647059,0.97647059}\makebox(0,0)[lt]{\lineheight{0}\smash{\begin{tabular}[t]{l}Right\\\end{tabular}}}}%
    \put(0.24887281,0.19759073){\color[rgb]{0.97647059,0.97647059,0.97647059}\makebox(0,0)[lt]{\lineheight{0}\smash{\begin{tabular}[t]{l}Sideways\end{tabular}}}}%
    \put(0.38306648,0.22615857){\color[rgb]{0.97647059,0.97647059,0.97647059}\makebox(0,0)[lt]{\lineheight{0}\smash{\begin{tabular}[t]{l}Rotate\end{tabular}}}}%
    \put(0.49884144,0.19608715){\color[rgb]{0.97647059,0.97647059,0.97647059}\makebox(0,0)[lt]{\lineheight{0}\smash{\begin{tabular}[t]{l}Backward\end{tabular}}}}%
    \put(0,0){\includegraphics[width=\unitlength,page=2]{figures/GPO/trial_summary.pdf}}%
  \end{picture}%
\endgroup%
}
	\caption{Test sequence used in the Bayesian Optimization. From the start pose (right top),
		the robot walks forwards, omni-directionally turns left and right followed by sideway 
		locomotion. Finally, the robot turns in place and walks backwards.}
	\figlabel{gpo_test_seq}
\end{figure}

The cost functions are evaluated through a complex, predefined test sequence, 
featuring forward, side-ways and backward gait commands, as shown in 
\figref{gpo_test_seq}. In simulation, we average the cost of $N=4$ runs, 
yielding more reproducible results.

For the \nopx, we optimized the P and D parameters for the 
Arm Angle corrective action in the sagittal direction. Nonetheless, the proposed 
method can be applied on all other parameters of the fused feedback mechanisms. 
To reduce hardware use, we limit the permitted number of real-world experiments 
to 15, which is comparable to the time an expert needs to hand-tune the 
parameters. This limit was reached after a total number of 161 iterations, thus 
including 146 evaluations of the gait inside the simulator. The performance of 
the resulting parameters were assessed in 5 test sequences and compared to the 
previous manual-tuned parameters. The optimized parametrization results in a 18\% 
improvement of 
the fused angle deviation and also shows a qualitatively convincing gait. 
Furthermore, the optimized gait is more robust to disturbance, allowing the 
robot to walk over obstacles (\figref{gpo_obstacle}).

\begin{figure}
	\centering
	\includegraphics[width=0.19\linewidth]{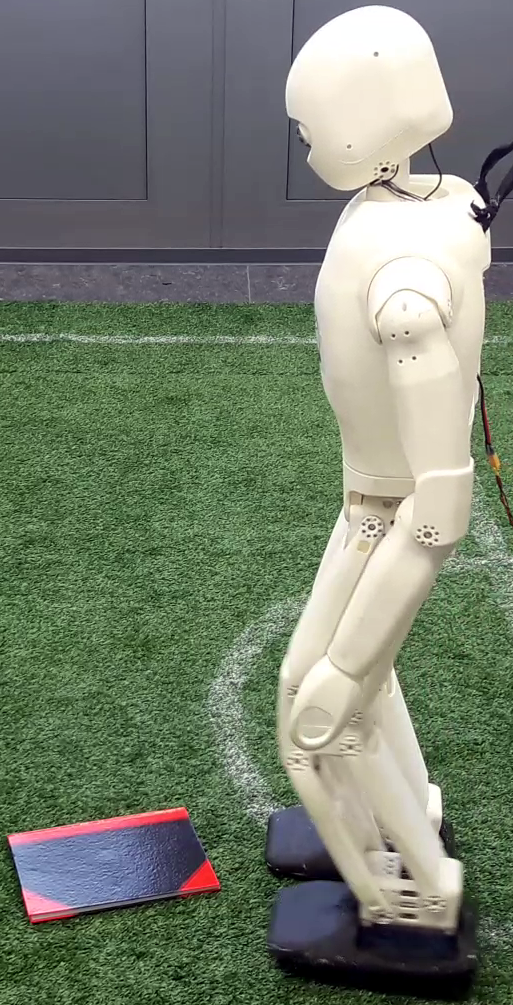}
	\includegraphics[width=0.19\linewidth]{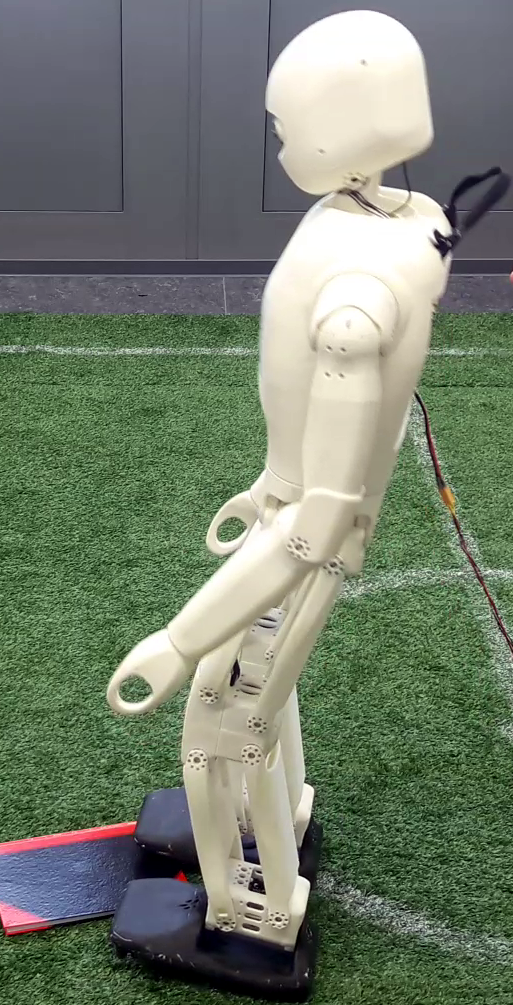}
	\includegraphics[width=0.19\linewidth]{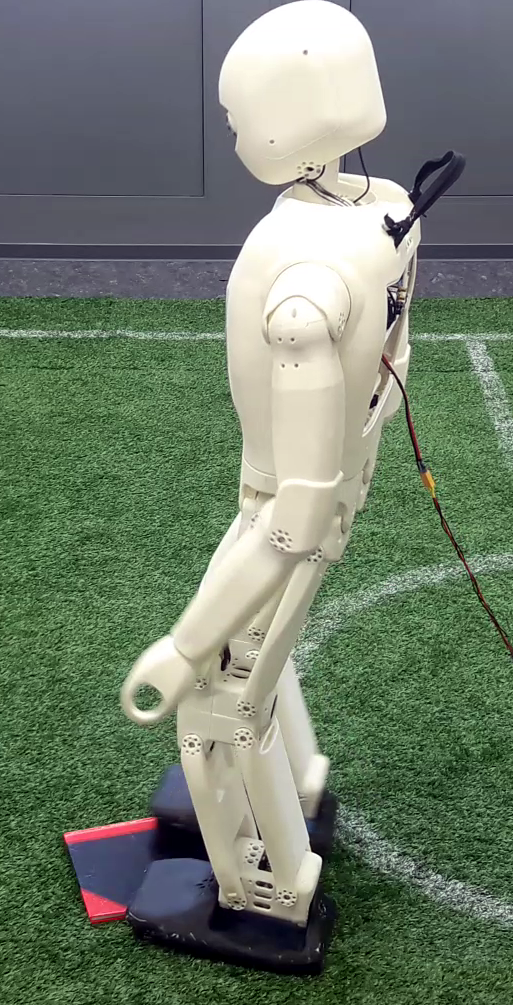}
	\includegraphics[width=0.19\linewidth]{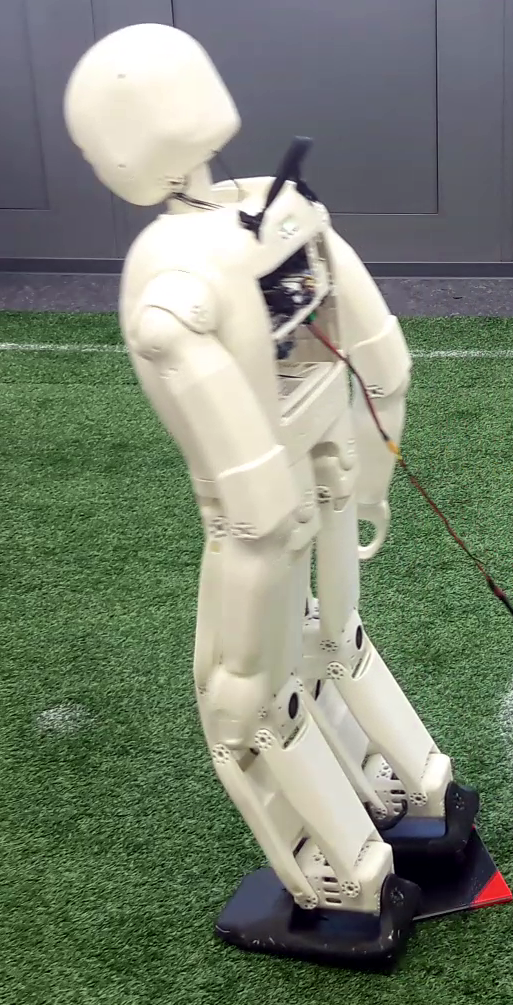}
	\includegraphics[width=0.19\linewidth]{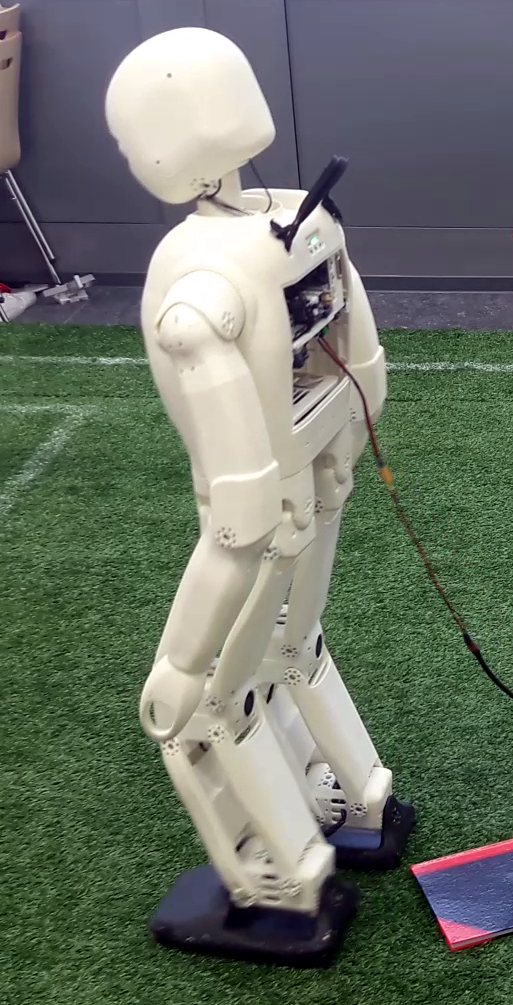}
	\caption{The \nopx successfully walks over an obstacle using optimized gait parameters. }
	\figlabel{gpo_obstacle}
\end{figure}

\subsection{Visual perception}
\label{vision_sec}

The visual perception system of the \nopx is able to recognize soccer related objects including a soccer ball, 
field boundaries, line segments and goal posts. Apart from this, other robots and non-soccer objects such as QR codes, 
human faces and skeletons can be detected through the effective usage of texture, shape, brightness and color information.

\subsubsection{Soccer Object Detection}
Our deep-learning based visual perception system can work with different brightnesses, 
viewing angles, and even lens distortions. To achieve this, we utilize a deep convolutional neural network followed by post-processing
we managed to outperform our previous approach to soccer vision~\cite{farazi2015} and include tracking and identification of our robots~\cite{Farazi2017b}.

The developed object detection pipeline uses an encoder-decoder architecture similar to pixel-wise segmentation 
models like SegNet~\cite{badrinarayanan2015segnet}, and U-Net~\cite{ronneberger2015u}. Due to computational limitations 
and the necessity of real-time perception, we have made several adaptations. One of these was using a `shorter' decoder than encoder in the network. 
With this design choice, we have minimized the number of parameters for the cost of losing fine-grained spatial information. 
To recover some of it, we use a method for finding the subpixel centroid in the post-processing steps. To minimize the effort to annotate 
data and greatly save on training time we utilized transfer-learning. A pre-trained ResNet-18 which is the lightest 
version in the pre-trained ResNet family is chosen as the encoder. Since ResNet was originally designed for classification 
tasks, we removed the Global Average Pooling (GAP) and the fully connected layers in the model. Transpose-convolutional layers are used
to upsample the representations. We took advantage of the lateral connections between 
the encoder and decoder parts, similar to the U-Net model, with the aim of preserving the high-resolution spatial information 
in the decoder part. The proposed visual perception architecture, which in total has 23 convolutional layers, is illustrated in \figref{net}.
\begin{figure}[t]
	\centering
	\includegraphics[width=0.55\linewidth,angle=90,origin=c]{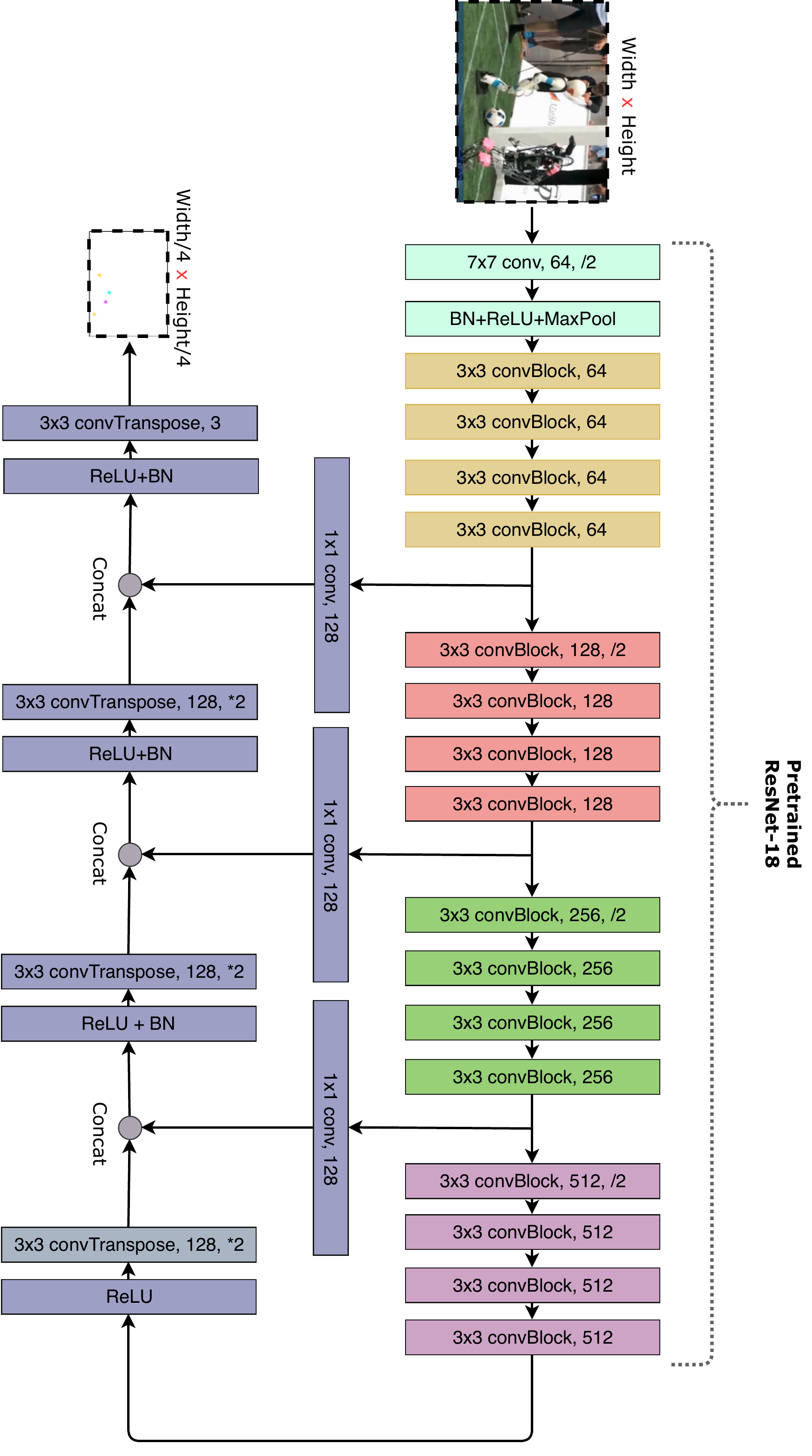}
	\vspace{-15ex}
	\caption{Visual perception architecture. Similar to original ResNet architecture, each convBlock consists of two convolutional layers followed by batch-norm and ReLU activations. Note that for simplicity, residual connections in ResNet are not depicted.}
	\figlabel{net}
	\vspace{-1ex}
\end{figure}

The following soccer related object classes were detected using the presented network: goal posts, soccer balls, and robots. For our soccer behavior, 
we only need to perceive predefined center locations of the aforementioned objects. Similar to SweatyNet~\cite{schnekenburger2017detection}, 
instead of a full segmentation loss, we used the mean squared error on the desired output. The target is constructed by Gaussian blobs around the 
ball center and bottom-middle points of the goal posts and robots.

Despite using the Adam optimizer, which has an adaptable per-parameter learning rate, finding a decent learning rate is still a challenging 
prerequisite for training. To determine an optimal learning rate, we followed the approach presented by Smith et al.~\cite{smith2017cyclical}. 
We have also tested the recently proposed AMSGrad optimizer \cite{2018on}, but did not see any benefit in our application over the employed Adam.

We used progressive image resizing that uses small pictures at the beginning of training, gradually increasing the dimensions as training 
progresses, a method inspired by Brock et al.~\cite{brock2017freezeout} and by Yosinski et al.~\cite{yosinski2014transferable}. In early 
iterations, the inaccurate randomly initialized model will make fast progress by learning from large batches of small pictures. Within the 
initial fifty epochs, we used downsampled training images whereas the weights on the encoder part are frozen. Throughout the following fifty 
epochs, all parts of the models are jointly trained. In the last fifty epochs, full-sized pictures are used to learn fine-grained details. 
A lower learning rate is employed for the encoder part, with the reasoning that the pre-trained model requires less training. With the above described method, 
the entire training process with around 3000 samples takes less than forty minutes on a single Titan Black GPU with \SI{6}{GB} of memory. 
Four examples from the test set are pictured in \figref{vision_output}. Some portions of the used dataset were taken from the ImageTagger 
library~\cite{imagetagger2018}, that have annotated samples from different angles, cameras, and brightness.
We extract the object coordinates by post-processing the blob-shaped network outputs. We apply morphological erosion and dilation to eliminate 
negligible responses on the thresholded output channels. Finally, the object center coordinates are computed. The output of the network is of 
lower resolution and has less spatial information than the input image. To account for this effect, we calculate sub-pixel level coordinates 
based on the center of mass of a detected contour. To find the contours, we use the connected component analysis~\cite{suzuki1985topological} 
on each of the output channels. 
\begin{figure*}[tb]
	\parbox{\linewidth}{
	\centering
	\includegraphics[width=0.99\linewidth]{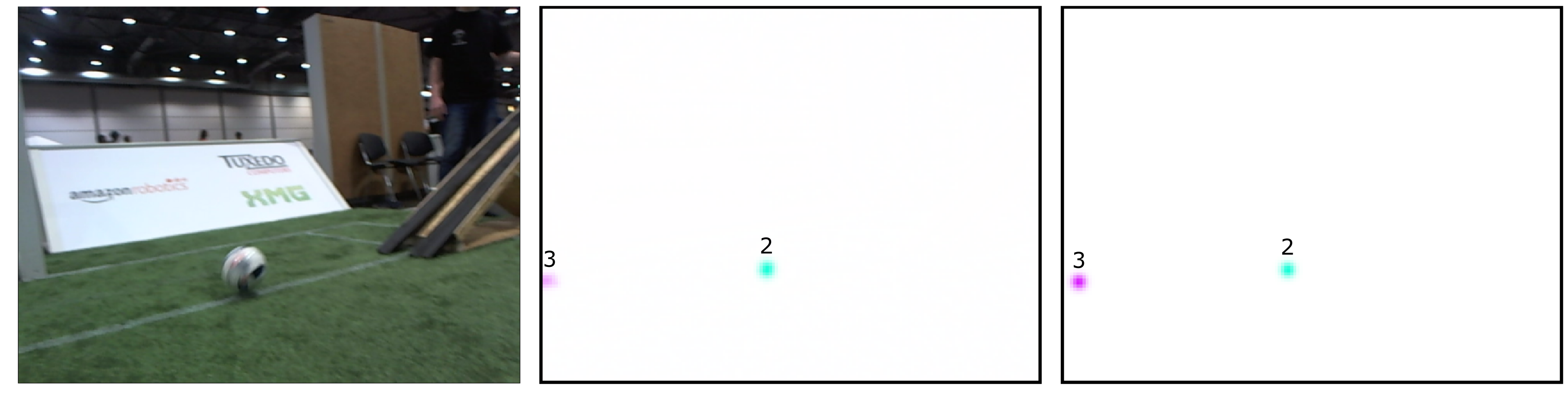}
	\centering
	\includegraphics[width=0.99\linewidth]{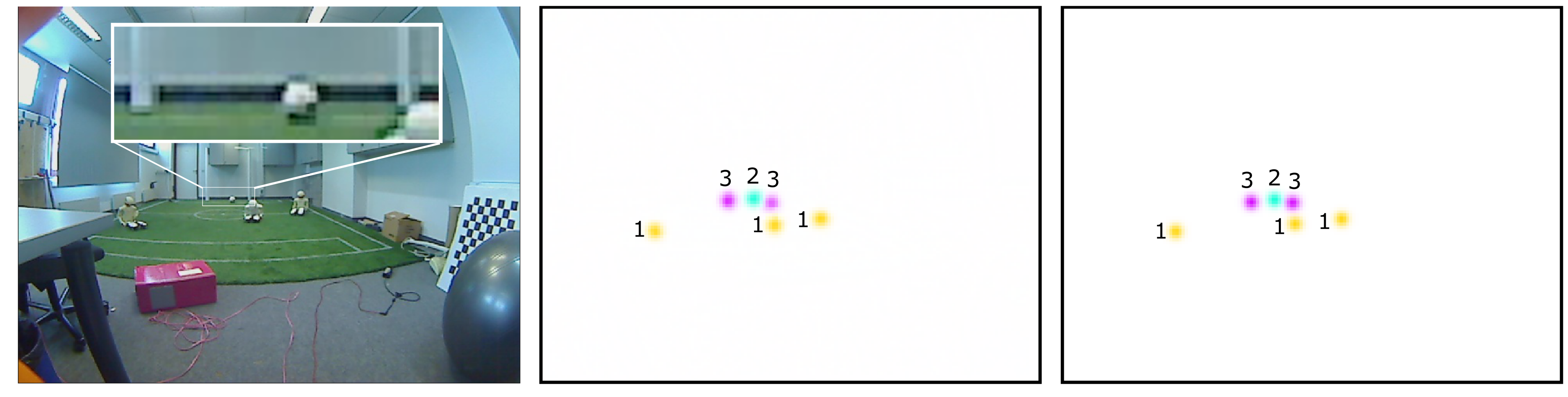}
	\centering
	\includegraphics[width=0.99\linewidth]{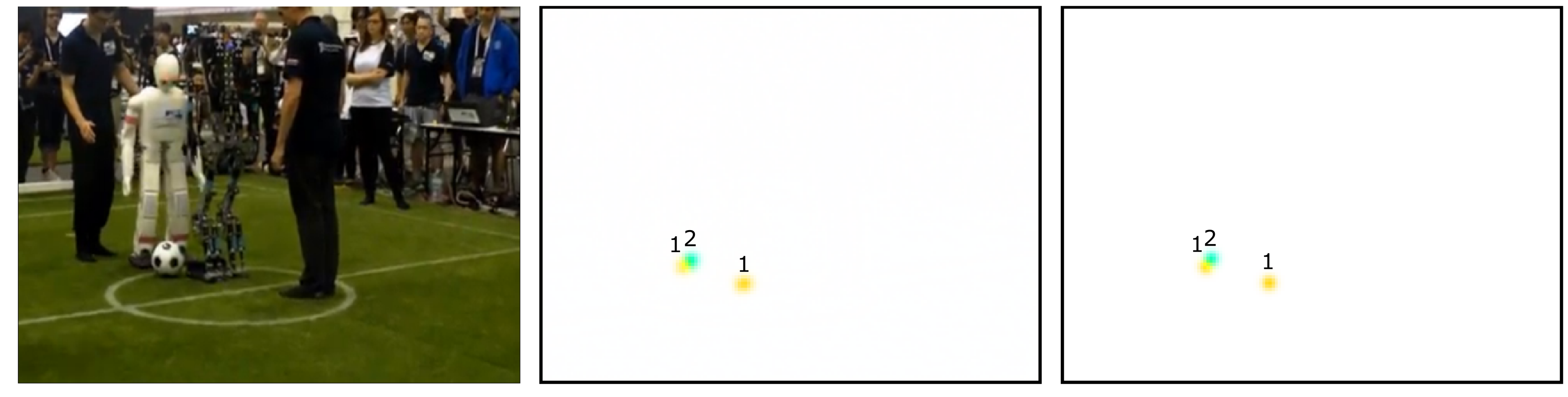}
	\centering
	\includegraphics[width=0.99\linewidth]{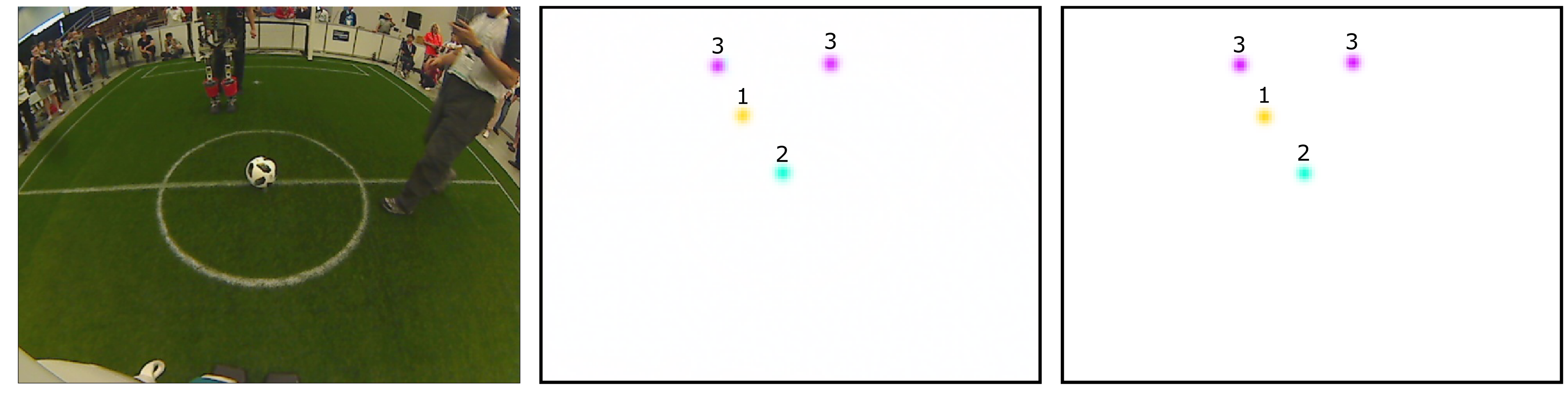}
}
	\caption{Object detection. Left column: A captured image from the robot in the test set. Middle column: The output of the network with robots (1), ball (2), and goal posts (3) annotated. Right column: Ground truth. Note that we can detect a very far ball (\SI{7}{m} away) which is only a few pixels in size.}
	\figlabel{vision_output}
	\vspace{-3ex}
\end{figure*}

After detecting soccer-related objects, we filter them and project each object location into egocentric world coordinates. These coordinates then are further processed in the behavior node of our ROS-based 
open-source software for decision making. 

Using the mentioned convolutional network, we can detect objects which are up to 8 meters away with excellent performance. 
We evaluated our visual perception pipeline in Sec.~\ref{evaluation_perception}. For field and line detections, we are still using non-deep learning approaches~\cite{farazi2015}. 
The complete perception pipeline including a forward-pass of the network takes approximately \SI{20}{ms} on the robot hardware.

\subsubsection{Human perception}

Detecting humans is a separate function, which is essential for human-robot interaction. For real-time face detection, we use the deep cascaded multi-task framework~\cite{zhang2016joint}. 
The cascaded framework includes three-stage multi-task deep convolutional networks. Firstly, an image pyramid is built using different scales, after which candidate windows are produced using a fast Proposal Network (P-Net). 
Then, generated candidates are further refined through a combination of non-maximum suppression and a Refinement Network (R-Net). 
In the third stage, the Output Network (O-Net) produces the final bounding box and facial landmark positions. An example result of the network output is shown on the left of \figref{faces}.
From there, the robot can track and follow the position of the biggest bounding box with the highest confidence value. A complete pass of the face detection and tracking pipelines on the \nopx take approximately \SI{50}{ms}.

Estimating 2D human poses and reacting to specific gestures is also possible. For the estimation part, the Part Affinity Fields method~\cite{cao2017realtime} is used. 
This approach uses a non-parametric representation, called \textit{Part Affinity Fields}, to learn to associate different body parts with individuals in the image. 
After detecting said body parts, a greedy bottom-up parsing step connects the various identified body parts to form the skeleton. 
An example result is illustrated on the right of \figref{faces}. Detecting a single skeleton using the current robot hardware can take up to about \SI{100}{ms}.

\begin{figure}[t]
	\centering
	\includegraphics[width=0.7\linewidth]{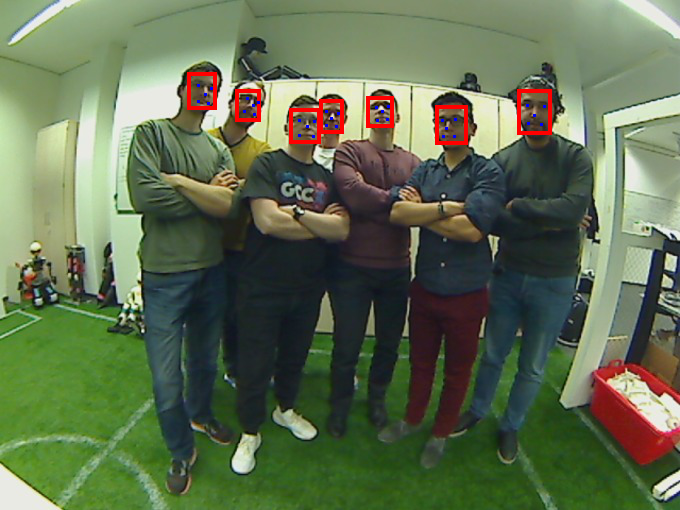}\hspace{0.1cm}\includegraphics[width=0.291\linewidth]{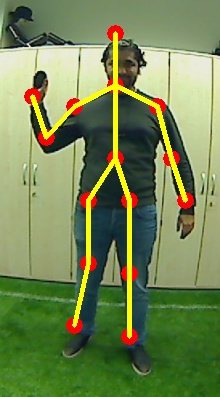}

	\caption{Left: Face landmarks detection results on a picture of team NimbRo. Right: Human pose estimation result. }
	\figlabel{faces}
	\vspace{-1ex}
\end{figure}

\subsubsection{General object detection}

Detecting non-soccer related objects is done with a general purpose pre-trained object detection method called YOLOv3~\cite{redmon2018yolov3}.
In contrast to classifier-based systems like R-CNN~\cite{Girshick2014RichFH}, which apply a classification model to different locations and scales, 
YOLO applies the model to the entire image at once. The network outputs bounding boxes with corresponding probabilities on the detected objects. 
By thresholding the probability results we discard the false positives. This method is a great choice for real-time applications, 
and it is approximately a thousand times faster than R-CNN and a hundred times faster than Fast R-CNN. 
Depending on the use case, we use different architectures. One of these is YOLOv3-tiny, which can run with \SI{100}{FPS}. It is suitable for coarser detection of big and unoccluded objects. 
The more precise YOLOv3-608 performs exceptionally well with objects of any size. It runs at \SI{7}{FPS}, which can still be considered a good result, given the functionality it provides. 
This difference in performance is the result of the different network architectures employed. The tiny version has only 13 convolutional layers, while the bigger model utilizes 75. 
An example result produced by these architectures is shown in \figref{yolo}. As these objects can be detected using the onboard computer, no additional 
server connection is required to offload the computations. Detections provided by these networks could be then further used for motion planning and grasping, given that the robot would be equipped with a gripper.

\begin{figure}[t]
	\centering
	\includegraphics[width=0.49\linewidth]{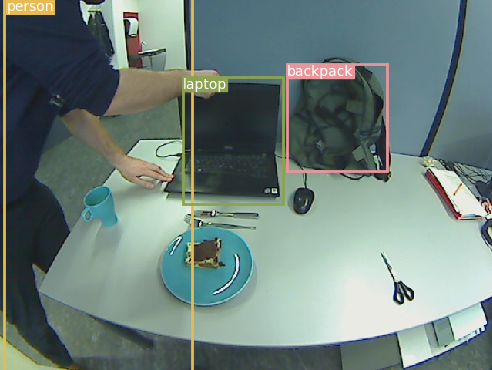}\hspace{0.1cm}\includegraphics[width=0.49\linewidth]{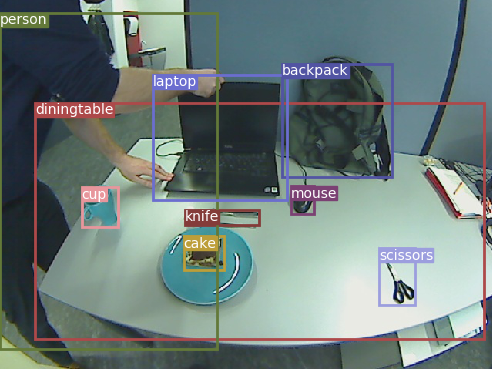}
	
	\caption{Left: A sample result of object detection using YOLOv3-tiny. Right: A sample result of object detection using YOLOv3-608. }
	\figlabel{yolo}
	\vspace{-1ex}
\end{figure}

\subsection{Simulation} 
\begin{figure}
	\centering
	\includegraphics[height=0.4\linewidth]{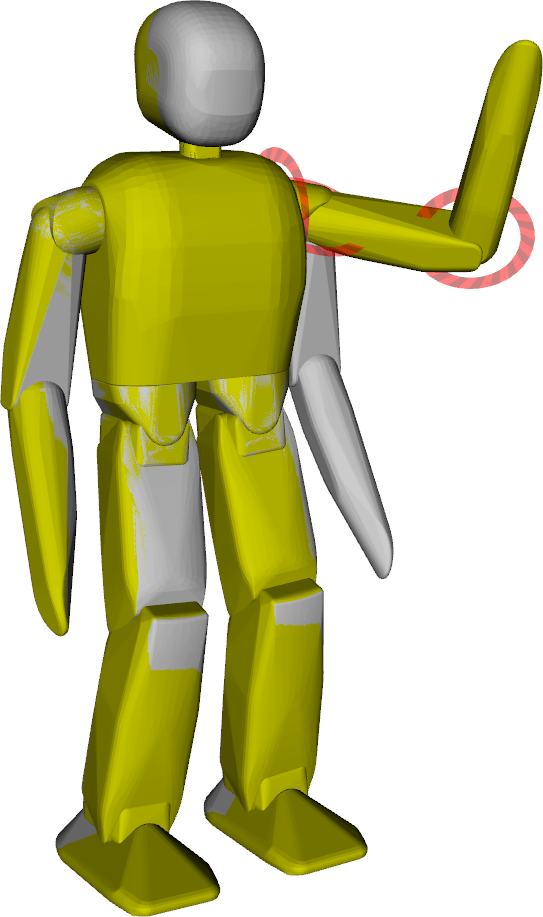}\hfil
	\includegraphics[height=0.4\linewidth]{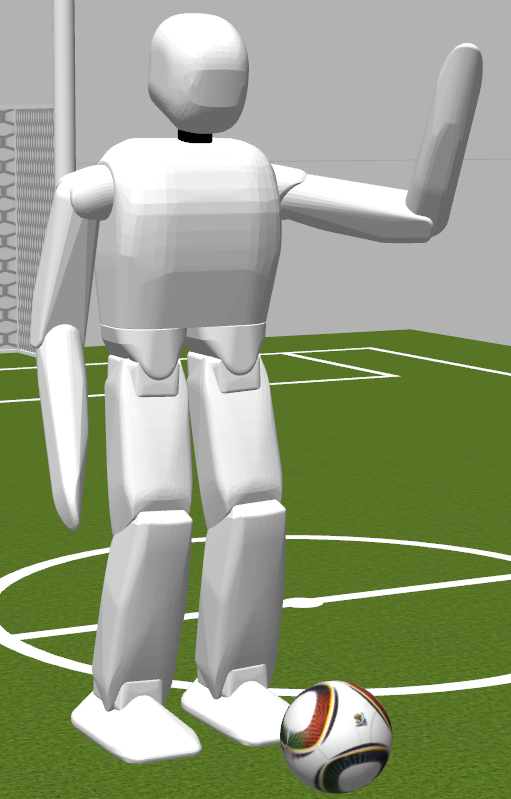}\hfil
	\includegraphics[height=0.4\linewidth]{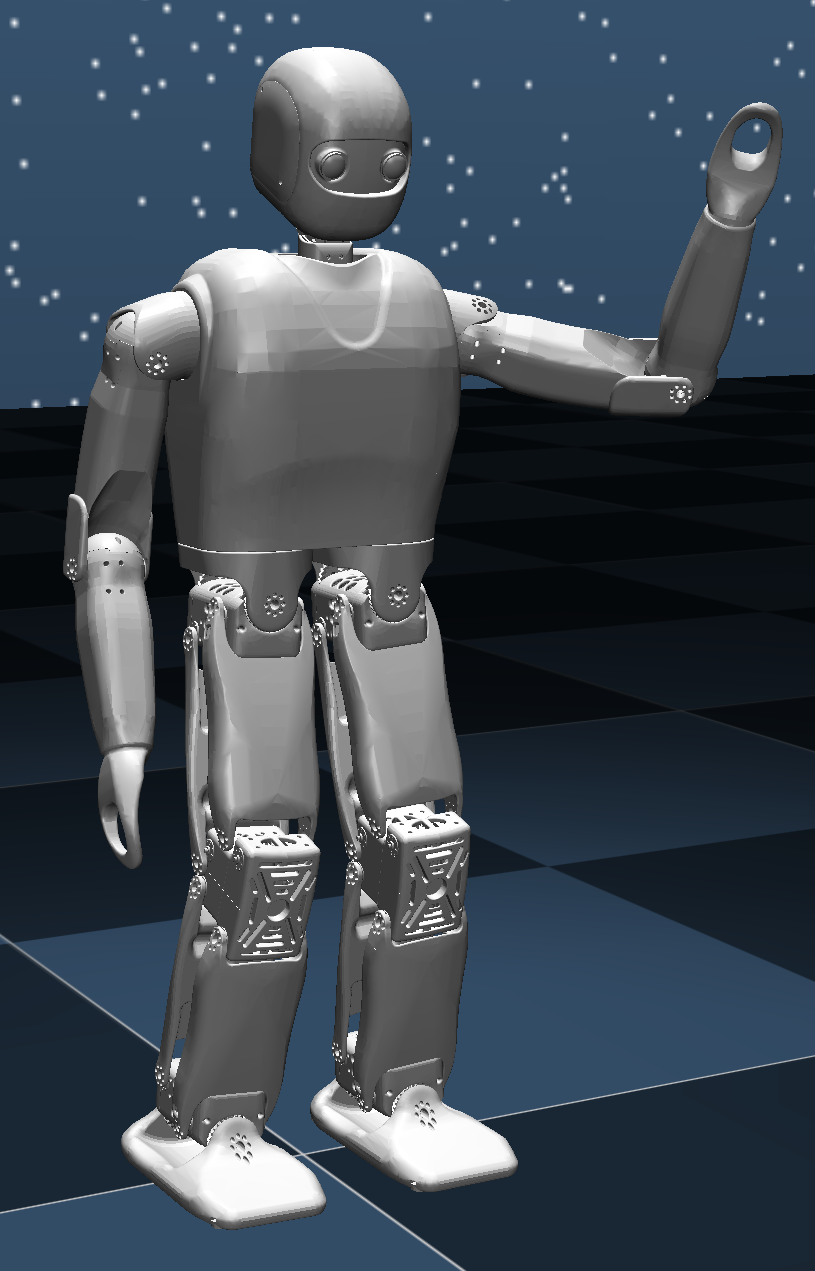}
	\caption{Motion modules used with different simulators. 
		On the left, a desired joint configuration is designed using a keyframe editor.
		The motion is executed in two different simulators: Gazebo (middle) and MuJoCo (right).
		The motion generation, interpolation and control is the same for both simulators, 
		only the interface to each simulator is different.}
	\label{fig:sims}
\end{figure}

One of the benefits of our highly modular framework is the integration of different simulators without requiring changes in the motion generation modules.
For instance, the Gazebo simulator\footnote{~Gazebo: \url{http://gazebosim.org}} is the preferred option when the use sensors such as RGB-D cameras or laser scanners are required.
On the other hand, the MuJoCo simulator\footnote{~MuJoCo: \url{www.mujoco.org}} is a more suitable alternative for learning related tasks, 
specially, for Deep Reinforcement Learning (DRL) approaches due to its stability, high accuracy and fast evaluation time\cite{erez2015}.
For each simulator, we only need an interface that defines how to write and to read the data from the simulators,
other components, including: control, motion generation and planning, are employed without any modification.
In a similar manner, the execution of motions with the real robot only requires a corresponding hardware interface.
Figure~\ref{fig:sims} displays an operator interface that generates a motion executed in two different simulators.

\begin{figure}
	\centering
	\includegraphics[height=0.35\linewidth]{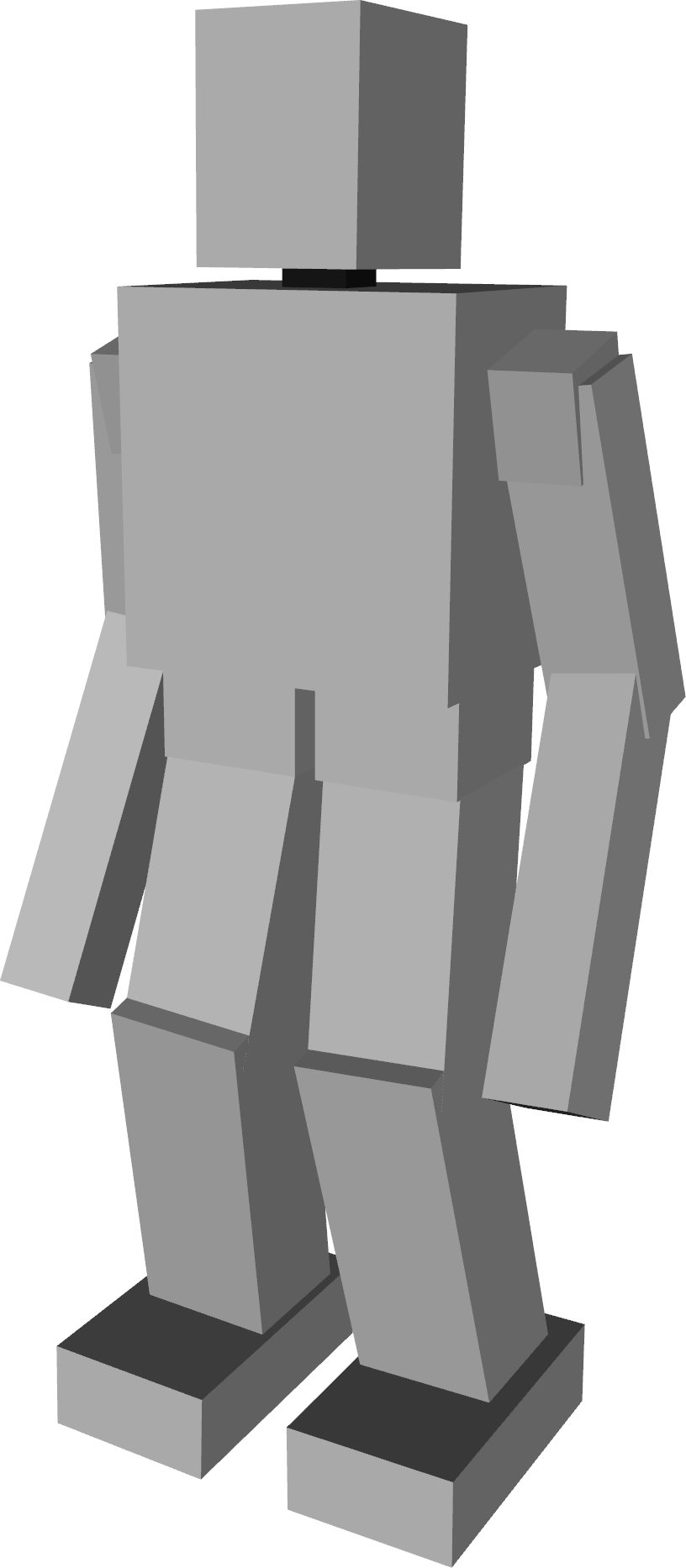}\hfil
	\includegraphics[height=0.35\linewidth]{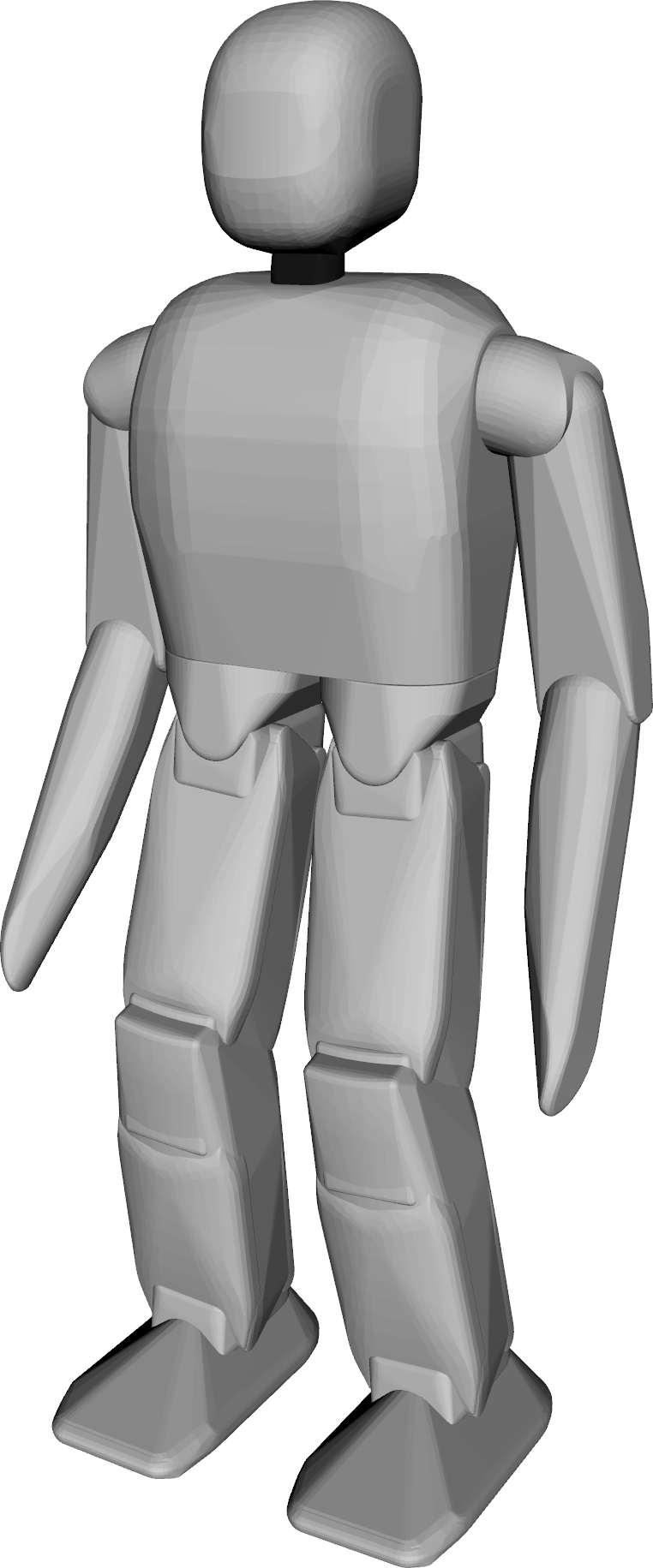}\hfil
	\includegraphics[height=0.35\linewidth]{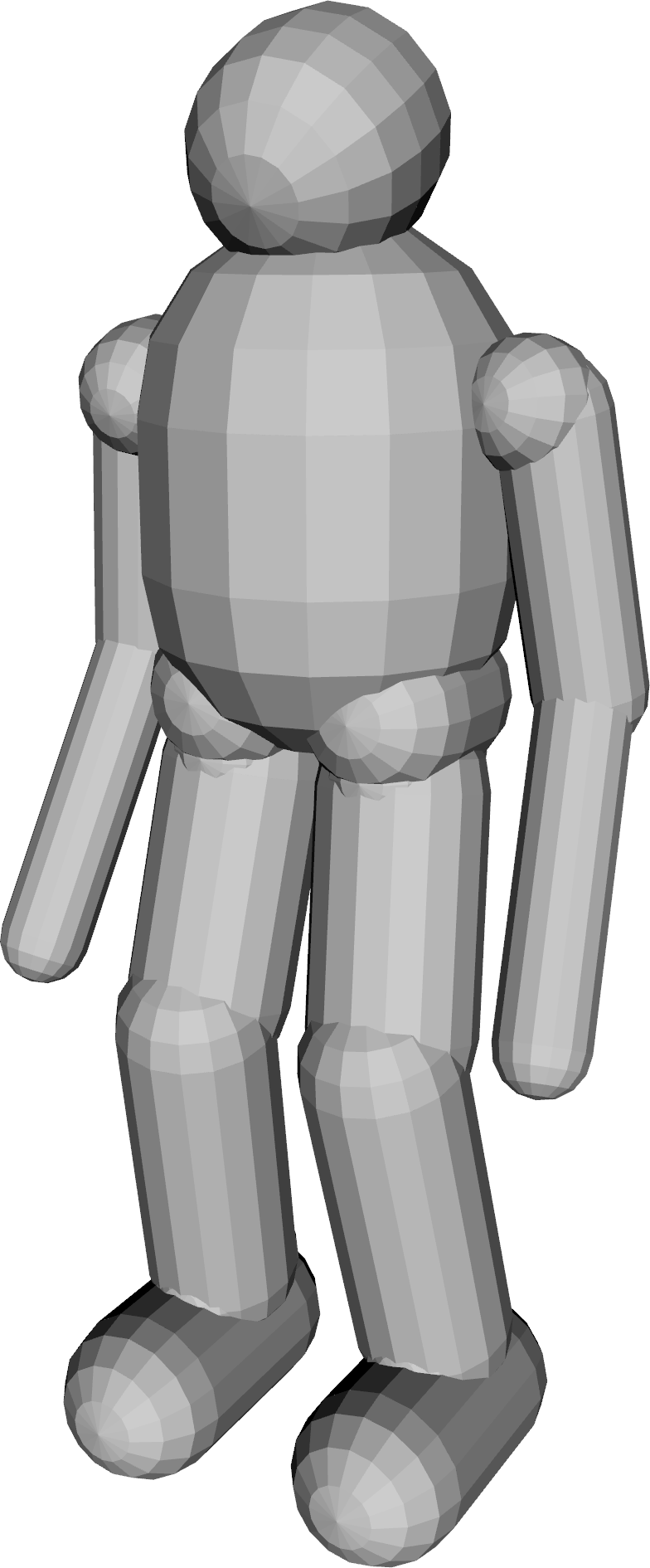}\hfil
	\includegraphics[height=0.35\linewidth]{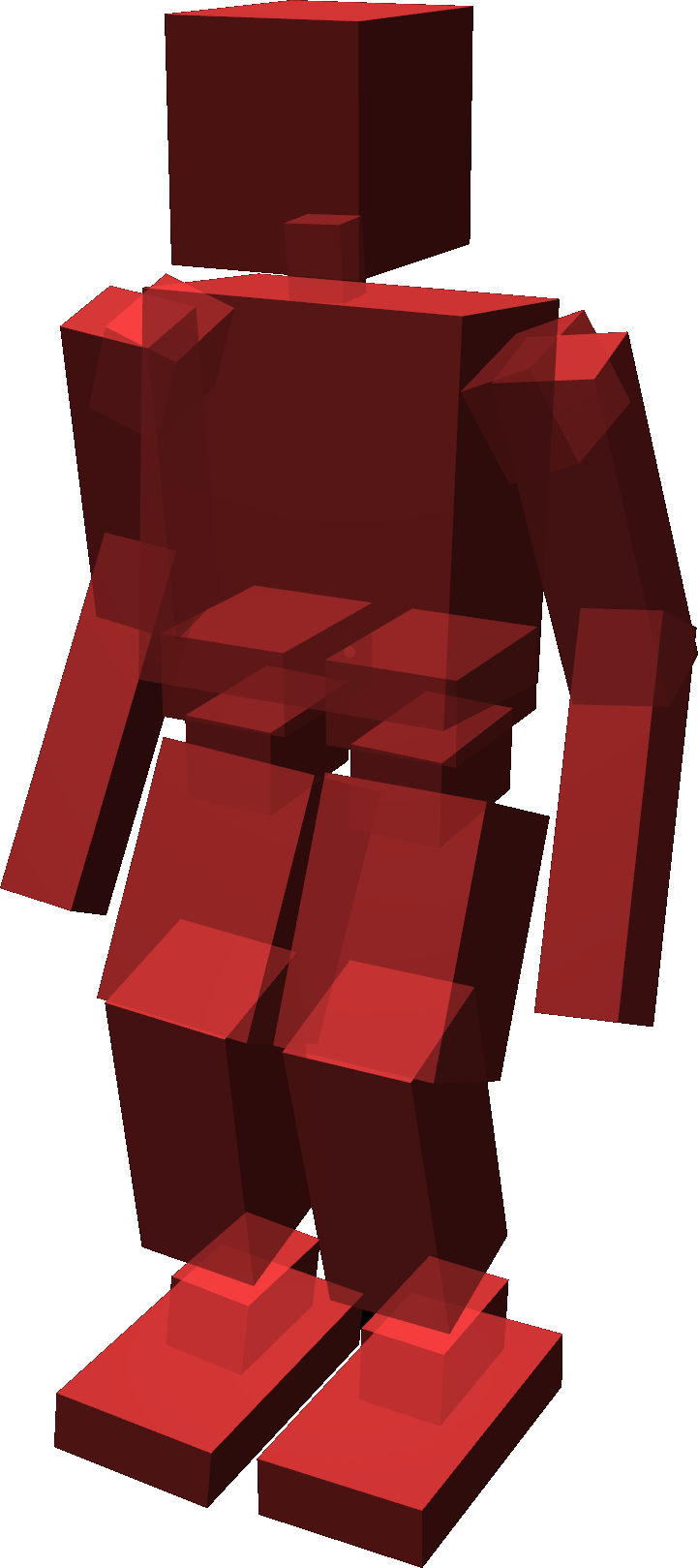}
	\caption{For speeding up the collision checking in simulation,
		 different collision models are generated, from left to right: optimal oriented boxes, convex hulls and oriented capsules. On the rightmost the inertia tensors are visualized.}
	\label{fig:capsule_model}
\end{figure}

The simulation of humanoid robots is computationally expensive due to the underactuation of the robot base and the repetitive contacts with the floor.
In order to reduce the execution time of the simulation, 
the body geometries of the robot are approximated in different levels.
The first approximation converts the model meshes into convex hulls.
This is normally employed when contacts on the trunk and limbs are expected and need to be analyzed precisely. 
In the second level, all model meshes are converted either into oriented capsules or oriented bounding boxes.
Collision checking between capsules is faster than boxes, but the use of boxes sometimes is necessary, 
for instance, for modeling the feet.
Simulations with these geometric primitives are performed when the real time factor needs to be at its maximum,
e.g., for learning approaches.
Additionally, simulators require the inertia tensors of each body for calculating the dynamics, 
these values are extracted directly from CAD files.
We developed and open sourced an automatic URDF file converter that calculates the optimal oriented boxes and capsules given the model meshes~\footnote{~\url{https://github.com/AIS-Bonn/primitive\_fitter}}.
The converter is based on the roboptim library\cite{Khoury} and the ApproxMVBB library\cite{barequet2001}.
If the CAD files of the hardware are not available, 
an automatic inertia approximator is also implemented based on the geometry of the mesh and the mass of the body.
The NimbRo OP2X robot with different collision models and inertia visualization is shown in Figure~\ref{fig:capsule_model}.

\begin{figure}
	\begin{center}
		\includegraphics[width=0.13\linewidth, trim={600pt 0 600pt 0}, clip]{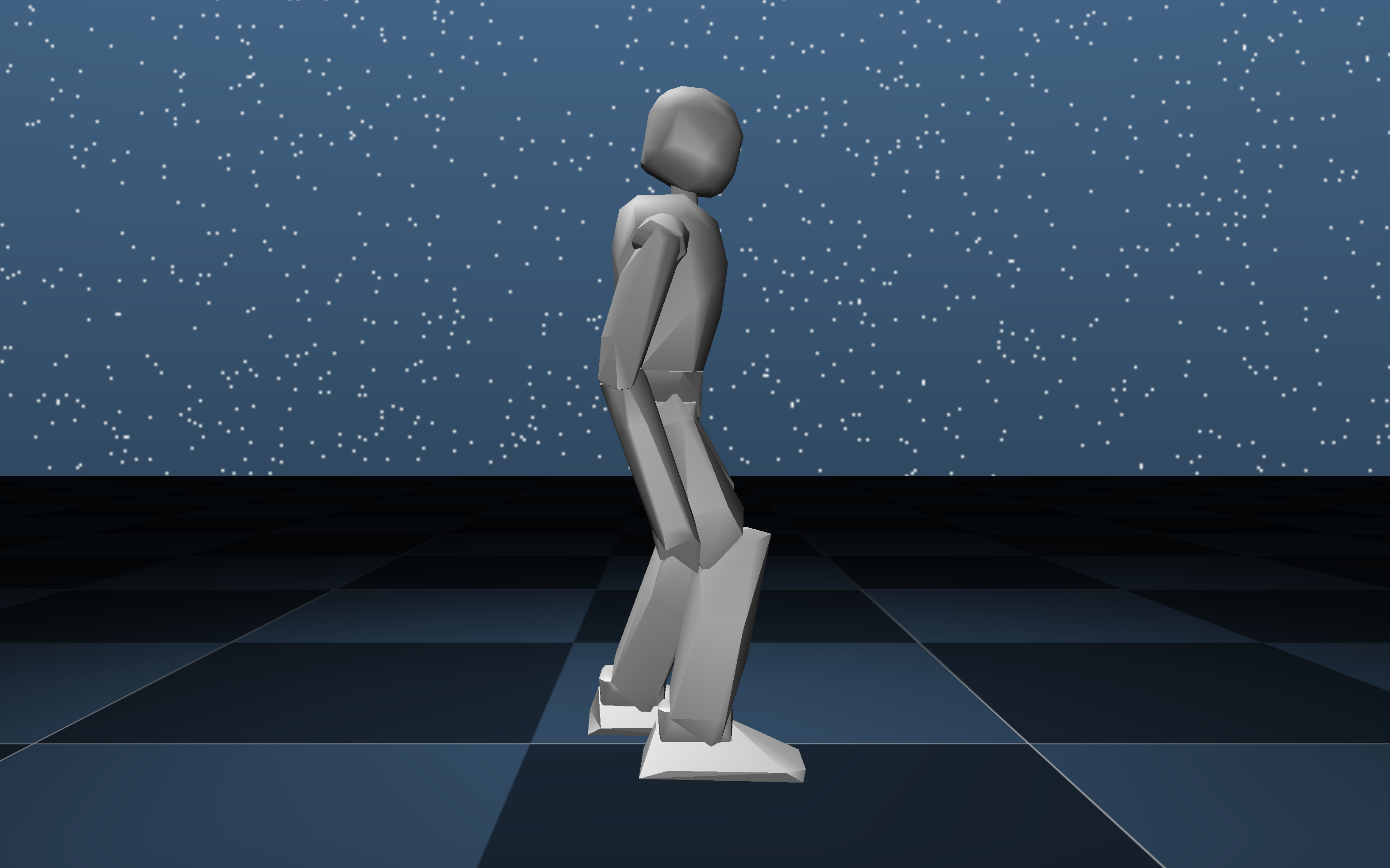}
		\includegraphics[width=0.13\linewidth, trim={600pt 0 600pt 0}, clip]{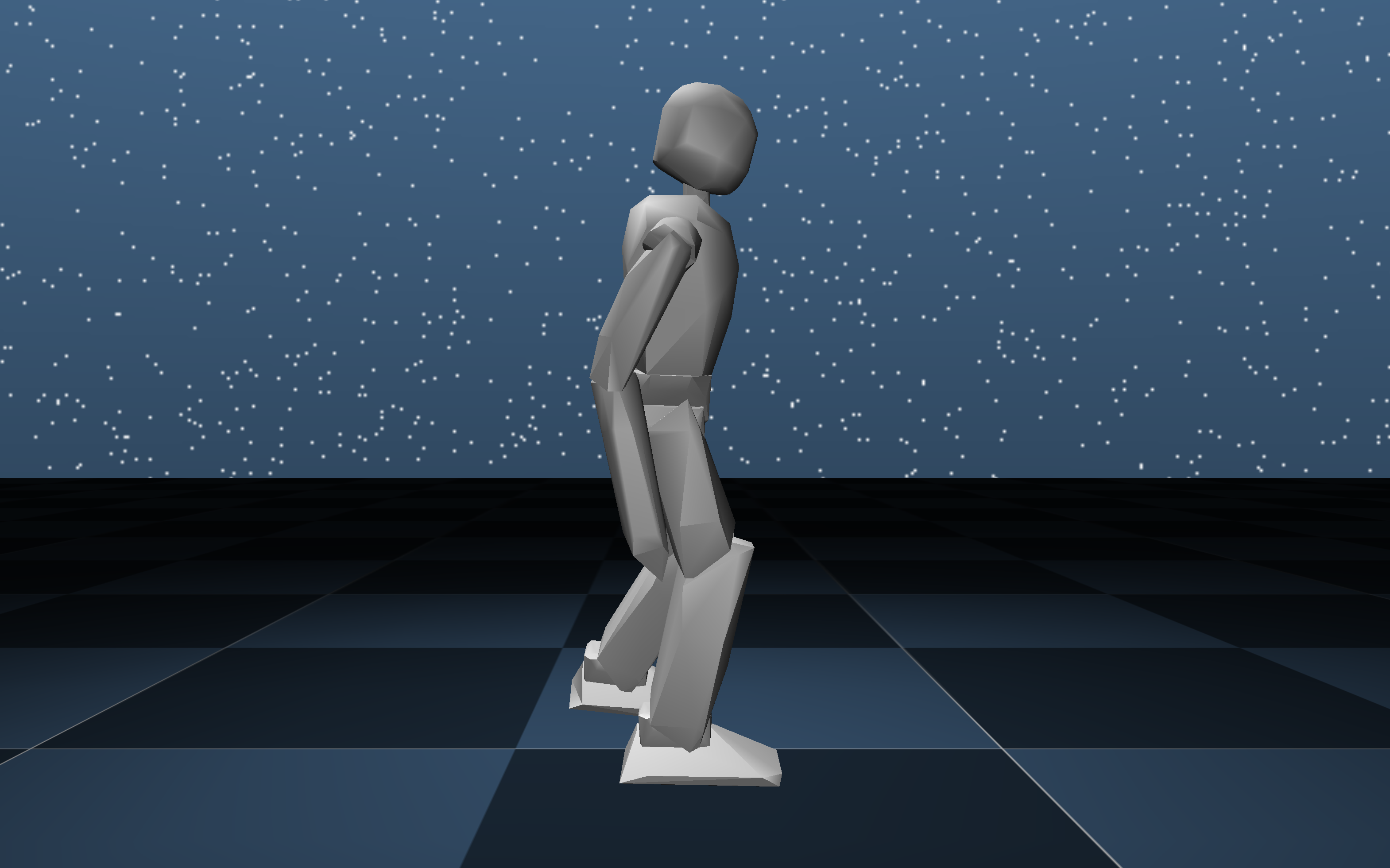}
		\includegraphics[width=0.13\linewidth, trim={600pt 0 600pt 0}, clip]{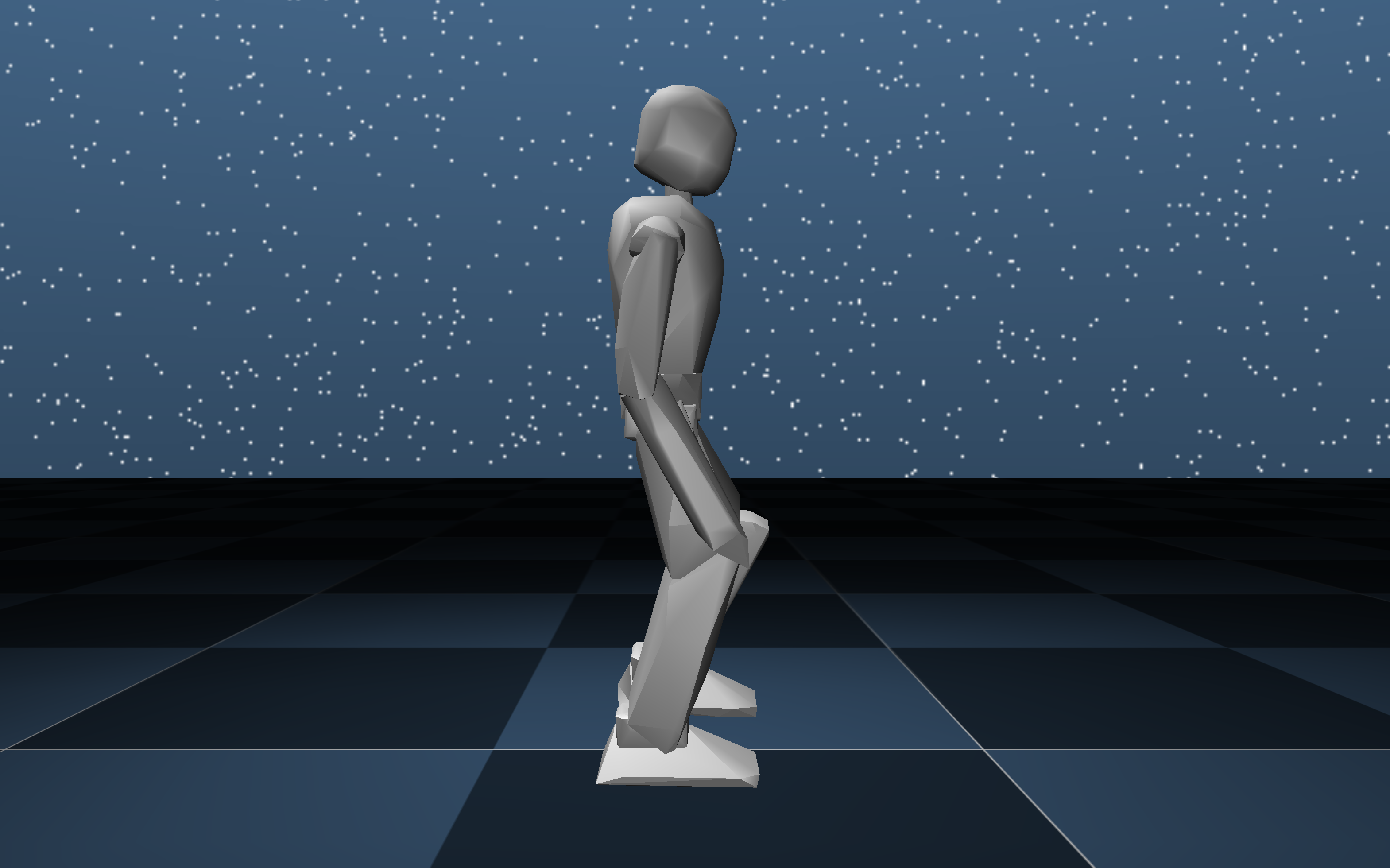}
		\includegraphics[width=0.13\linewidth, trim={600pt 0 600pt 0}, clip]{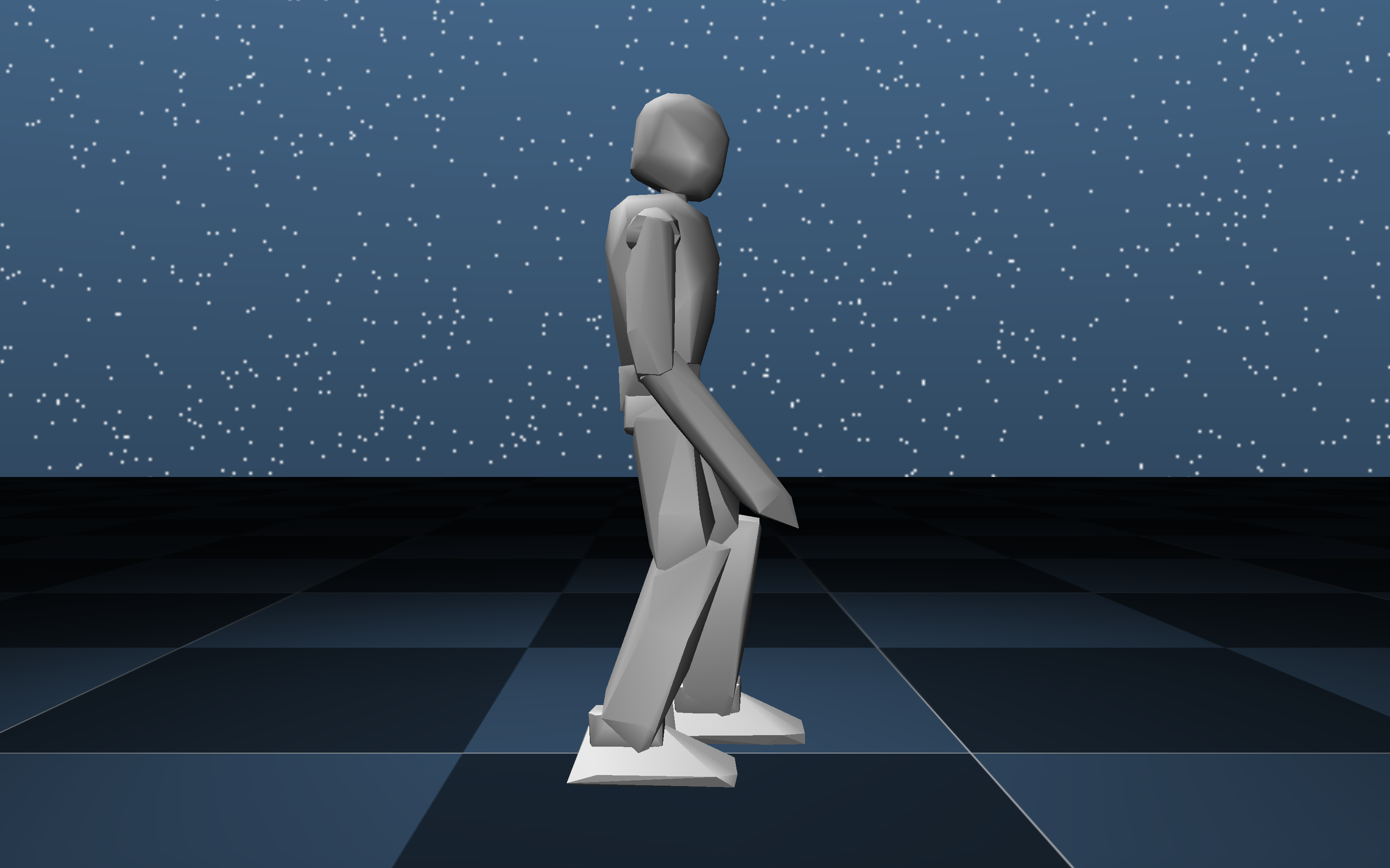} 
		\includegraphics[width=0.13\linewidth, trim={600pt 0 600pt 0}, clip]{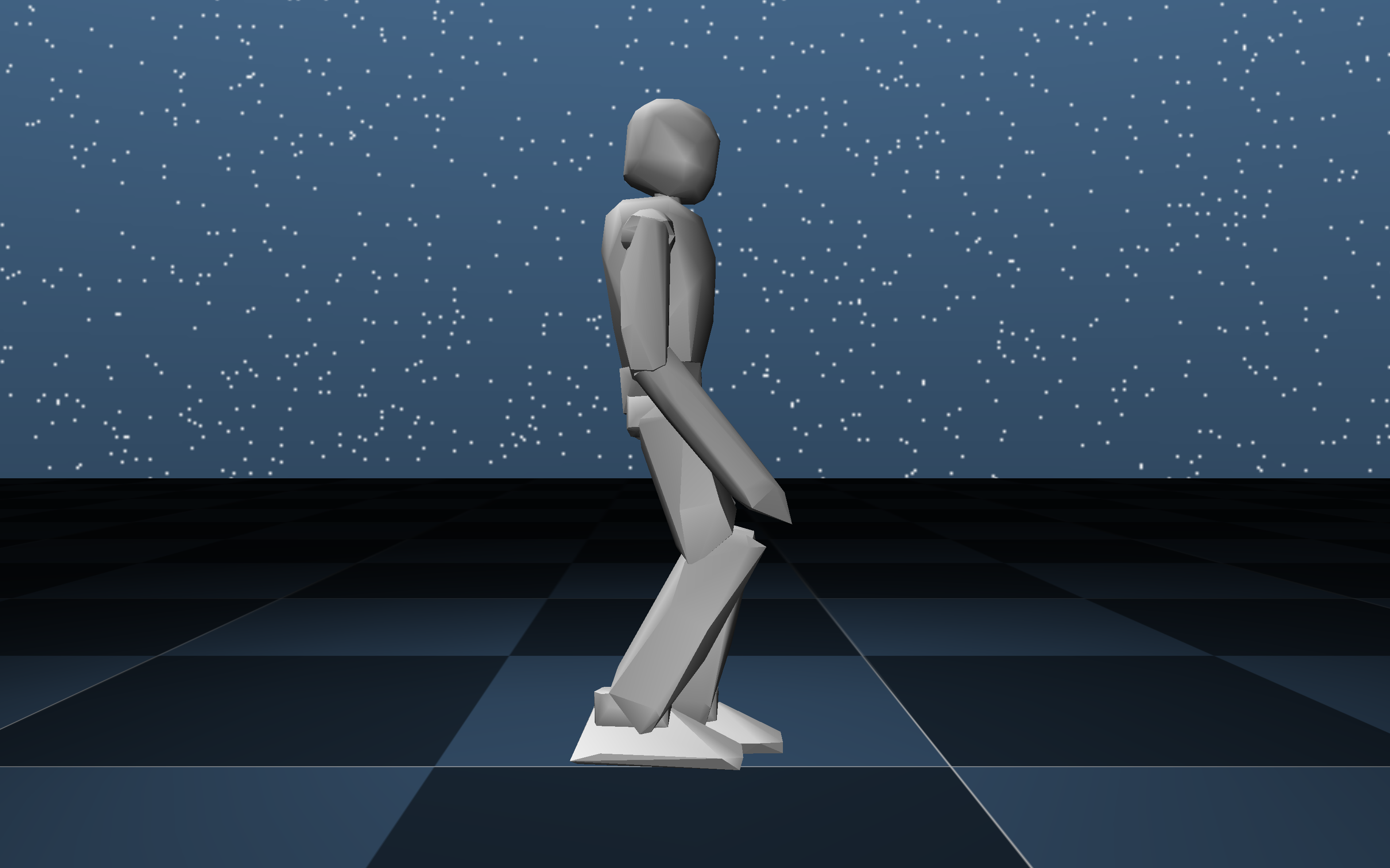}
		\includegraphics[width=0.13\linewidth, trim={600pt 0 600pt 0}, clip]{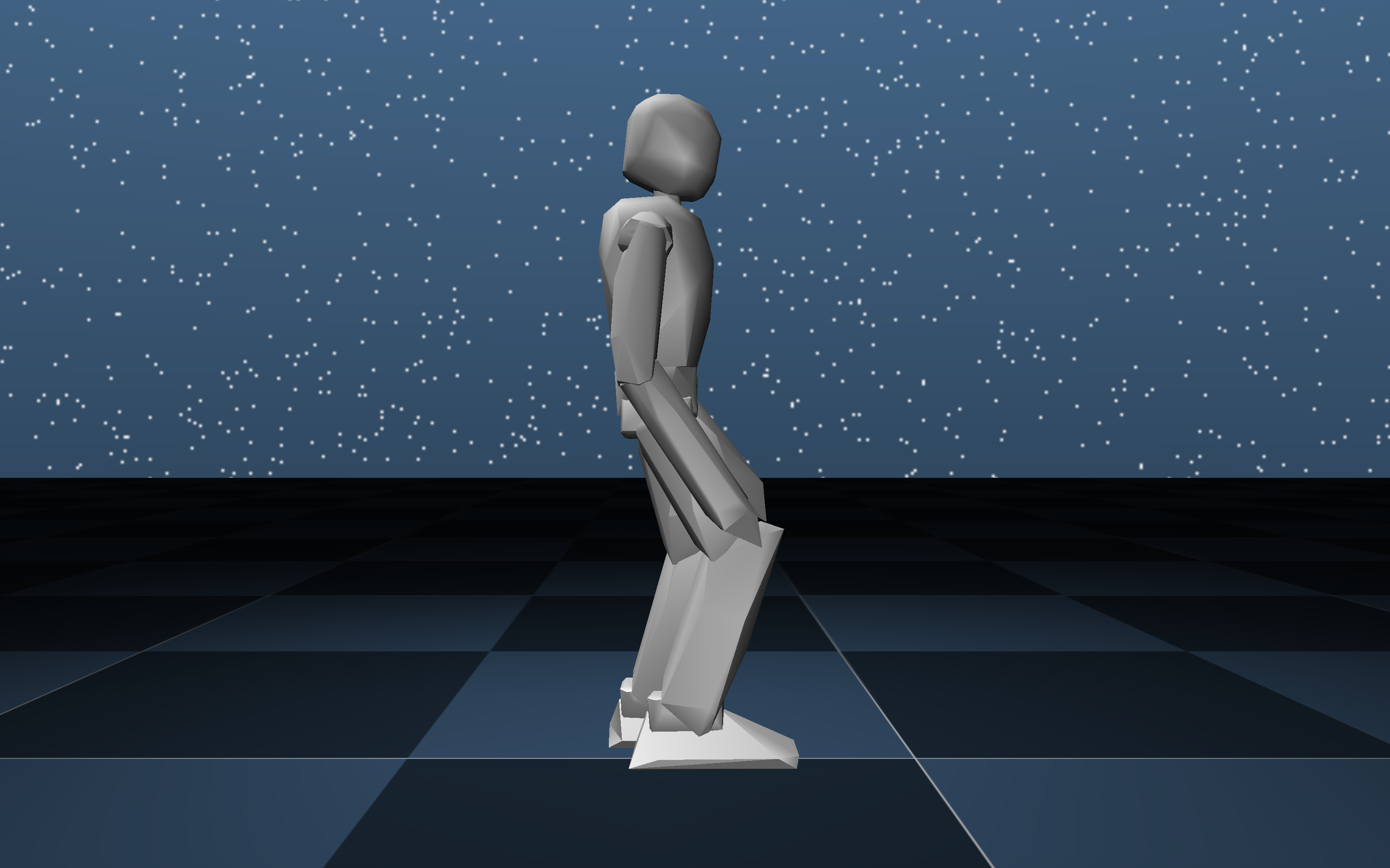}
		\includegraphics[width=0.13\linewidth, trim={600pt 0 600pt 0}, clip]{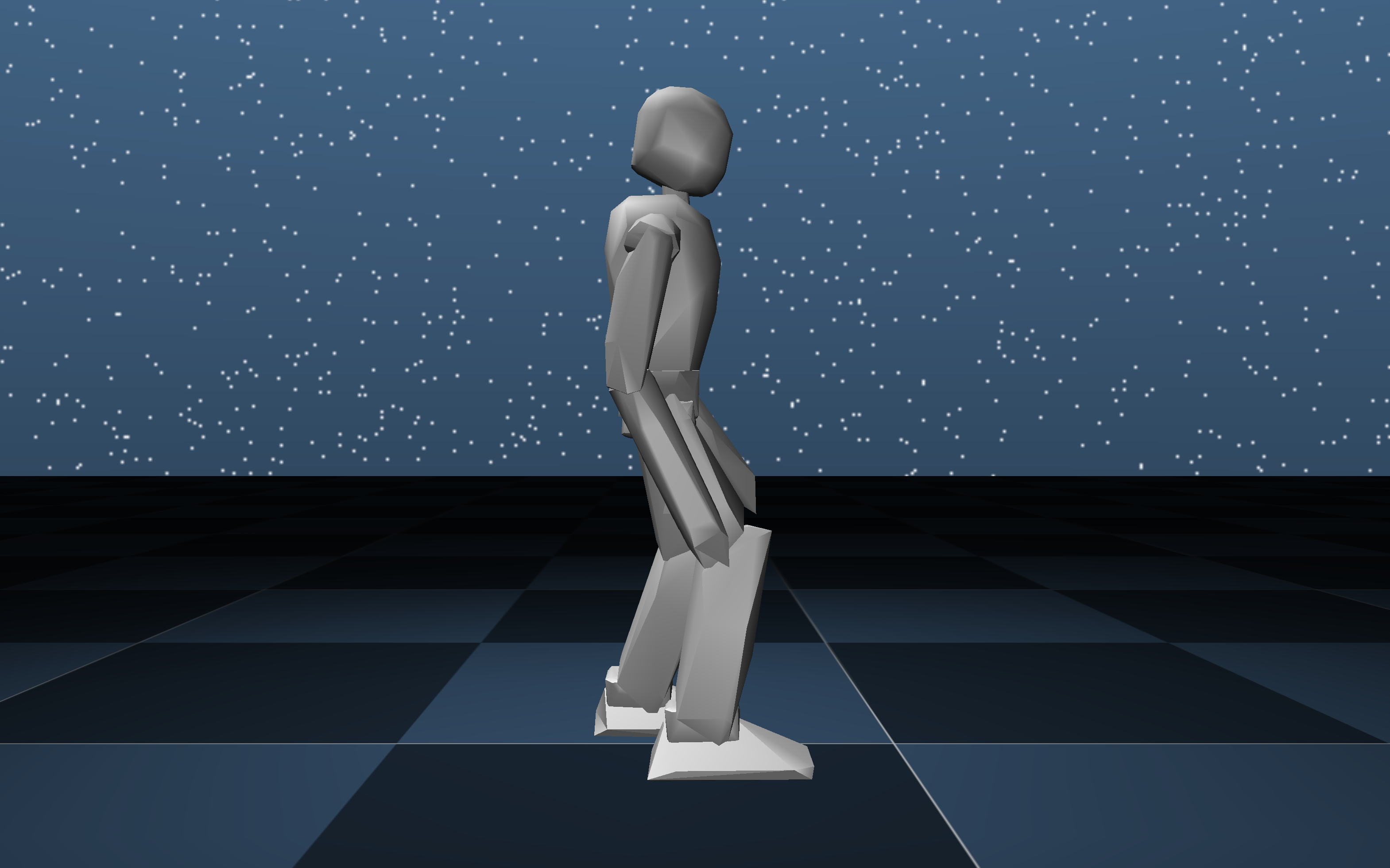}
		\caption{Bipedal gait of the NimbRo OP2X in simulation using the same gait parameters as used with the real robot.
			The shown sequence starts and finishes when the right foot makes contact with the floor depicting a full gait cycle.}
		\label{fig:mujoco_gait}
	\end{center}
\end{figure}

For developing and testing new approaches, e.g., bipedal gaits, 
the simulators are used before experiments are done with the real robot.
In this manner, the risk of damage and the initialization time are reduced almost to zero.
This requires fine tuning of the simulator such that it resemble as much as possible reality.
The contact with the floor is modeled as elliptic friction cones considering translational and torsional reaction forces.
Joint and torque limits together with gear ratios are incorporated in the model.
Because the real robot is controlled through joint position commands, 
PD controllers are implemented in simulation.
Hwangbo et al.~\cite{hwangbo2019learning} have demonstrated that learning the actuator model (input to output mapping) plays a key role for sim-to-real transfer.
Similarly,
we tune the gains of the controllers to obtain similar tracking errors as the real robot actuators~\cite{ficht2017nop2},
such that the actuator model with the controller together produce a similar control output as the real actuators given the same control input.
In this manner, the difference between the simulation model and the real actuators is reduced and the complex actuator modeling is avoided.
Figure~\ref{fig:mujoco_gait} shows the NimbRo OP2X robot walking in simulation using the same gait parameters that are used with the real robot.

\section{Evaluation} 
\label{eval}

The \nopx was developed in a total time of three months, which included the hardware design, manufacturing and software adaptation stages.
This was enough time to prepare the robot for the RoboCup humanoid robot soccer competition in June of 2018, in Montr\'eal, Canada.
In this section, a report on the results of the competition along with a more extensive statistical analysis of the robot's performance is presented.

\subsection{RoboCup 2018 Performance}
At RoboCup, different robots from all around the world face against each other in various competitions. One of the leagues focuses
on Humanoid robot soccer with three size classes: KidSize, TeenSize and AdultSize. In the AdultSize class where a robot needs to be at least \SI{130}{cm} tall, 
one vs. one soccer games and two vs. two mixed-team drop-in games are played. Additionally four technical challenges that evaluate 
certain features of the robot in isolation, are performed. The soccer games are carried out on a $6\times9$\,\SI{}{m} 
field, covered with artificial grass. In the competition, the \nopx showed an outstanding performance by winning all of the 
possible awards, which in turn resulted in obtaining the Best Humanoid Award.

During the soccer tournament \nopx participated in six one on one and five drop-in games. This summed up to 220 minutes of official play time, 
where 66 goals were scored, while only 5 were conceded. \nopx during the games is shown in \figref{soccer}.
\begin{figure}[]
	\centering
	\includegraphics[height=0.5\linewidth]{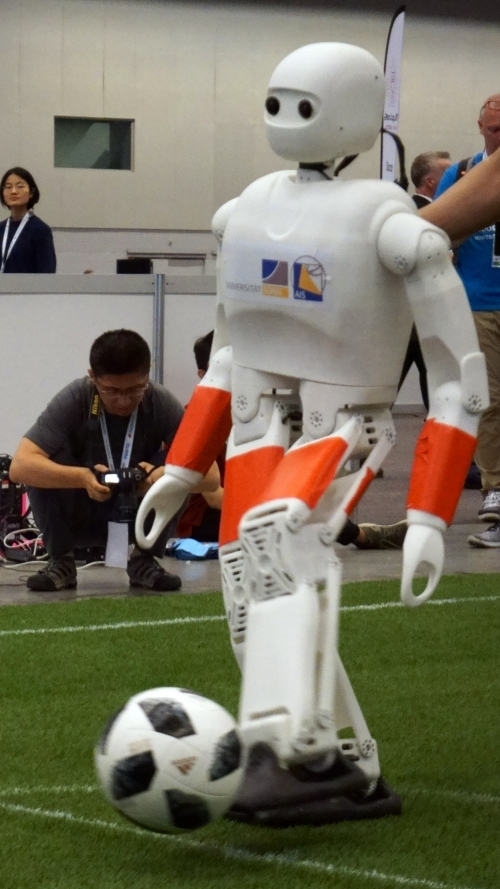}\hspace{1px}
	\includegraphics[height=0.5\linewidth]{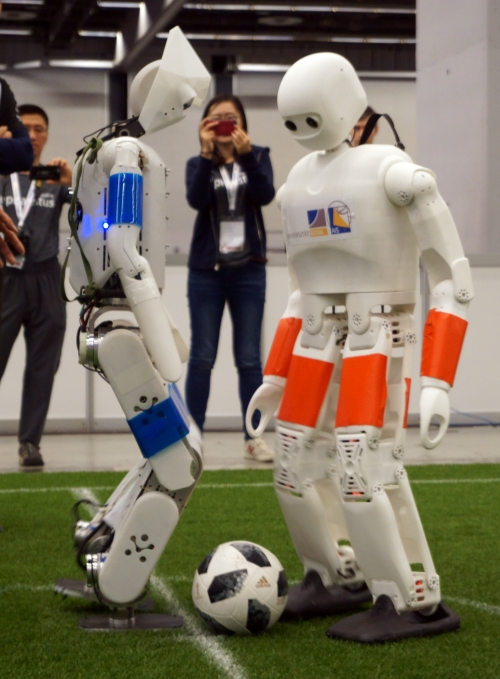}\vspace*{-1ex}
	\caption{\nopx during RoboCup 2018. Left: Performing a kick. Right: Competing for the ball.}
	\figlabel{soccer}\vspace{-2ex}
\end{figure}

The gait showed sufficiently fast walking speeds~(approx. \SI{0.5}{m/s}) and a high level of maintaining balance at the same time.
As a result, \nopx never fell while walking or dribbling in free space. In total, \nopx lost its balance four times. 
This was caused by collisions with other robots or stepping on their feet right before the beginning of their foot lifting motion.

Our vision system demonstrated good results as well. In the mixed and ever-changing lighting conditions, we could reliably
detect and track the ball position up to a distance of \SI{7}{m}, at which it was represented by only a few pixels in the image. 
Our robust robot detections allowed for efficient path planning and dribbling the ball around the opponents.

In the \textit{Push Recovery} technical challenge, an external disturbance is applied by a \SI{3}{kg} heavy pendulum swinging on a \SI{1.5}{m} long 
string. The weight hits the robot at the height of the Center of Mass (COM). By varying the distance $d$ of the pendulum to the robot before 
releasing it, the value of the applied impulse is controlled. \nopx withstood pushes with $d = 0.8$\,m. 
One of these trials that was captured during the competition can be seen on \figref{push_recovery}. 
The \textit{Moving Ball} challenge, requires the robot to score a goal after a pass from a human. It took \nopx only 3 seconds to score
a goal, counting from the moment when the ball was passed. This was mostly possible thanks to our fast vision pipeline and capable hardware.

\begin{figure*}[]
	\centering
	\includegraphics[height=0.27\linewidth]{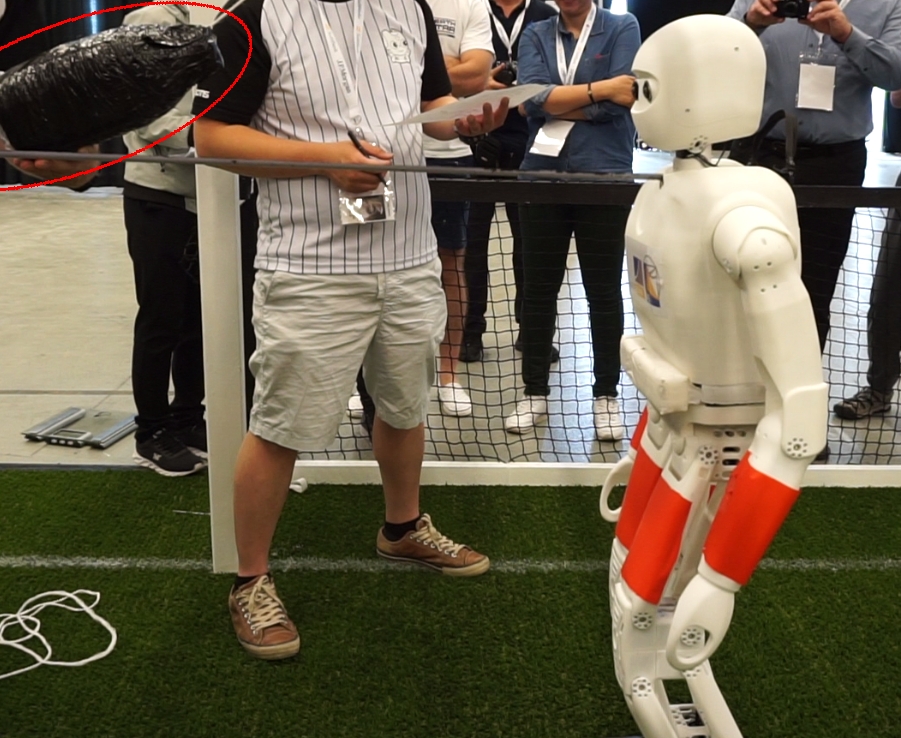}\hspace{1px}
	\includegraphics[height=0.27\linewidth]{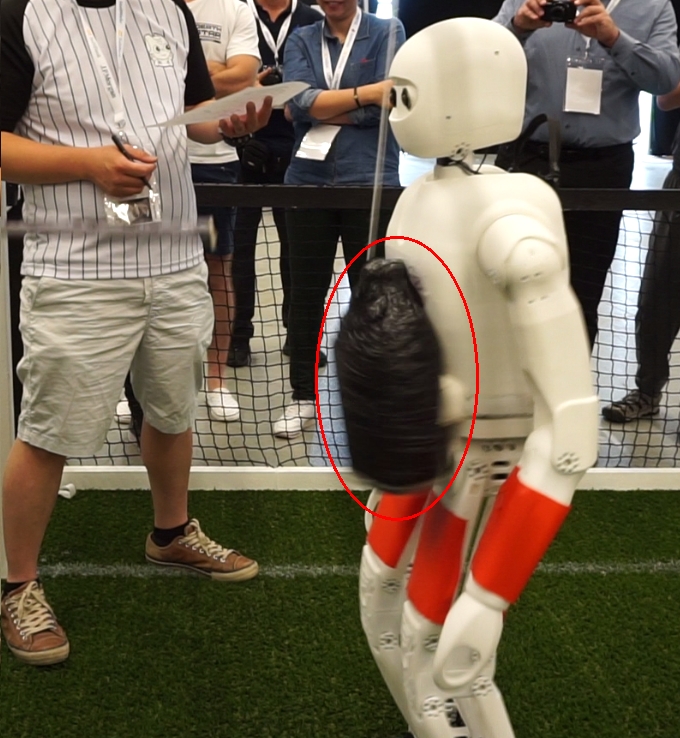}\hspace{1px}
	\includegraphics[height=0.27\linewidth]{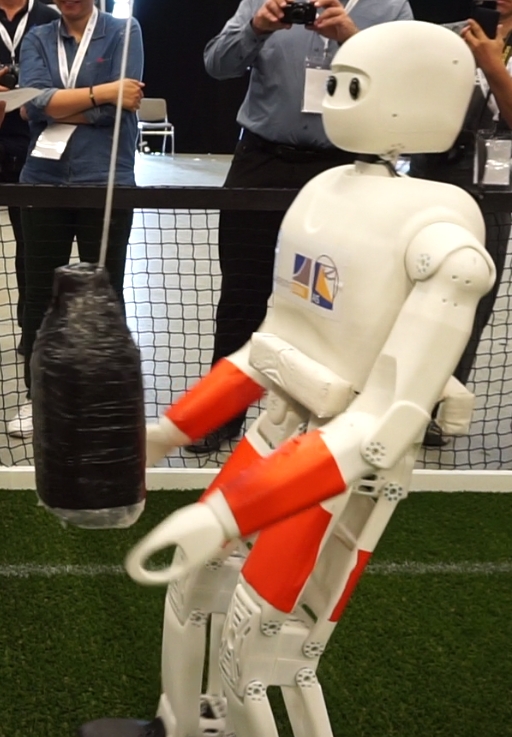}\hspace{1px}
	\includegraphics[height=0.27\linewidth]{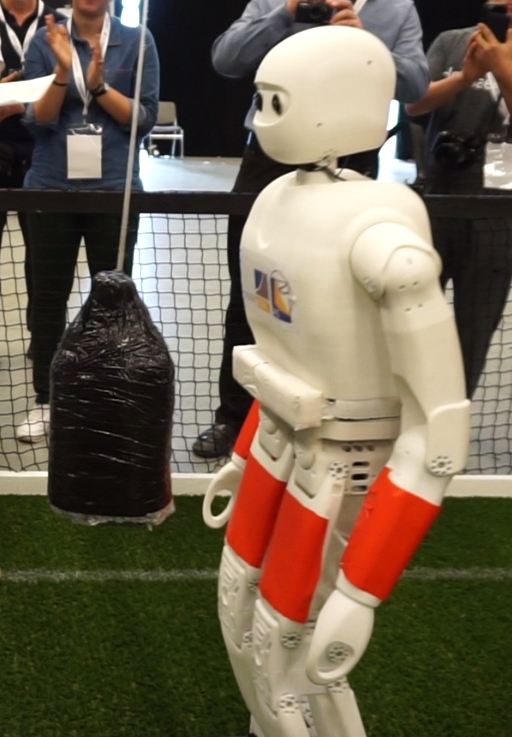}\vspace*{-1ex}
	\caption{\nopx withstanding a push from the front. The weight is annotated with a circle. After receiving the push, the robot performs corrective actions to recover balance.}
	\figlabel{push_recovery}\vspace{-2ex}
\end{figure*}

\subsection{Balanced Walking}
\seclabel{walking_stability}

Despite the fact that performance during RoboCup 2018 competition allowed to evaluate the capabilities of the platform in strict 
conditions (imperfect artificial grass with mixed and varying lighting at different times of the day), 
such an evaluation lacks the statistical accuracy in certain aspects. We have therefore performed several experiments to 
quantitatively assess the robot's capabilities. These were focused on the balanced walking of \nopx and consisted of: walking straight on the 
grass field and withstanding a push while walking in place. 

For the following experiments, we evaluate the balancing capabilities with respect to the fused angle deviations. Due to the 
low-cost nature of the \nopx, the available feedback during locomotion is joint position sensing and attitude estimation.
Although obtaining estimates of other balance indicators from joint position feedback is theoretically possible, it has 
been largely inaccurate in our experience, due to inconsistent results provided by the sensors. Therefore the verification of balance is based solely on the IMU measurements.
A maximum angle for the fused pitch and fused roll of the trunk are established, where exceeding these values results
in the robot tipping over and falling, despite any corrective actions taken. The measurements taken are then represented in the 
form of phase plots, which accurately represent the system behavior.

The first evaluation has been performed by commanding the robot to walk straight on artificial grass for \SI{4}{m}. The test was performed with different 
speeds, starting at a low \SI{0.03}{m/s} and gradually increasing it by \SI{0.1}{m/s} until the robot was losing balance more than \SI{50}{\%} of 
the attempts, which happened with a speed of more than \SI{0.7}{m/s}. The results for each speed is an average of four trials.
In two of the trials the robot is walking with the direction of the artificial grass blades, in the other two --- against the blades. 
The maximum walking speed that the robot was able to achieve without falling a single time was \SI{0.67}{m/s} or \SI{2.41}{km/h},
which is almost that of the walking speed of Honda's Asimo (given as \SI{2.7}{km/h}).

To precisely determine the behavior of the gait, we recorded the fused pitch value and its rate of change~(velocity).
\figref{walking1} presents the phase plot of the fused pitch velocity for different walking speeds: \SI{0.03}{m/s}, 
\SI{0.2}{m/s}, and \SI{0.5}{m/s}. While at low speed the fused pitch velocity stays close to the 
nominal zero(measured when the robot is standing still), it starts scattering when the speed is increased.

The standard deviations of the fused pitch and its derivative shown on \figref{walking2} clearly exhibit a smooth 
non-linear growth of these values with an increase of walking speed. Trends of these values for the case of walking 
with a forward velocity of \SI{0.67}{m/s} are shown in \figref{walking3}. Note that during the initial acceleration 
phase the gait is irregular. After achieving the set velocity, the robot starts walking in a predictable, stable manner.

\begin{figure}
	\centering
	\includegraphics[width=0.32\linewidth, trim={0 0 40pt 20pt}, clip]{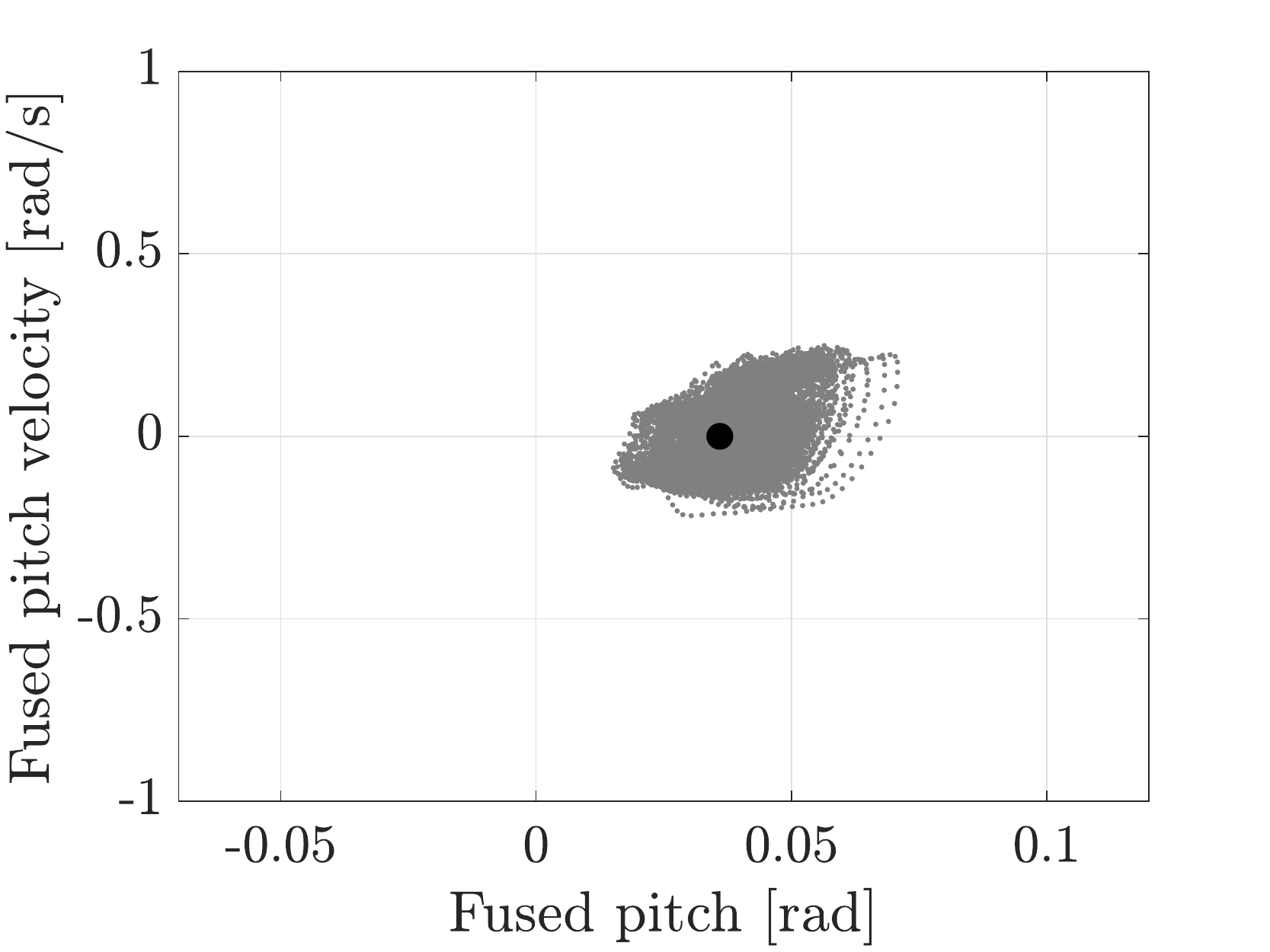}
	\includegraphics[width=0.32\linewidth, trim={0 0 40pt 20pt}, clip]{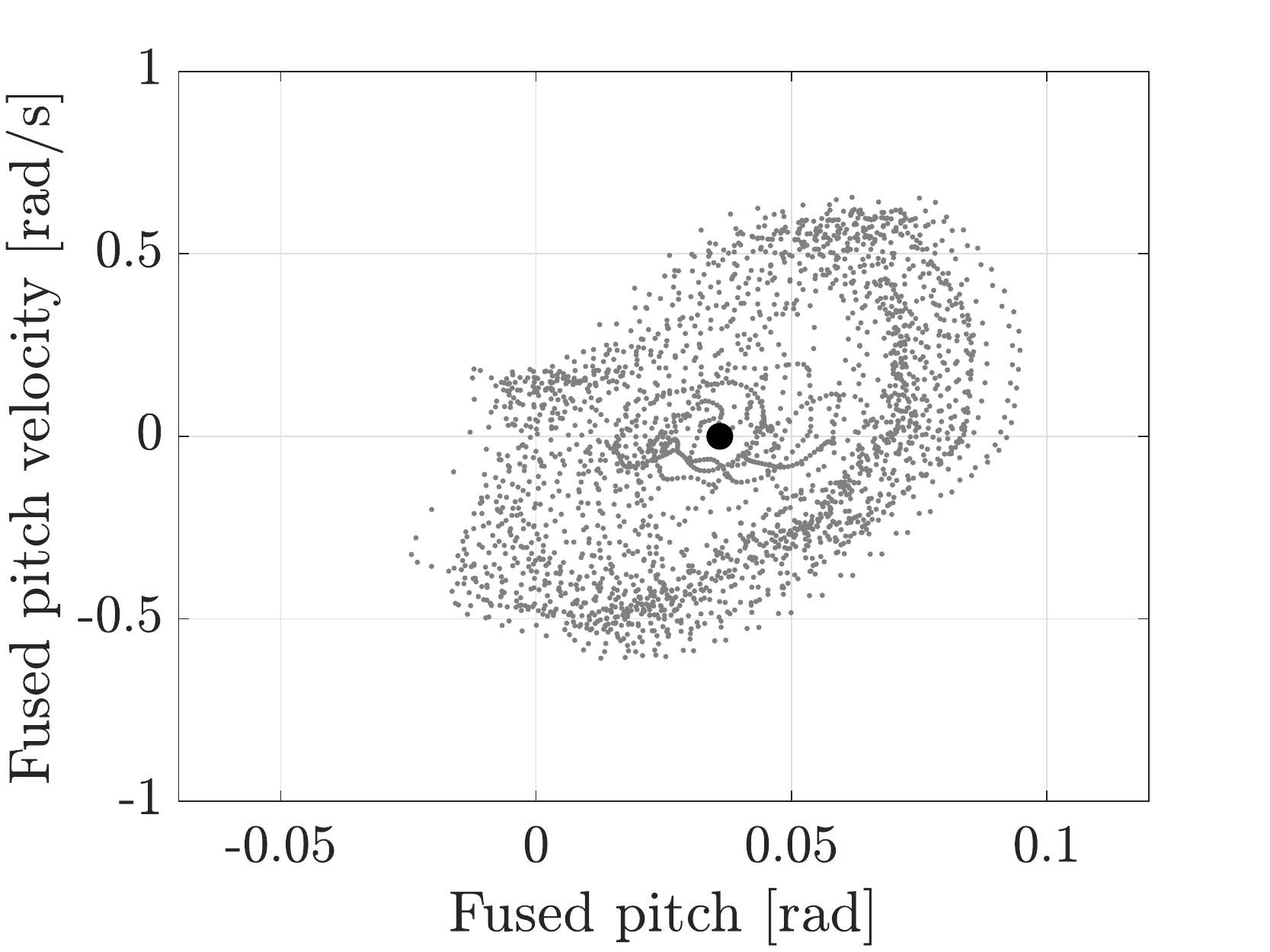}
	\includegraphics[width=0.32\linewidth, trim={0 0 40pt 20pt}, clip]{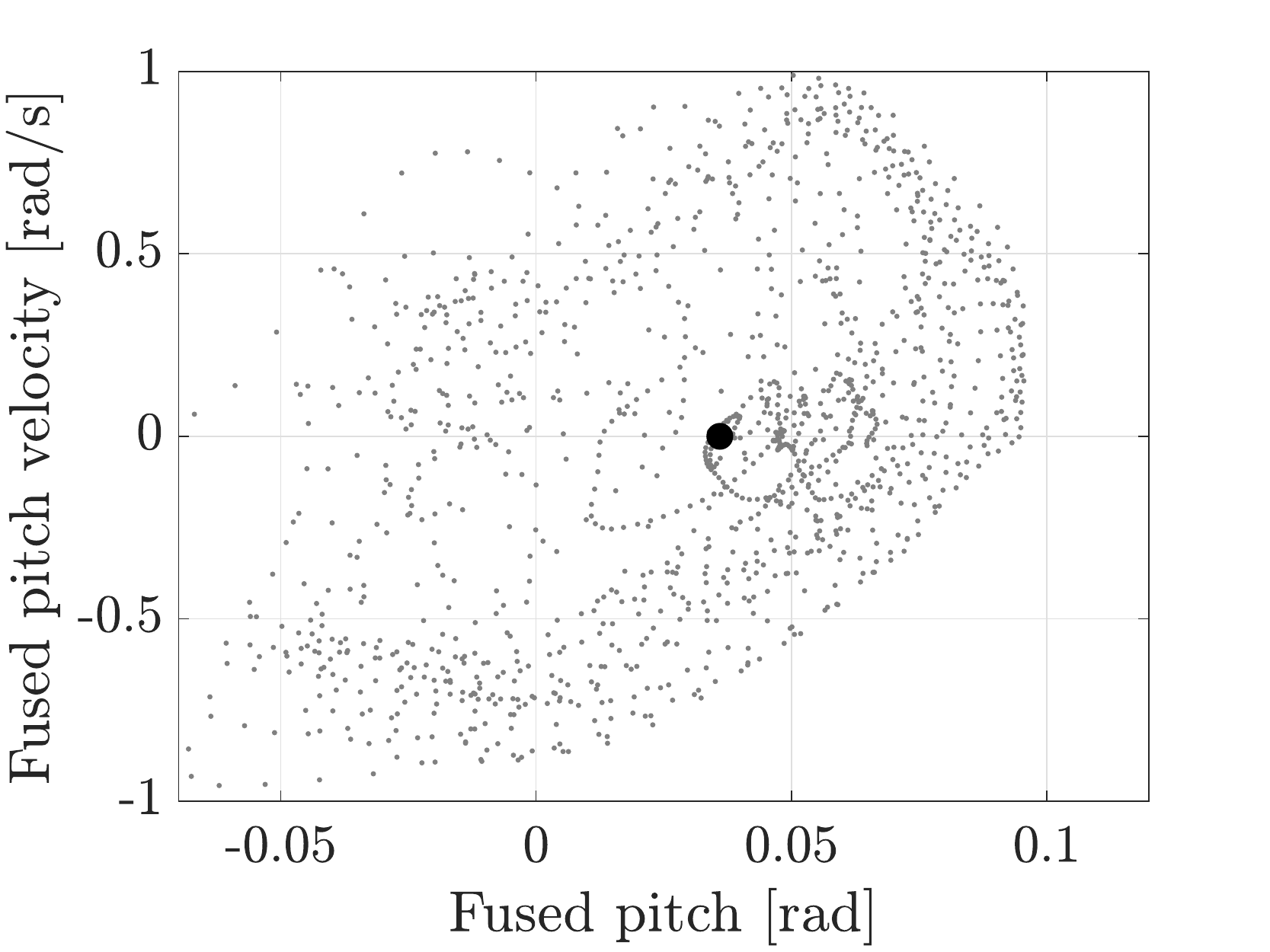}
	\caption{Phase plot of the fused pitch velocity during walking with varying forward velocity. From left to right: \SI{0.03}{m/s},  
	\SI{0.2}{m/s}, \SI{0.5}{m/s}. Nominal zero is marked with a black dot. 
	Note that velocities start to vary significantly when walking faster, while staying close to the nominal zero during a slow walk.}
	\figlabel{walking1}
\end{figure}

\begin{figure}
	\centering
	\includegraphics[width=0.99\linewidth, trim={50pt 5pt 80pt 10pt}, clip]{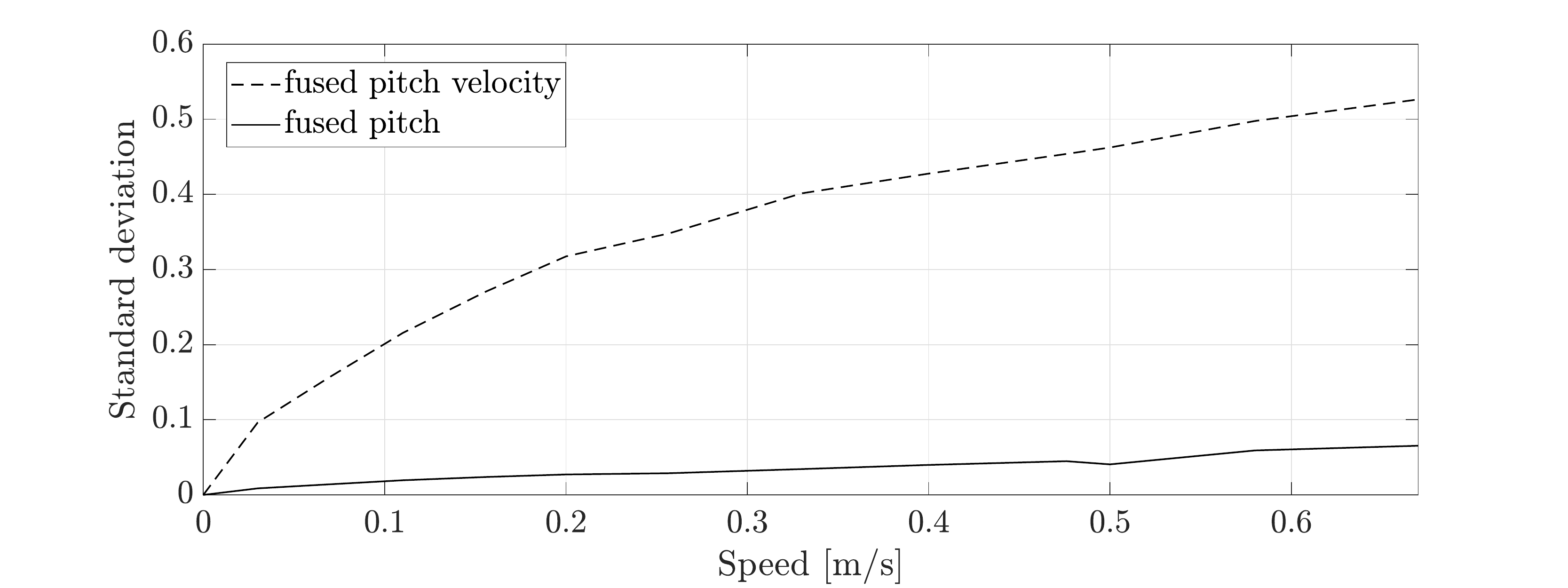}
	\caption{Standard deviations of fused pitch and its derivative versus speed of walking. Both values grow smoothly with the increase of walking speed.}
	\figlabel{walking2}
\end{figure}

\begin{figure*}
	\centering
	\includegraphics[width=0.99\linewidth, trim={50pt 5pt 80pt 10pt}, clip]{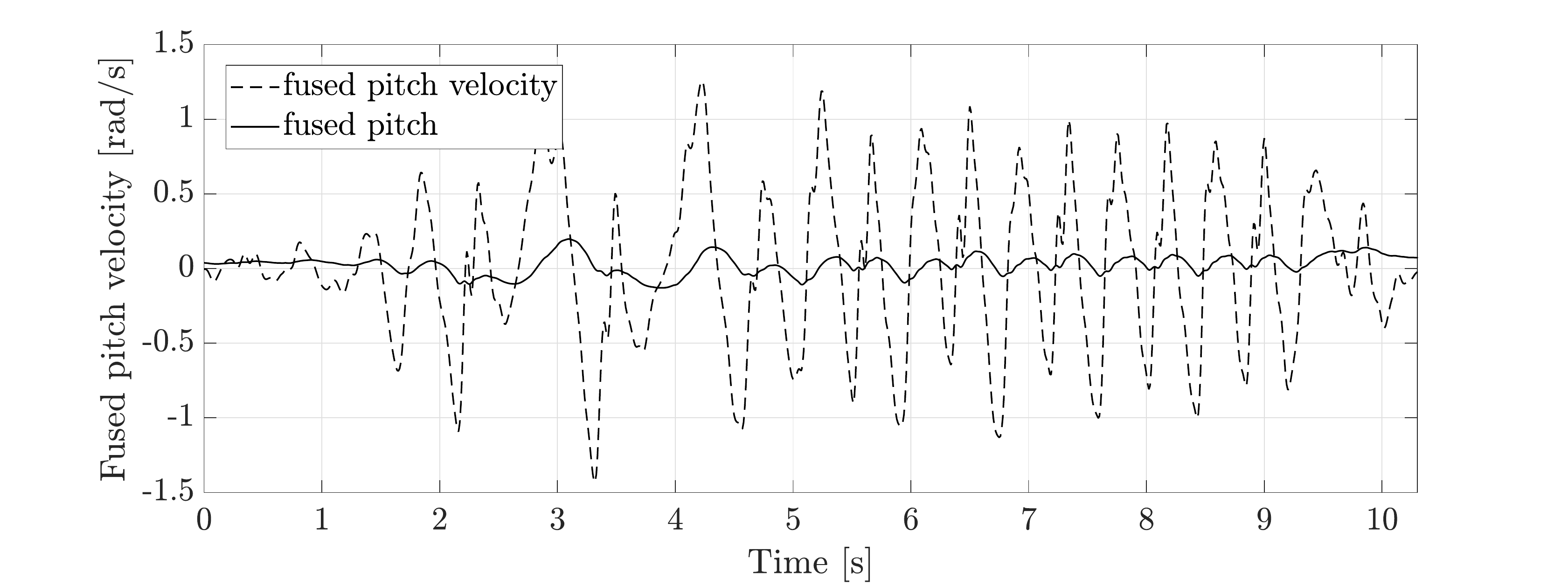}
	\caption{Fused pitch velocity versus time for walking with speed of 0.67\,m/s. The first half of the plot is more unstable as result of the robot acceleration actions to reach the set speed. Once it is achieved, the walking pattern is stable.}
	\figlabel{walking3}
\end{figure*}

A second experiment evaluated the ability of \nopx to withstand pushes while walking on the spot. 
The experiment was performed with a pendulum which was released freely after being retracted by a distance $d$. 
A weight of \SI{3}{kg} was attached to the end of a rope of length \SI{1}{m}. The impact was dealt at the height of the robot's center of mass. 
Pushes from the front and from the back of the robot were performed, 10 from each side.

The robot demonstrated better balance when pushed from the front, since it naturally slightly leans forward. 
The best result for pushes from the back was withstanding 40\,\% 
of the pushes with a retraction distance of the pendulum $d=90$\SI{}{cm} which corresponds to impulse of \SI{9.51}{kg}$\cdot$\SI{}{m/s}. 
However, in case of front pushes the robot successfully withstood 60\,\% of pushes with $d=100$\SI{}{cm} which corresponds 
to an impulse of \SI{10.14}{kg}$\cdot$\SI{}{m/s}. In \figref{pushing2}, phase plots of fused pitch velocity are shown, which were 
recorded during pushes from the front with $d=50$\SI{}{cm}, and $d=100$\SI{}{cm}. The strength of the push and reaction of the system can be clearly observed
by how the points are more spread out on the right side of the figure, but still manage to return to the nominal value. 
An example of a successful trial from a frontal push with $d=100$\SI{}{cm} can be seen on \figref{pushing3}.
\begin{figure}
	\centering
	\includegraphics[width=0.48\linewidth, trim={0 0 40pt 20pt}, clip]{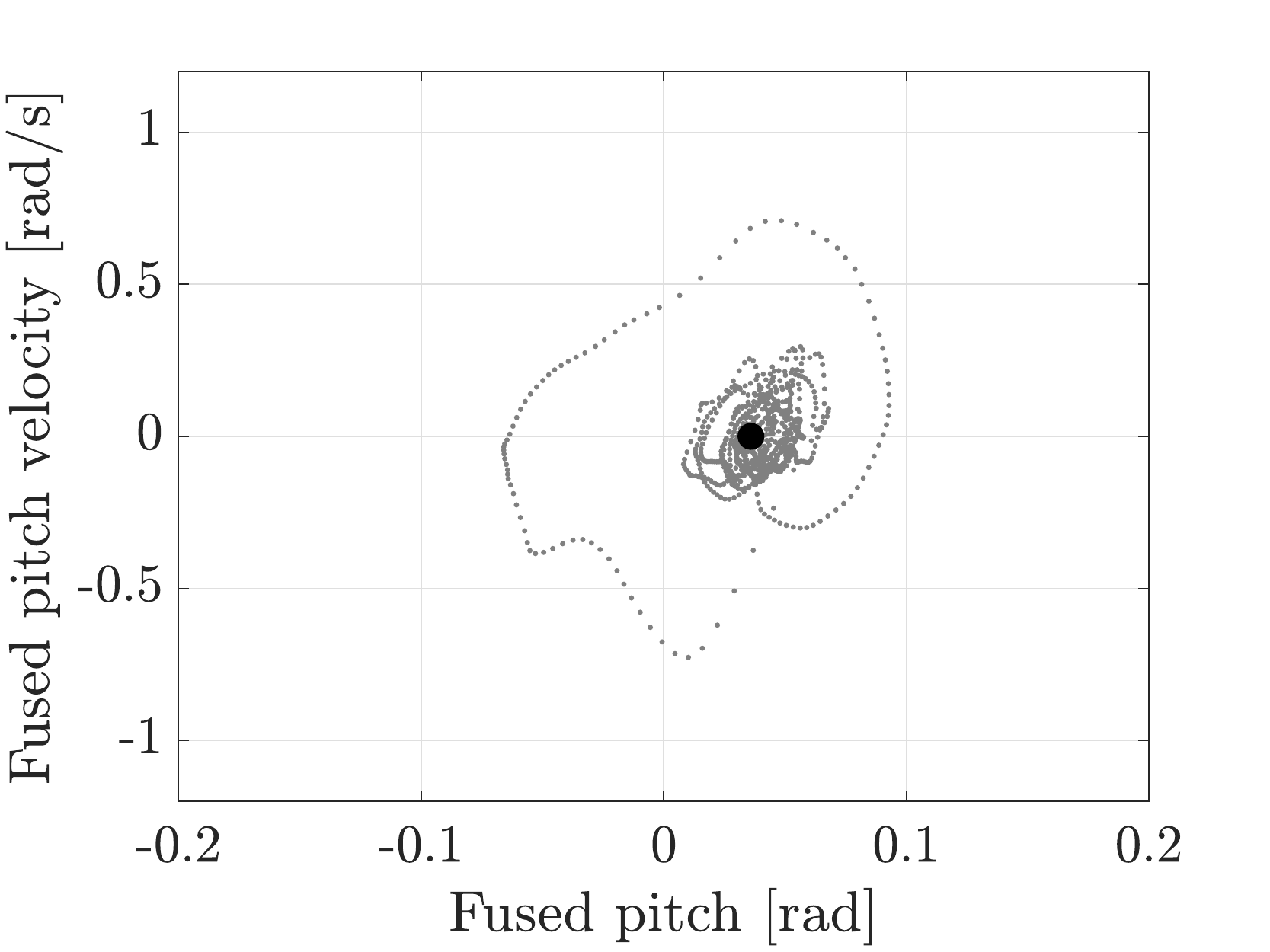}
	\includegraphics[width=0.48\linewidth, trim={0 0 40pt 20pt}, clip]{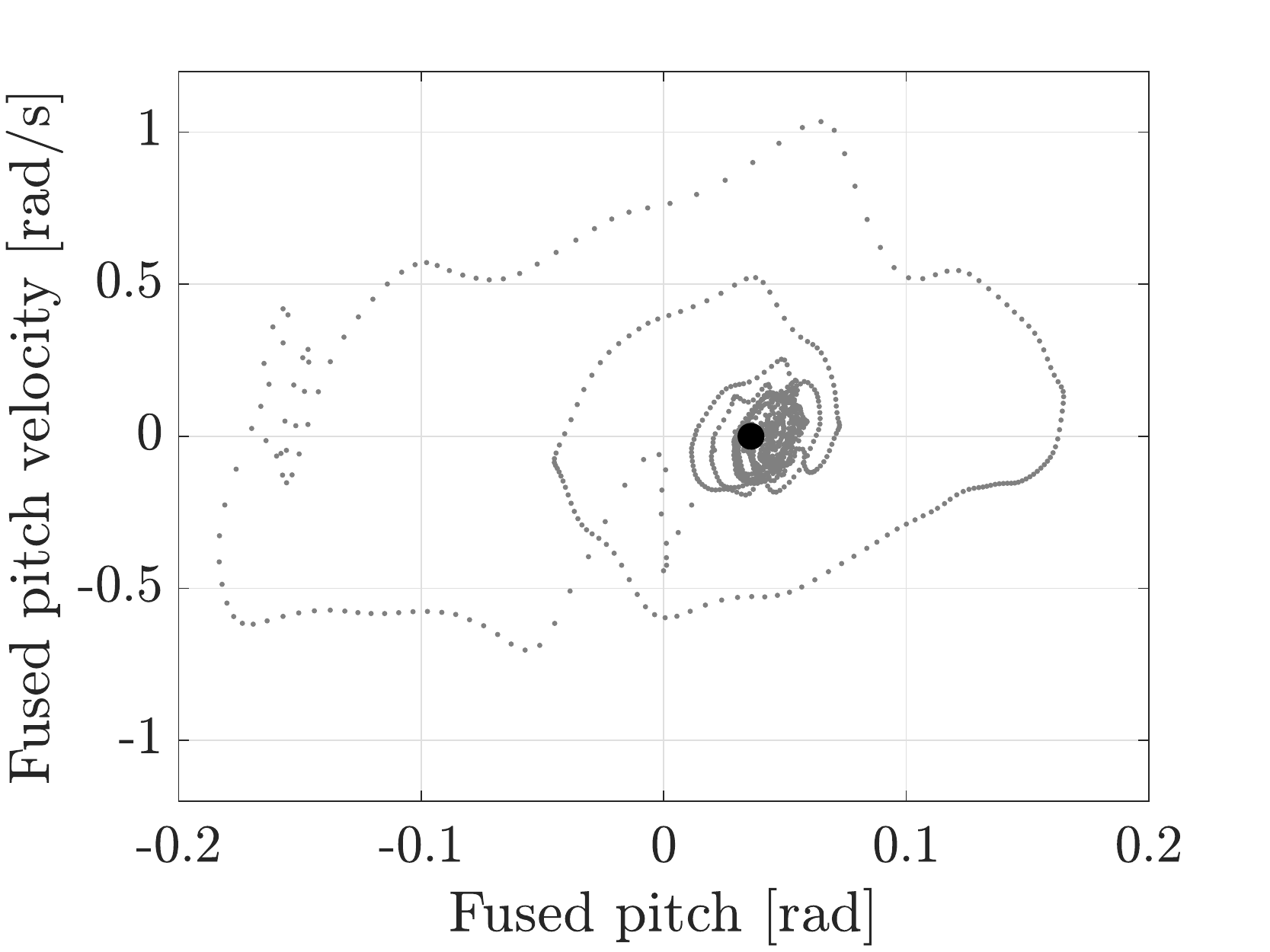}
	\caption{Phase plot of the fused pitch velocity during experiencing a push from the front. Left: 
	push with $d=50$\,cm; Right: push with $d=100$\,cm. Nominal zero is marked with a black dot.}
	\figlabel{pushing2}
\end{figure}

\begin{figure*}
	\centering
	\includegraphics[width=0.99\linewidth, trim={50pt 5pt 80pt 10pt}, clip]{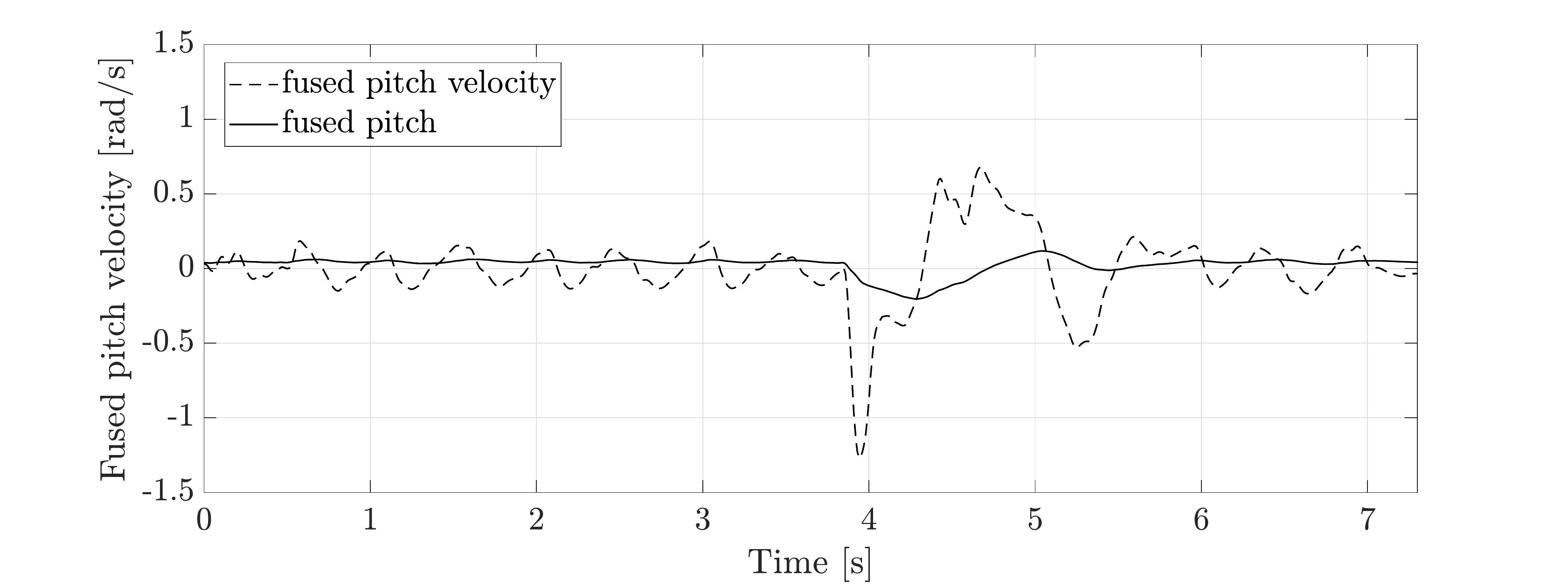}
	\caption{Fused pitch velocity versus time while withstanding a push from the front with retraction 
	distance of the pendulum $d=100$\,cm. The push can be observed in the middle and the recovered stable pattern afterwards.}
	\figlabel{pushing3}
\end{figure*}

\subsection{Soccer Object Detection} 
\label{evaluation_perception}

\begin{table}[t]
	\caption {Results of our visual perception network.} \label{vis_res} 
	\begin{center}
		\begin{tabular}{ | l || c  c  c  c  c |}
			\hline
			Type & F1 & Accuracy & Recall & Precision & FDR \\
			\hline
			Ball (ours) & \textbf{0.997} & \textbf{0.994} & \textbf{1.0} & \textbf{0.994} & \textbf{0.005} \\
			Ball (SweatyNet-1~\cite{schnekenburger2017detection}) & 0.985 & 0.973 & 0.988 & 0.983 & 0.016 \\
			\hline
			Goal (ours) & \textbf{0.977} & \textbf{0.967} & \textbf{0.988} & \textbf{0.966} & \textbf{0.033} \\
			Goal (SweatyNet-1~\cite{schnekenburger2017detection}) & 0.963 & 0.946 & 0.966 & 0.960 & 0.039 \\
			\hline
			Robot (ours) & \textbf{0.974} & \textbf{0.971} & \textbf{0.957} & \textbf{0.992} & \textbf{0.007} \\
			Robot (SweatyNet-1~\cite{schnekenburger2017detection}) & 0.940 & 0.932 & \textbf{0.957} & 0.924 & 0.075 \\
			\hline
			Total (ours) & \textbf{0.983} & \textbf{0.977} & \textbf{0.982} & \textbf{0.984} & \textbf{0.015} \\
			Total (SweatyNet-1~\cite{schnekenburger2017detection}) & 0.963 & 0.950 & 0.970 & 0.956 & 0.043 \\
			\hline
		\end{tabular}
	\end{center}
\end{table}

To evaluate our visual perception pipeline, we compare its resuts on different soccer-related objects against SweatyNet~\cite{schnekenburger2017detection}.
The comparison can be seen in Table.~\ref{vis_res}. It is important to note that both methods were evaluated using the same training and test sets. It can be observed that
our usage of the transfer learning helped us in achieving excellent results with limited training samples. 
We have outperformed SweatyNet, for which the results were one of the best-reported in terms of detecting soccer objects. 
Training on the same machine equipped with a single Titan Black GPU with \SI{6}{GB} of memory, our method achieved superior results.
Our visual perception pipeline is also approximately three times faster than SweatyNet. 
The reduced time can be attributed to the progressive image resizing and transfer learning techniques. Overall our robots can detect soccer objects 
with excellent accuracy and very little false detection rate.

\section{Conclusions} %

In this paper, we have presented the \nopx, an adult-sized open-source humanoid robot platform 
aimed towards research. The hardware specifications and software modules of the robot were described in detail. 
A thorough evaluation has been carried out, which showcased the robot's physical and visual capabilities. The minimalistic and modular 3D-printed 
hardware structure makes maintenance effortless and welcomes user modifications. Through the use of inexpensive, 
yet effective means, the robot displayed its excellent performance both in a controlled and uncontrolled environment.
We have shown that the \nopx has enough computing power, and a mature enough software framework to easily implement 
methods for more general use cases that work in real-time. These are not limited to playing soccer, but also can
include human interaction, object detection and even manipulation (after an addition of a gripper).
Both the software framework and hardware (3D-printable CAD files along with a Bill of Materials) of the \nopx are 
openly available online, free of charge. This open-source aspect, paired with the inexpensive and rapid 
production techniques greatly contributed to the fast production time. We hope that the \nopx will bring 
more active participants to the field of humanoid robotics, which will result in its wider dissemination.

\section*{Acknowledgements} 

This work was partially funded by grant BE 2556/13 of the German Research Foundation (DFG).

\vfill
\noindent%
\parbox{5truein}{
\begin{minipage}[b]{1truein}
\centerline{{\includegraphics[width=1in]{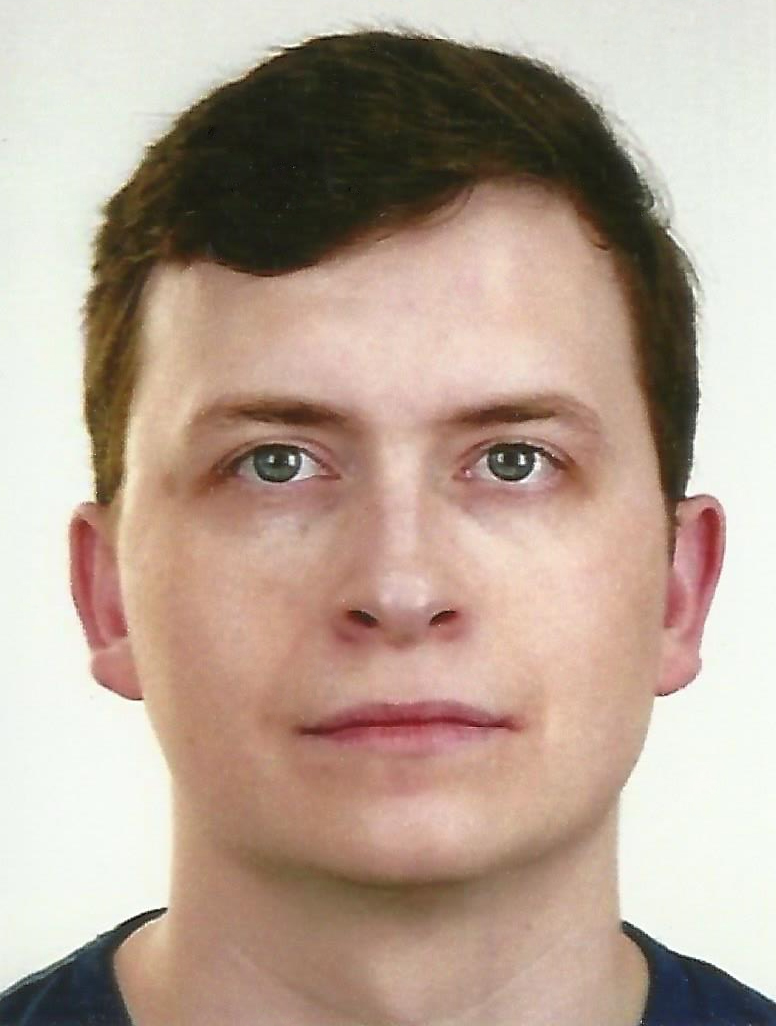}}} %
\end{minipage}
\hfill %
\begin{minipage}[b]{3.85truein}
{{\bf Grzegorz Ficht} received his M.S. degree in Robotics and Control Engineering
	from Gdansk University of Technology, Poland in 2013. Since 2015, he 
	is a researcher at the University of Bonn, Germany in the Autonomous Intelligent Systems group,
	where he is working towards obtaining a Ph.D. degree in Computer Science.
	Apart from designing~the~\noptwo~and~\nopx~robots, he has actively contributed to the 
	EuRoC project. He has also won nu-\hfilneg}
\end{minipage} } %

\vspace*{4.8pt}  
\noindent
merous RoboCup awards in the Humanoid Soccer League. Grzegorz Ficht is the author of over 15 technical publications. His research interests include the 
design of humanoid robots and using template models for their control with respect to motion and balancing.

\vspace*{13pt}  
\noindent%
\parbox{5truein}{
	\begin{minipage}[b]{1truein}
		\centerline{{\includegraphics[width=1in]{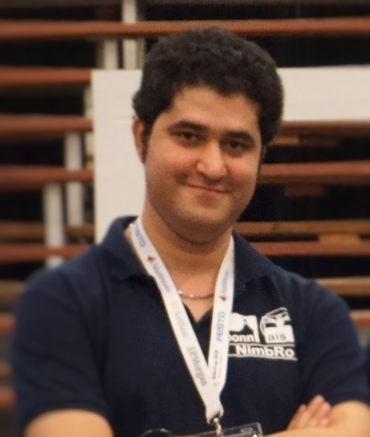}}}
	\end{minipage}
	\hfill %
	\begin{minipage}[b]{3.85truein}
		{{\bf Hafez Farazi} received his M.S. degree in Computer Sciences, Intelligence Systems, from the Amirkabir University of Technology, Iran in 2014. Since 2015, he has been conducting his doctoral studies at the Autonomous Intelligent Systems Institute of the University of Bonn, Germany, where he has actively contributed to humanoid soccer research and Anticipating Human Behavior project. He is a member of the Technical Committee in RoboCup.\hfilneg}
\end{minipage} } %

\vspace*{4.8pt}  
\noindent

Hafez Farazi is the author of over 20 technical
publications. His research interests include machine vision, machine learning, video prediction, 
localization, and bipedal locomotion applied to the field of humanoid and mobile robotics.

\clearpage

\noindent%
\parbox{5truein}{
\begin{minipage}[b]{1truein}
\centerline{{\includegraphics[width=1in]{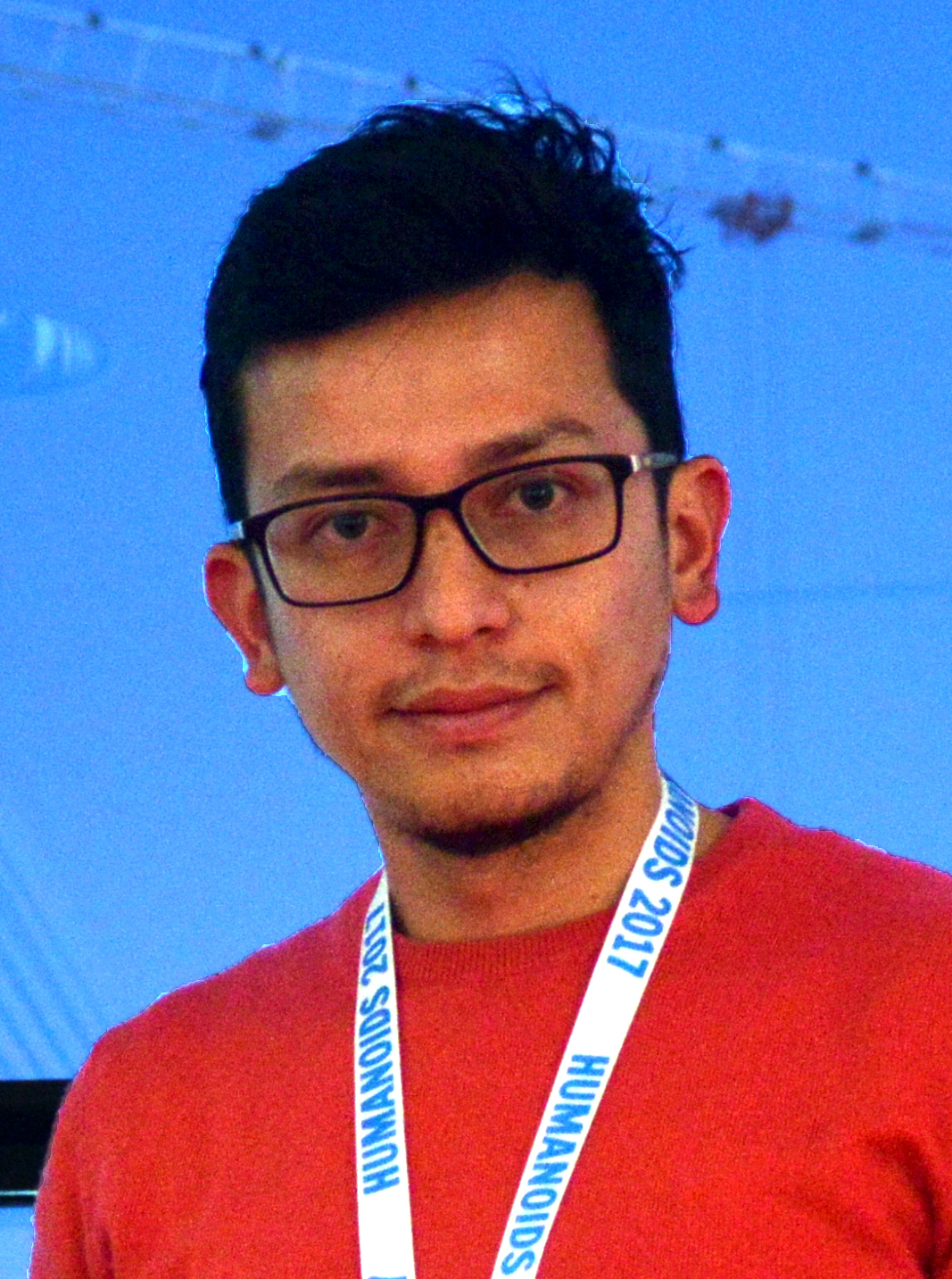}}}
\end{minipage}
\hfill %
\begin{minipage}[b]{3.85truein}
{{\bf Diego Rodriguez} received his M.S. degree in Robotics, Cognition and Intelligence
	from the Technical University of Munich, Germany in 2016. Since that year, he 
	has been conducting his doctoral studies at the Autonomous Intelligent Systems 
	Institute of the University of Bonn, Germany, where he has actively contributed
	in international projects such as the CENTAURO project, and he has been the winner of 
	several robotics competitions including the MBZIRC challenge and RoboCup.\hfilneg}
\end{minipage} } %

\vspace*{4.8pt}   
\noindent
Diego Rodriguez is the author of over 15 technical
publications. His research interests include learning, grasping, manipulation and locomotion
applied to the field of humanoid and mobile robotics.

\vspace*{13pt}  
\noindent%
\parbox{5truein}{
	\begin{minipage}[b]{1truein}
		\centerline{{\includegraphics[width=1in]{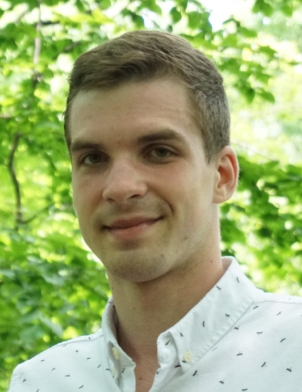}}}
	\end{minipage}
	\hfill %
	\begin{minipage}[b]{3.85truein}
		{{\bf Dmytro Pavlichenko} received his M.S. degree in Computer Science
			from the University of Bonn, Germany in 2016. Since 2017, he has been conducting his doctoral studies at the Autonomous Intelligent Systems Institute of the University of Bonn, Germany, where he has actively participated in international projects such as the CENTAURO project, and contributed to winning of several robotics competitions including the MBZIRC 2017 and RoboCup 2016/2017/2018.\hfilneg}
	\end{minipage} } %
	
\vspace*{4.8pt}  
\noindent
Dmytro Pavlichenko is the author of numerous technical publications in the field of robotics. 
His main research interests lie in the field of robotic manipulation planning, path planning and machine learning.

\vspace*{13pt}  
\noindent%
\parbox{5truein}{
\begin{minipage}[b]{1truein}
\centerline{{\includegraphics[width=1in]{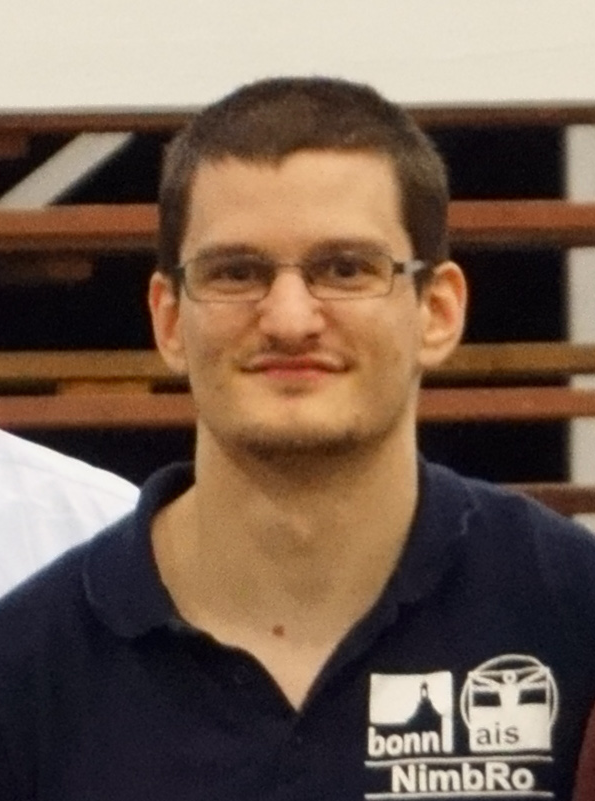}}}
\end{minipage}
\hfill %
\begin{minipage}[b]{3.85truein}
{{\bf Philipp Allgeuer} received his Honours degree in Mechatronic Engineering 
in combination with Mathematical and Computer Sciences from the University of 
Adelaide, Australia in 2012, with majors in robotics and pure mathematics. From 
2013 onwards he was a PhD researcher and candidate at the Autonomous Intelligent 
Systems group at the University of Bonn, Germany, focusing on the dynamic 
bipedal locomotion of humanoid robots, and making contributions to various 
projects and competitions,\hfilneg} 
\end{minipage} } %

\vspace*{4.8pt}   
\noindent 
including in particular RoboCup. He has published numerous papers in the field 
of state estimation and humanoid walking, and has further published to multiple 
venues about novel contributions to the field of rotation formalisms in 3D.

\clearpage

\noindent%
\parbox{5truein}{
\begin{minipage}[b]{1truein}
\centerline{{\includegraphics[width=1in]{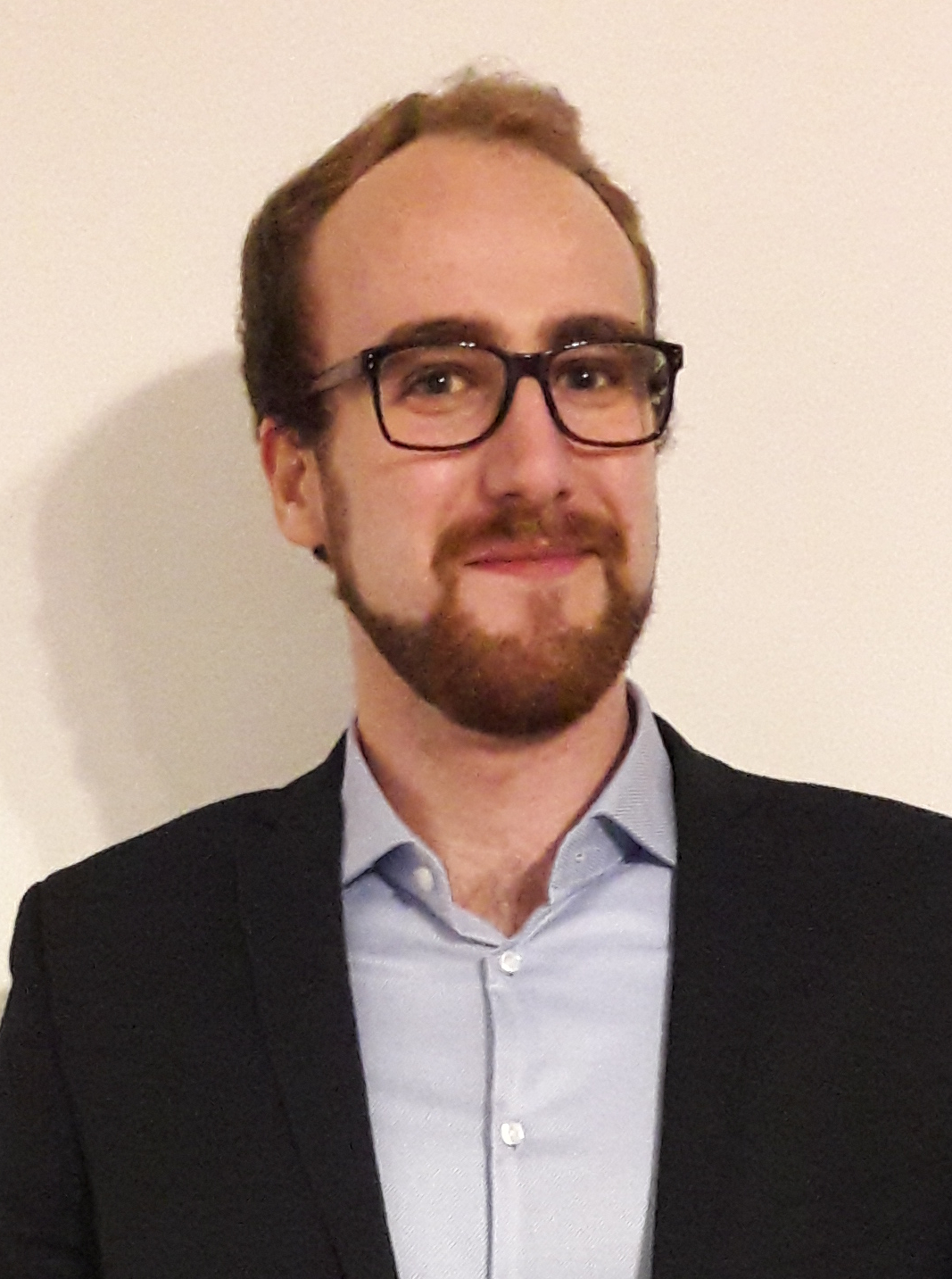}}}
\end{minipage}
\hfill %
\begin{minipage}[b]{3.85truein}
{{\bf Andr\'e Brandenburger} received his B.S. degree in Computer Science from the University of Bonn, Germany in 2018. Since that year, he has continued with his Master studies with a focus on Intelligent Systems. From 2016 onwards, he has been employed by the Autonomous Intelligent Systems group of the University of Bonn, where he actively contributed to winning several robotics competitions and supported research about humanoid robotics and machine learning. \hfilneg}
\end{minipage} } %

\vspace*{4.8pt}  
\noindent
Andr\'e Brandenburger is the author of numerous technical publications in the field of robotics. His main research interests lie in the fields of machine learning, computer vision, robotics and sensor data fusion. 

\vspace*{13pt}  
\noindent%
\parbox{5truein}{
	\begin{minipage}[b]{1truein}
		\centerline{{\includegraphics[width=1in]{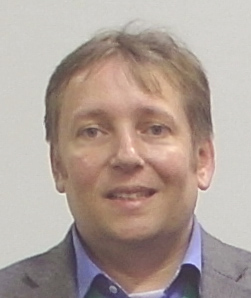}}}
	\end{minipage} 
	\hfill %
	\begin{minipage}[b]{3.85truein}
		{{\bf Sven Behnke} received his M.S. degree in Computer Science (Dipl.-Inform.) in 1997 from Martin-Luther-Universit\"at Halle-Wittenberg. 
		In 2002, he obtained a Ph.D. in Computer Science (Dr. rer. nat.) from Freie Universit\"at Berlin. He spent the year 2003 as postdoctoral 
		researcher at the International Computer Science Institute, Berkeley, CA. From 2004 to 2008, Professor Behnke headed the Humanoid 
		Robots Group at\hfilneg}   
	\end{minipage} } %
	 
\vspace*{4.8pt}  
\noindent
Albert-Ludwigs-Universit\"at Freiburg. Since April 2008, he is professor for Autonomous Intelligent Systems at the 
University of Bonn and director of the Institute of Computer Science VI. His research interests include cognitive robotics, computer vision, and machine learning.

\vfill\eject

\end{document}